\pdfoutput=1

\documentclass[11pt]{article}

\usepackage[preprint]{acl}

\definecolor{customRed}{RGB}{209,41,32}

\usepackage{times}
\usepackage{latexsym}

\usepackage[T1]{fontenc}

\usepackage[utf8]{inputenc}

\usepackage{microtype}

\usepackage{inconsolata}

\usepackage{xurl} 
\usepackage{hyperref} 
\usepackage{longtable} 

\usepackage{amsmath}

\usepackage{graphicx}

\usepackage{xcolor}

\usepackage{tabularx}

\usepackage{booktabs}
\usepackage{siunitx}
\usepackage{array}

\usepackage{caption}
\usepackage{subcaption}

\usepackage{multirow}

\title{Unraveling Babel: Exploring Multilingual Activation Patterns of LLMs\\and Their Applications}

\author{
Weize Liu$^{1}$, Yinlong Xu$^{1}$, Hongxia Xu$^{1,2}$\thanks{\, Corresponding authors.}, Jintai Chen$^{3}$, Xuming Hu$^{4,5}$\footnotemark[1], Jian Wu$^{1,6}$\\
$^{1}$Liangzhu Laboratory, Zhejiang University~~~
$^{2}$AI Research Center, WeDoctor Cloud\\
$^{3}$University of Illinois Urbana-Champaign\\
$^{4}$The Hong Kong University of Science and Technology (Guangzhou)\\
$^{5}$The Hong Kong University of Science and Technology\\
$^{6}$School of Public Health, Zhejiang University\\
\texttt{\{weizeliu1115,xuminghu97\}@gmail.com} \ \ \ \texttt{einstein@zju.edu.cn}
}

\begin{document}
\maketitle
\begin{abstract}
Recently, large language models (LLMs) have achieved tremendous breakthroughs in the field of NLP, but still lack understanding of their internal neuron activities when processing different languages. We designed a method to convert dense LLMs into fine-grained MoE architectures, and then visually studied the multilingual activation patterns of LLMs through expert activation frequency heatmaps. Through comprehensive experiments on different model families, different model sizes, and different variants, we analyzed the similarities and differences in the internal neuron activation patterns of LLMs when processing different languages. Specifically, we investigated the distribution of high-frequency activated experts, multilingual shared experts, whether multilingual activation patterns are related to language families, and the impact of instruction tuning on activation patterns. We further explored leveraging the discovered differences in expert activation frequencies to guide sparse activation and pruning. Experimental results demonstrated that our method significantly outperformed random expert pruning and even exceeded the performance of unpruned models in some languages. Additionally, we found that configuring different pruning rates for different layers based on activation level differences could achieve better results. Our findings reveal the multilingual processing mechanisms within LLMs and utilize these insights to offer new perspectives for applications such as sparse activation and model pruning.
\end{abstract}

\section{Introduction}

Large language models (LLMs), proficient in utilizing a wide range of linguistic structures, have attained notable advancements in the field of natural language processing~\citep{zhao2023survey}. However, how an LLM uses multiple languages within one single structure remains elusive. Previous studies in neuroscience suggest that although there is considerable overlap in the brain regions involved in processing different languages, there are discernible differences~\citep{crinion2006language,videsott2010speaking,friederici2011brain}. Specifically, some certain regions appear to be specialized for particular languages, while some regions are language-agnostic. In the field of NLP, several recent studies~\citep{tang2024language,zhao2024large,kojima2024multilingual} have investigated language-specific neurons within LLMs. But the internal mechanisms of LLMs in processing different languages and how to leverage these mechanisms remain insufficiently explored. 

We still lack an intuitive understanding of the internal neuron activity of LLMs when processing different languages---it remains like a black box. This is due to the difficulty of decomposing LLMs into recognizable components. This presents a significant obstacle for us to better utilize LLMs. Therefore, we aim to investigate the differences and connections in internal neuron activations of LLMs when processing different languages.

We refer to the different activation patterns exhibited by internal neurons of LLMs when processing different languages as multilingual activation patterns. To open the black box and intuitively understand and explain the internal multilingual activation patterns of LLMs, we devised a method that involves converting dense LLMs into fine-grained MoE architectures and then calculating the activation frequencies of experts. Subsequently, we visualize the multilingual activation patterns of LLMs by heatmaps.

We conducted a comprehensive experimental study on the multilingual activation patterns of different LLMs, including various model families (Llama 2, Llama 3, Mistral), different sizes (7B--70B), and different variants (pre-trained and instruction tuned variants). We analyzed the distribution of high-frequency activated experts, multilingual shared experts, whether the activation patterns of different languages are related to language families, and the impact of instruction tuning on activation patterns.

Considering that dense LLMs can be transformed into MoE architectures~\citep{zhang-etal-2022-moefication}, based on obtaining multilingual activation patterns, we further propose to leverage the activation frequency differences of various experts to guide sparse activation and pruning, aiming to minimize the amount of computation and inference cost, while maintaining model performance as much as possible. We advocate using only high-frequency experts for inference, excluding the parameters of other experts to achieve sparse activation and pruning. To this end, we propose two specific pruning schemes: pruning based on frequency thresholds and pruning based on frequency sorting. In experiments across various models and metrics, our method significantly outperformed random expert pruning and even exceeded the performance of the original unpruned models in some languages. This further validates the effectiveness of the multilingual activation patterns discovered by our method, providing a new feasible path for model pruning. Meanwhile, we found that equal proportional pruning at each layer is inferior to that of unequal pruning, which verifies the inter-layer differences in activation levels of LLMs. Therefore, we suggest configuring different pruning rates for different layers based on differences in activation levels.

Overall, we have not only analyzed the multilingual activation patterns within various LLMs, but also explored their connections with language families and instruction tuning. Furthermore, based on the identified multilingual activation patterns, we provide a new and effective approach to achieve sparse activation and pruning.

\section{Related Work}

Initially, in the task of neural machine translation (NMT) using small language models, some works~\citep{lin2021learning,zhang2021share,xie2021importance} explored language-specific components. \citet{armengol-estape-etal-2022-multilingual} studied the exceptional performance of GPT-3 in Catalan, despite the small proportion of Catalan in the training corpus. \citet{bhattacharya-bojar-2023-unveiling} evaluated the language specificity of the detector in the XGLM model using a parallel corpus of Czech and English, by conceptualizing the FFN as a system of detector, selector, and combiner. Several recent works~\citep{tang2024language,zhao2024large,kojima2024multilingual} investigated the existence of language-specific neurons in LLMs and the modification of these neurons. \citet{zhang2024unveiling} identified a core region corresponding to linguistic competence in LLMs through fine-tuning with various languages. In terms of MoE, the LLaMA-MoE model~\citep{llama-moe-2023} explored transforming Llama 2 7B into an MoE model and its continued training. However, exploring the multilingual activation patterns within LLMs and how to leverage these patterns remains an area that has not been sufficiently investigated.

\section{Exploring Multilingual Activation Patterns in LLMs}

\citet{zhang-etal-2022-moefication} proposed MoEfication, exploring the conversion of the feed-forward networks (FFNs) in pre-trained Transformers into MoE structures without altering the original model parameters, and maintaining the performance on downstream tasks by conditionally selecting experts. Inspired by this, we propose to study the internal neuron activation patterns of LLMs when processing different languages by visualizing the expert activation frequency and its variations after converting the model to an MoE structure.

\paragraph{Expert Construction} Our first step is to split the parameters of the FFNs into different experts. The FFNs in Llama/Mistral models comprise three layers: up-projection, gate-projection, and down-projection, as illustrated in the schematic diagram presented in Appendix~\ref{sec:appendix_a}. Based on the intuition that the neurons with similar parameters exhibit similar activation patterns, we adopt the parameter clustering split method to cluster the parameters of each FFN layer, dividing them into different experts. Specifically, we perform balanced $K$-Means clustering~\citep{malinen2014balanced} on the parameters of the up-projection layer, dividing it into 256 clusters. Then, we divide the neurons and their parameters of down-projection and gate-projection layers into different clusters based on the clustering results of the up-projection layer. Dividing neurons into different experts can significantly reduce the computational burden in subsequent experiments and enable us to directly observe the internal multilingual activation patterns of LLMs through heatmap visualization. To achieve fine-grained parameter splits while maintaining the visualization effects of heatmaps, we divide each FFN layer into 256 experts.

\begin{figure*}[htb]
  \centering
  \begin{minipage}{\linewidth}
    \centering
    \includegraphics[width=\linewidth]{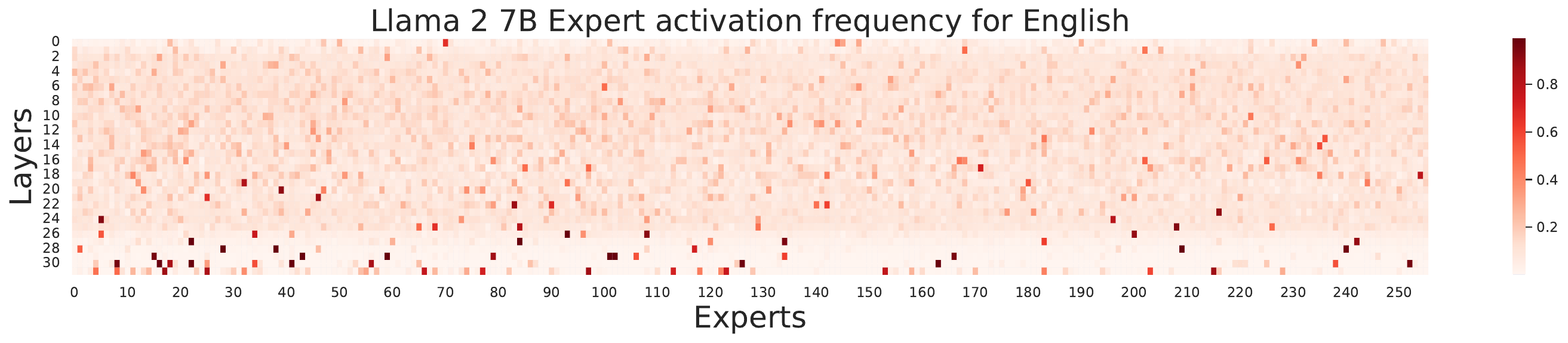}
    \vspace{-6mm}
  \end{minipage}

  \begin{minipage}{\linewidth}
    \centering
    \includegraphics[width=\linewidth]{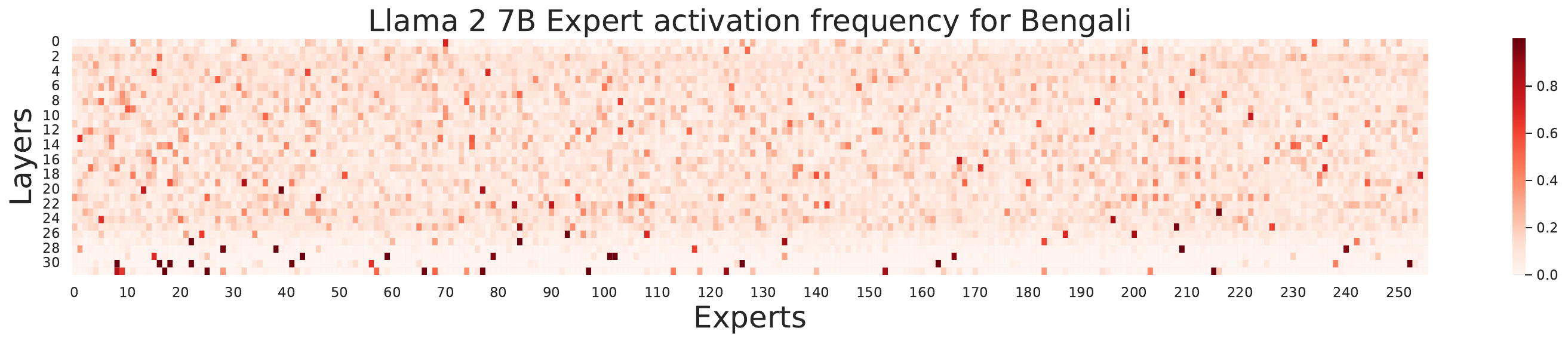}
    \vspace{-6mm}
  \end{minipage}

  \begin{minipage}{\linewidth}
    \centering
    \includegraphics[width=\linewidth]{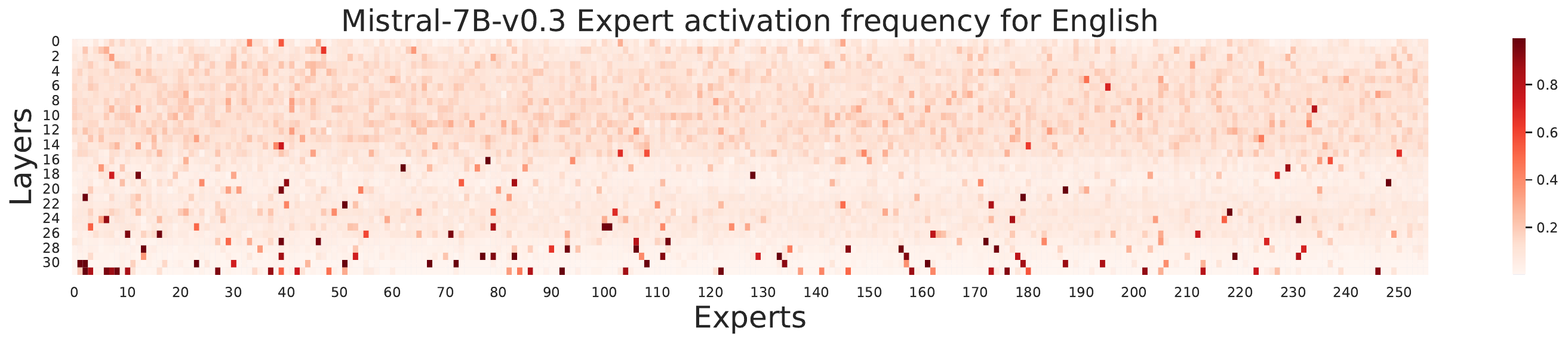}
    \vspace{-6mm}
  \end{minipage}

  \begin{minipage}{\linewidth}
    \centering
    \includegraphics[width=\linewidth]{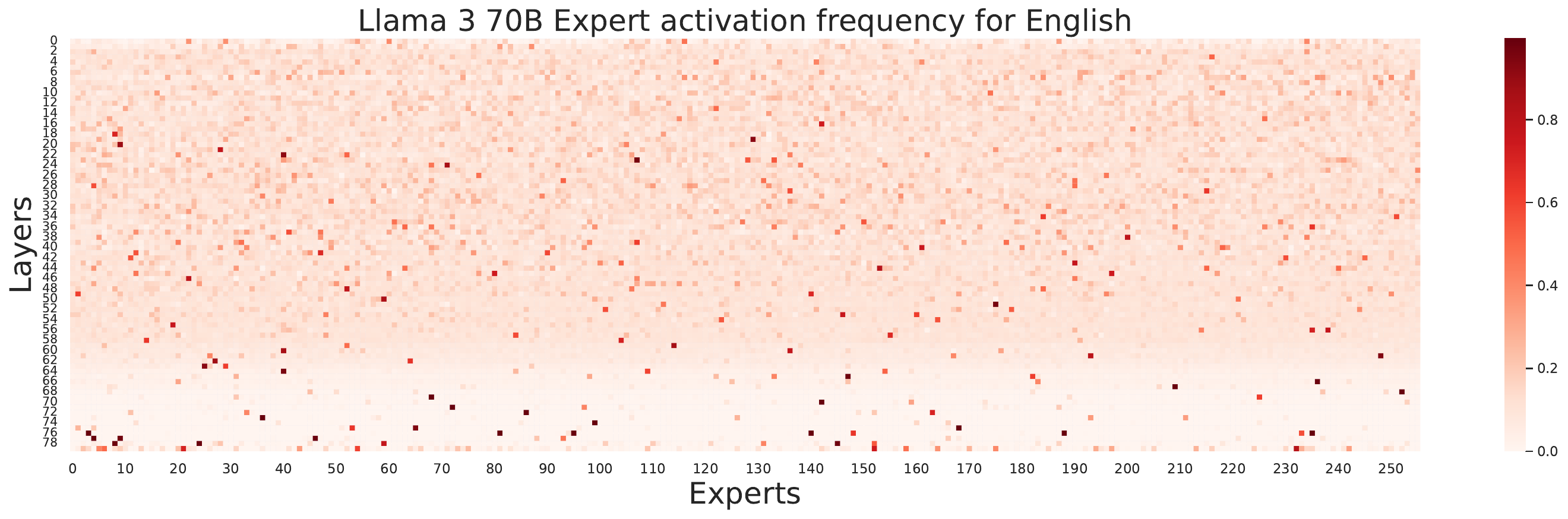}
    \vspace{-8mm}
  \end{minipage}
  
  \caption{Heatmaps of activation patterns for some models and languages. Each heatmap is 32*256 (number of layers * number of experts), with darker colors indicating higher activation frequencies.}
  \label{fig:heatmap_red}
  \vspace{-4mm}
\end{figure*}

\paragraph{Cross-layer Expert Selection} Next, we design the cross-layer expert selection method to identify experts with higher activation levels and frequencies as the language-specific experts for each language. LLMs contain multiple FFN layers, but the MoEfication method is limited to selecting experts within a single FFN layer, which fails to reflect the differences in activation levels across layers of varying depth within the LLM. Therefore, we need to extend the MoEfication method to cross-layer expert selection. For each input token, we use the sum of the activation values of all neurons for each expert as the score of this expert, representing the activation level. Given the direct incomparability of activation magnitudes across different FFN layers, to facilitate a global comparison of activation levels across layers, we perform a Z-score normalization on the scores of all experts within each layer. Subsequently, we rank the scores of all experts across all layers. We select approximately the top 10\% of experts (for the 7B/8B models, we select the top 800, and for the 70B model, we select the top 2000) as the activated experts for a given input token, increasing their activation counts by 1. By normalizing and then performing cross-layer comparison, we better identified the experts whose activation values stood out relative to other experts.

\paragraph{Expert Activation Patterns} After extensive testing with a multitude of tokens, we calculate the activation frequency for each expert (activation count/total number of tokens). Finally, the activation frequencies of all experts across $n$ layers are compiled into an $n \times 256$ activation matrix for heatmap visualization.

\section{Experiments}

\begin{figure*}[ht]
  \centering
  \begin{subfigure}[b]{0.325\textwidth}
    \includegraphics[width=\linewidth]{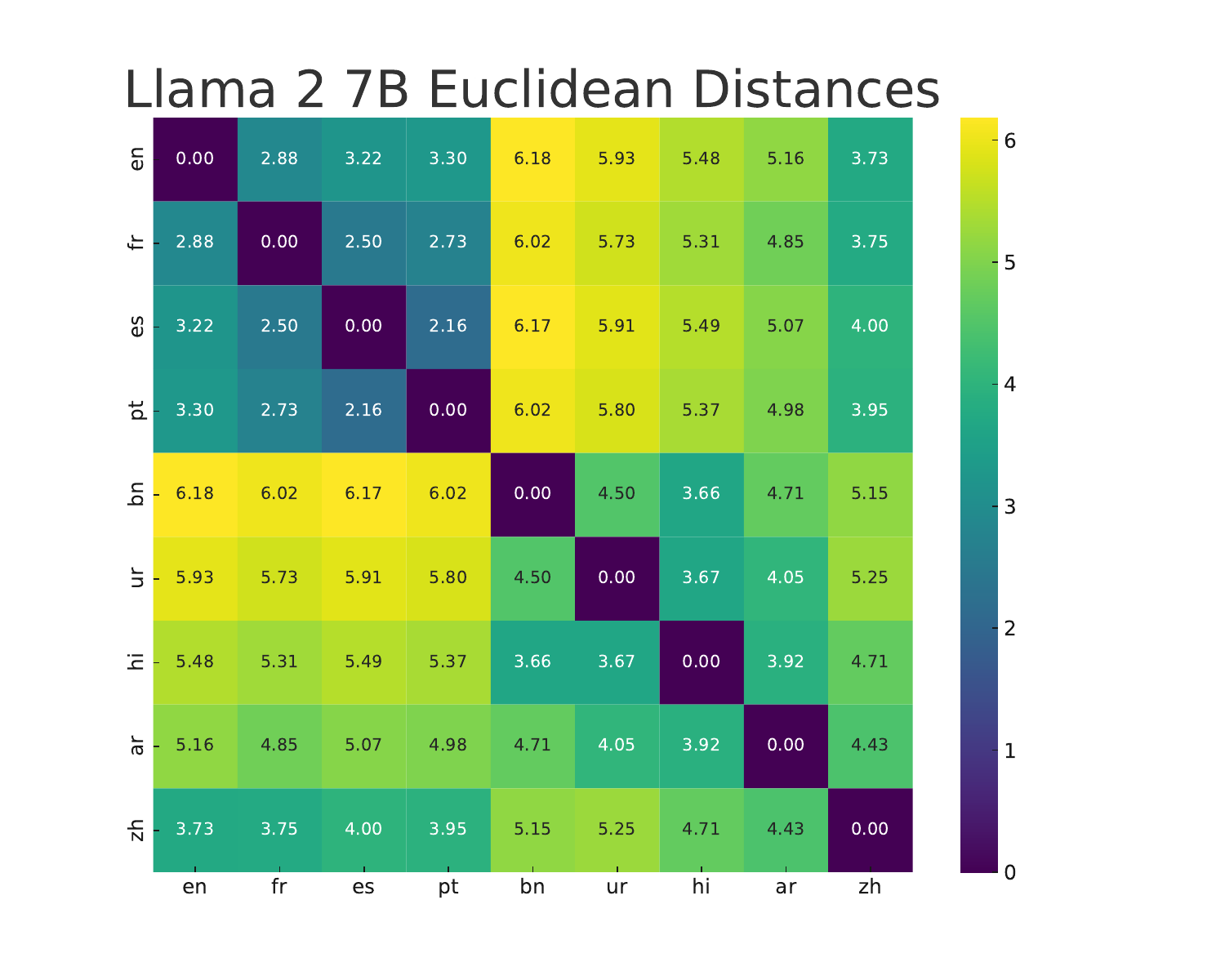}
  \end{subfigure}
  \vspace{-2mm}
  \begin{subfigure}[b]{0.325\textwidth}
    \includegraphics[width=\linewidth]{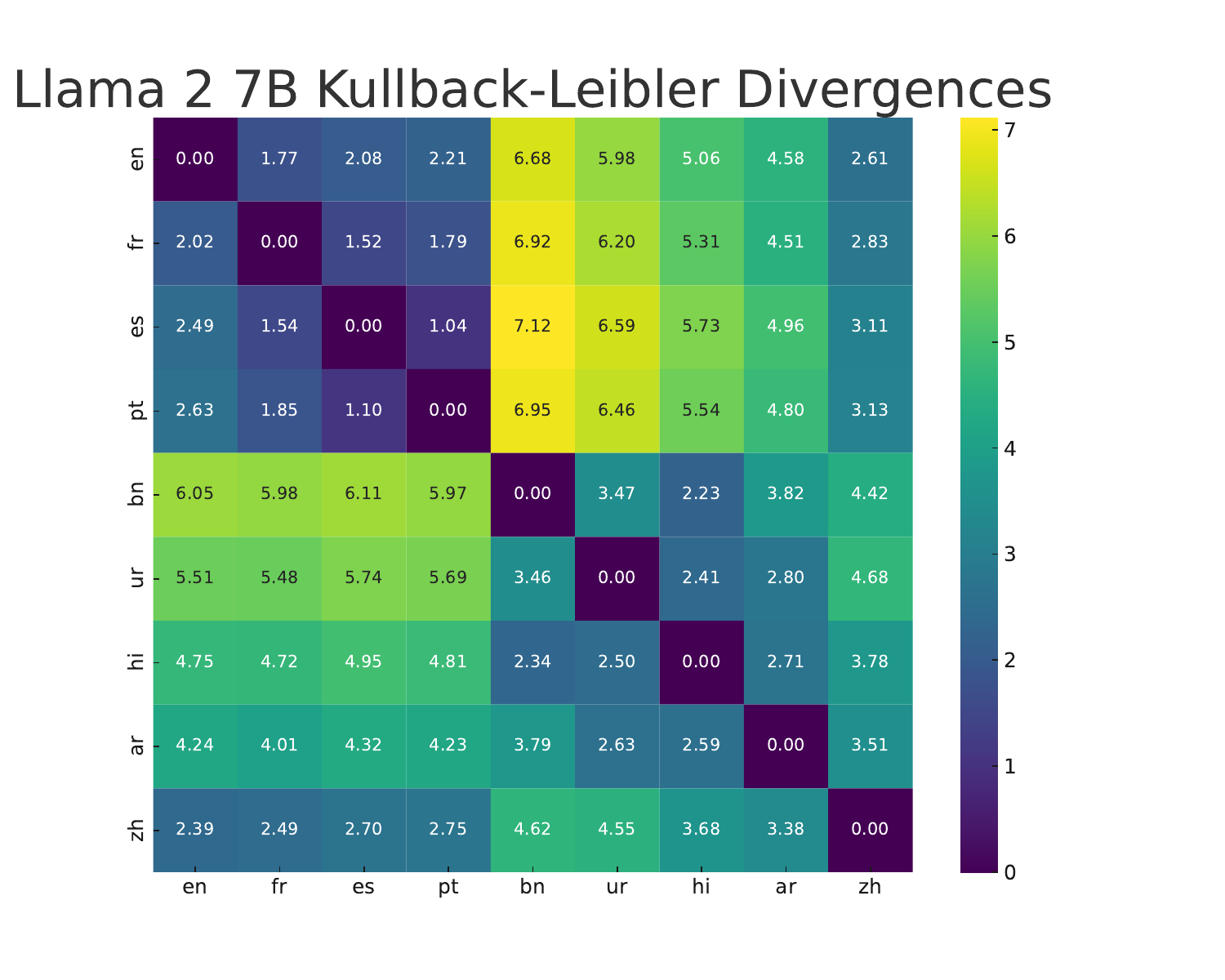}
  \end{subfigure}
  \begin{subfigure}[b]{0.325\textwidth}
    \includegraphics[width=\linewidth]{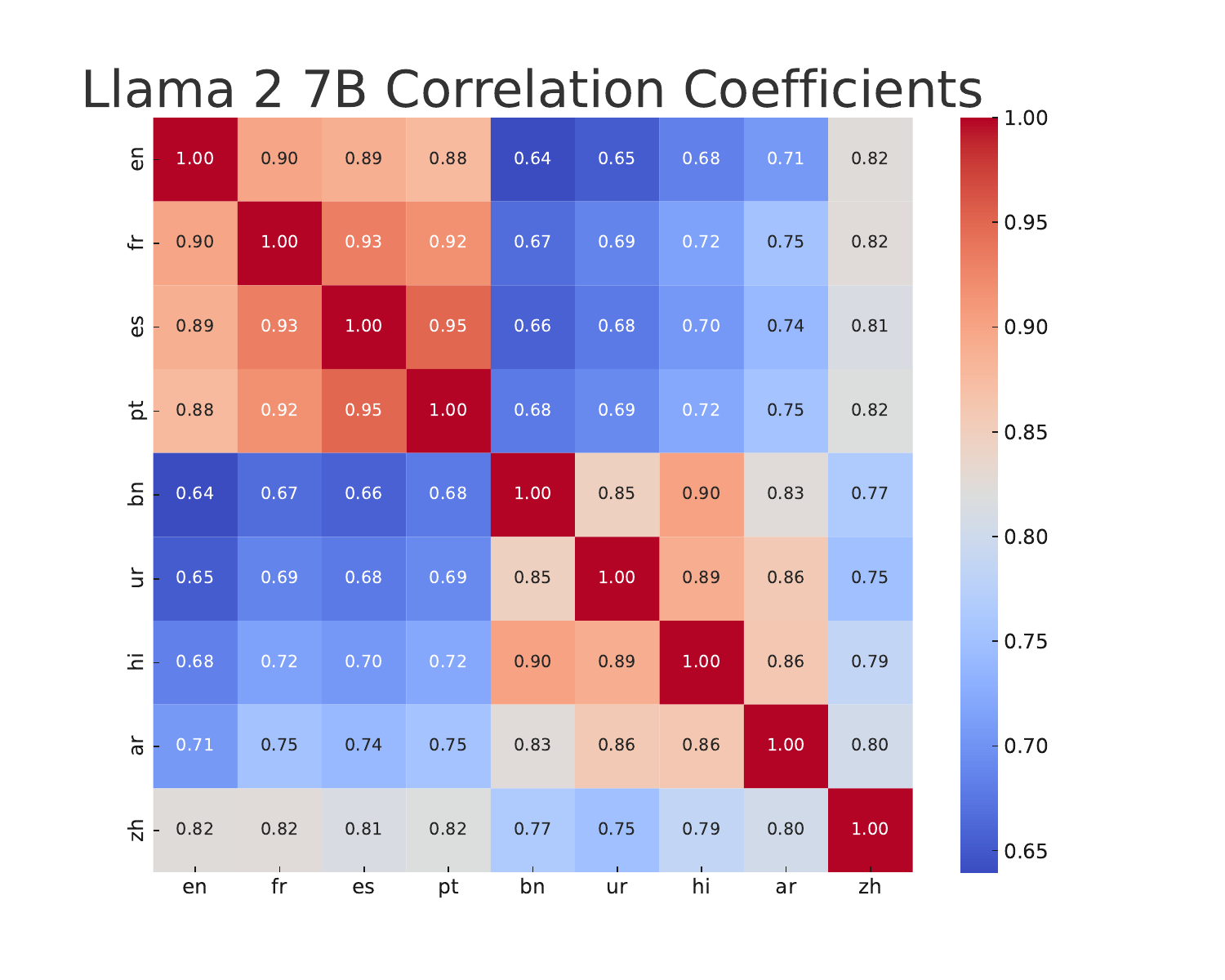}
  \end{subfigure}
  \begin{subfigure}[b]{0.325\textwidth}
    \includegraphics[width=\linewidth]{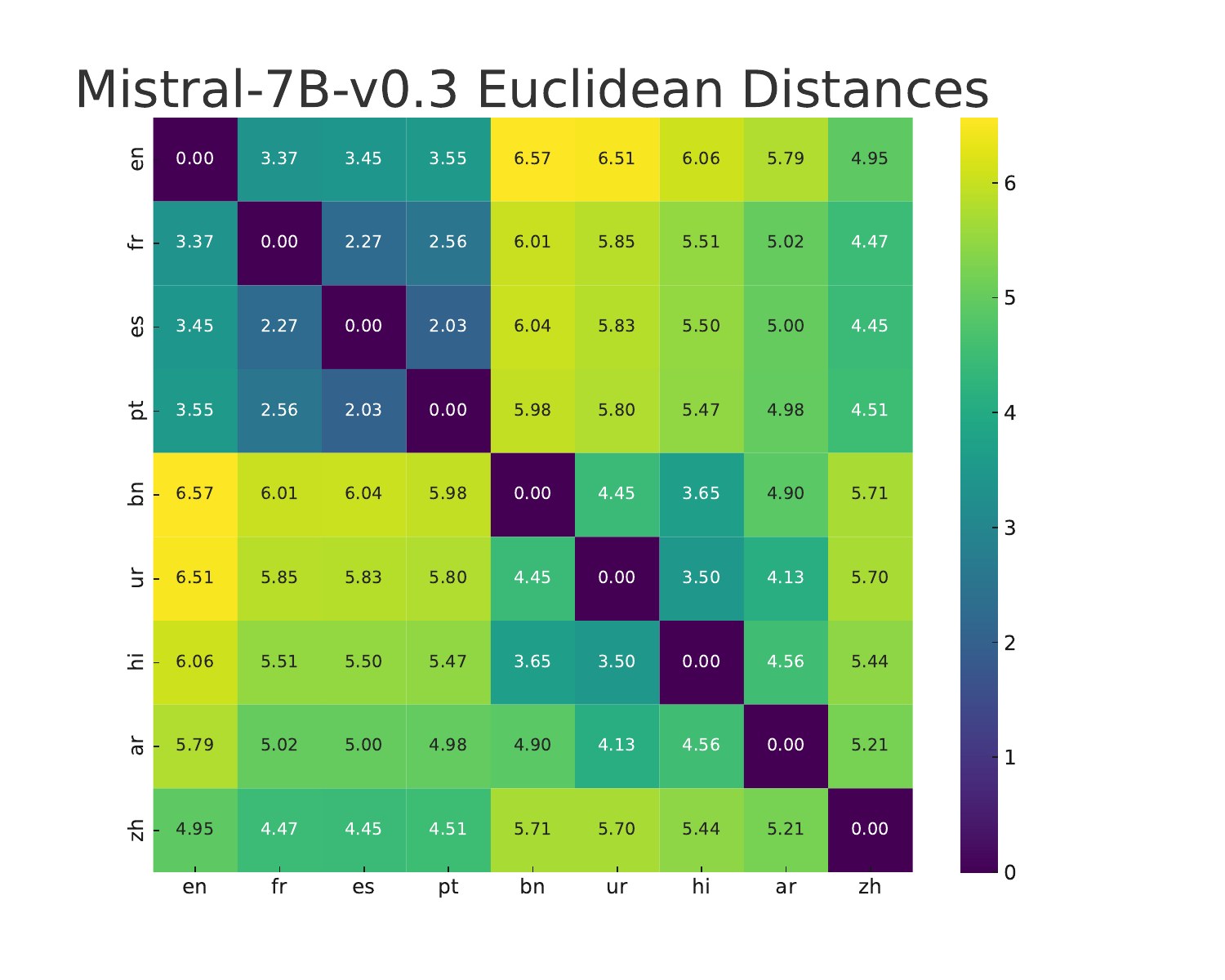}
  \end{subfigure}
  \begin{subfigure}[b]{0.325\textwidth}
    \includegraphics[width=\linewidth]{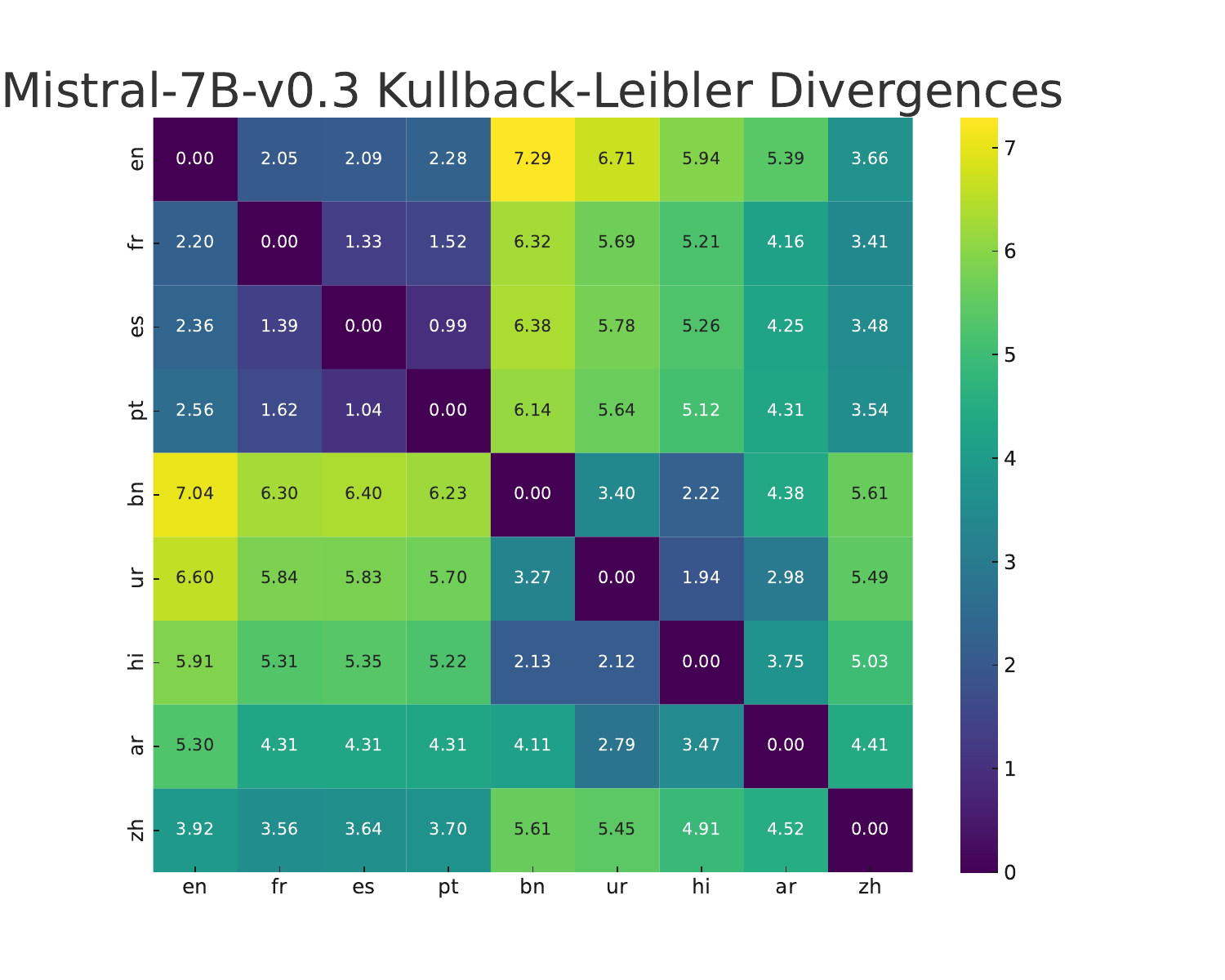}
  \end{subfigure}
  \begin{subfigure}[b]{0.325\textwidth}
    \includegraphics[width=\linewidth]{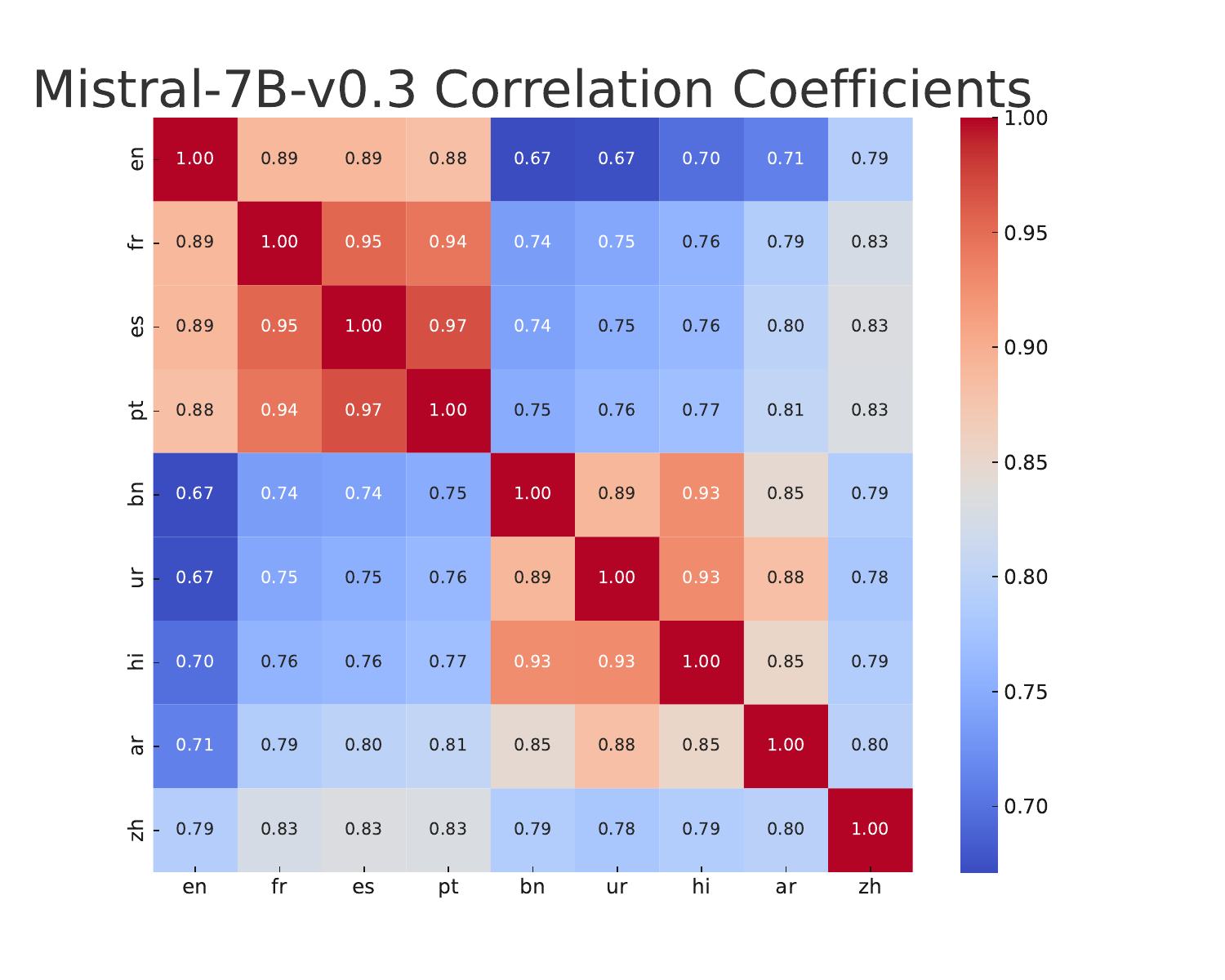}
  \end{subfigure}
  \vspace{-4mm}
  \caption{Heatmaps of similarity between activation pattern matrices for different languages in Llama 2 7B and Mistral-7B-v0.3. Each value in the Euclidean distance heatmaps represents the square root of the sum of the squares of the differences between corresponding elements of two matrices. Each value in the Kullback-Leibler (KL) divergence heatmaps represents the cumulative sum of the KL divergences computed row-wise between two matrices. Each value in the correlation coefficients heatmaps represents the mean of the Pearson correlation coefficients calculated row-wise between two matrices. The smaller the Euclidean distance/KL divergence, the more similar the two matrices are. The larger the correlation coefficient, the more similar the two matrices are.}
  \label{fig:combined_heatmaps}
  \vspace{-4mm}
\end{figure*}

\paragraph{Models.} Our experiments involve four models: Llama 2 7B~\citep{touvron2023llama}, Llama 3 8B, Llama 3 70B, and Mistral-7B-v0.3~\citep{jiang2023mistral}, along with their respective instruction tuning variants, to conduct a comprehensive study on the effects of different model families, model sizes, and instruction tuning.

\paragraph{Data.} We selected the nine most widely spoken languages from the 46 languages contained in the ROOTS corpus~\citep{laurenccon2022bigscience} for experiments. The data sources are in Appendix~\ref{sec:appendix_b}. The language families and genera of these languages are presented in Table~\ref{tab:data}. For each language, we test its activation pattern using 10,000 data samples, with a maximum input length of 200 tokens per sample.

\begin{table}[ht]
\centering
\begin{tabular}{lll}
\toprule
Family & Genus & Language \\
\midrule
Indo-European & Germanic & English (en) \\
Indo-European & Romance & French (fr) \\
Indo-European & Romance & Spanish (es) \\
Indo-European & Romance & Portuguese (pt) \\
Indo-European & Indic & Bengali (bn) \\
Indo-European & Indic & Urdu (ur) \\
Indo-European & Indic & Hindi (hi) \\
Afro-Asiatic & Semitic & Arabic (ar) \\
Sino-Tibetan & Chinese & Chinese (zh) \\
\bottomrule
\end{tabular}
\caption{The language families and genera. The ISO 639-1 language codes for each language are shown in parentheses in the Language column.}
\label{tab:data}
\end{table}

\begin{figure*}[ht]
  \centering
  \begin{minipage}{\linewidth}
    \centering
    \includegraphics[width=\linewidth]{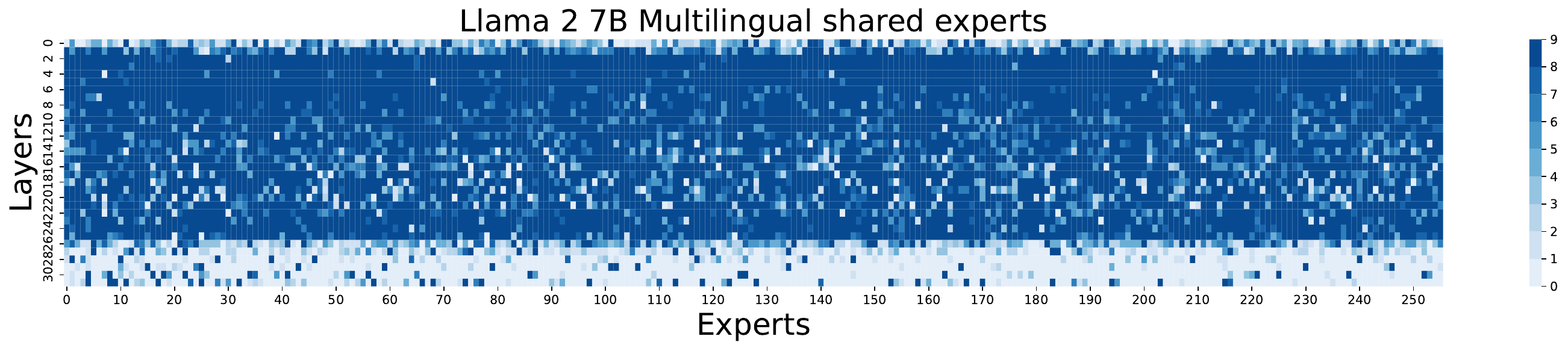}
    \vspace{-6mm}
  \end{minipage}

  \begin{minipage}{\linewidth}
    \centering
    \includegraphics[width=\linewidth]{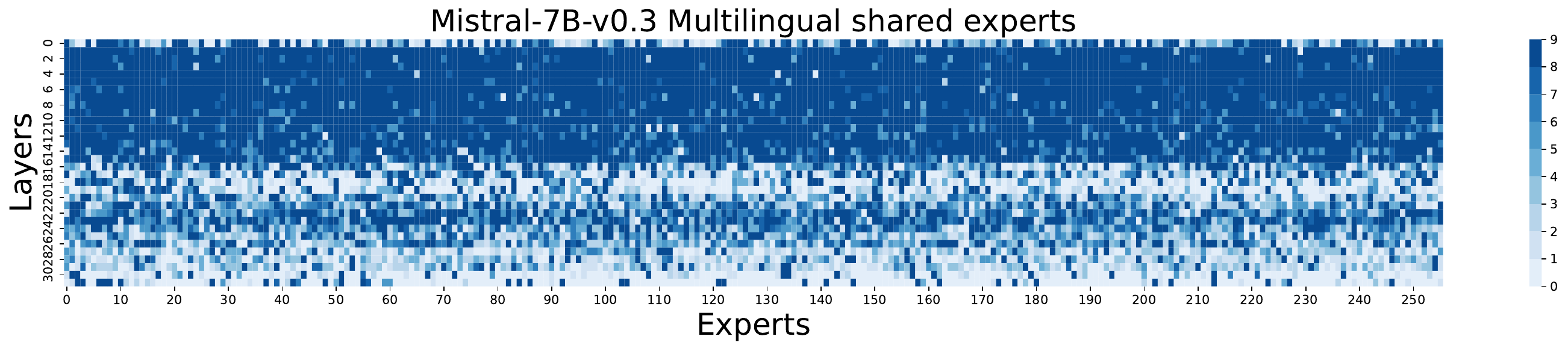}
    \vspace{-8mm}
  \end{minipage}
  
  \caption{The heatmaps of multilingual shared experts for Llama 2 7B and Mistral-7B-v0.3. The color shade of each cell indicates how many languages the expert is a high-frequency expert in.}
  \label{fig:heatmap_shared}
  \vspace{-4mm}
\end{figure*}

\subsection{Multilingual Activation Patterns of LLMs}
\label{subsec:41}

We present the activation pattern heatmaps of some models and languages in Figure~\ref{fig:heatmap_red}, with the heatmaps of the remaining languages and models provided in Appendix~\ref{sec:appendix_c}. We define layers closer to the input as shallow layers, with the shallowest layer being layer 0, and layers closer to the output as deep layers, with the deepest layer being layer 31/79. From the figure, it can be observed that for the Llama family models, the activation frequency differences between different experts in the shallow and middle layers are relatively small. But starting from a certain layer, the sparsity of expert activations significantly increases. Most experts exhibit lower activation frequencies, but a few experts have very high frequencies, indicating that significant differences in activation levels occur between different experts. Unlike the Llama family models, the Mistral model exhibits a distinct light-colored band in the middle layers, indicating that there is a portion of the middle layers with generally lower activation levels. By examining the heatmaps of other languages in Appendix~\ref{sec:appendix_c}, we can observe that the positions of these light-colored bands in the Llama and Mistral models are consistent across all languages. However, the activation patterns for the same model also vary across different languages. For example, in Figure~\ref{fig:heatmap_red}, it is evident that the activation frequency of shallow and intermediate experts is generally higher for Bengali compared to English. By observing the activation patterns of different languages, we find that regardless of the model, the activation frequency of shallow- and middle-layer experts for English, French, Portuguese, and Spanish is significantly lower than that for Bengali, Urdu, and Hindi. This led us to speculate whether the activation patterns of different languages are related to their respective language families.

\subsection{Are there connections at the language family level in the activation patterns?}

To study the relationship between the multilingual activation patterns of LLMs and the language families and genera shown in Table~\ref{tab:data}, we calculated the similarity between the activation pattern matrices of different languages. We employed Euclidean distance, Kullback-Leibler (KL) divergence, and Pearson correlation coefficient as three measures to comprehensively reflect the similarity between different activation pattern matrices. The results of Llama 2 7B and Mistral-7B-v0.3 are illustrated in Figure~\ref{fig:combined_heatmaps}, with similar results for other models listed in Appendix~\ref{sec:appendix_d}. The results indicate that for all models, the activation patterns of three languages belonging to the Romance genus (French (fr), Spanish (es), Portuguese (pt)) and three languages of the Indic genus (Bengali (bn), Urdu (ur), Hindi (hi)) exhibit high similarity within the same genus. However, there is a significant difference in the activation patterns between these two genera. The activation pattern of English belonging to the Germanic genus is closer to that of the three languages belonging to the Romance genus, although there are still differences between them. This may be because English and Romance languages both use the Latin alphabet system and have a large number of loanwords and cognates (usually with the same or similar forms and meanings). Similarly, the activation pattern of Arabic is more similar to the languages of the Indic genus, possibly because Arabic has a large number of loanwords in Bengali, Urdu, and Hindi. Particularly, Urdu uses a modified Arabic alphabet system, with many words and expressions being very similar to Arabic, and both are written from right to left. In summary, we can confirm that the activation patterns of LLMs for different languages are closely related to the language families and genera to which these languages belong. Additionally, they may also be related to the alphabet systems and surface form similarities.

\begin{figure*}[ht]
  \centering
  \begin{minipage}{\linewidth}
    \centering
    \includegraphics[width=\linewidth]{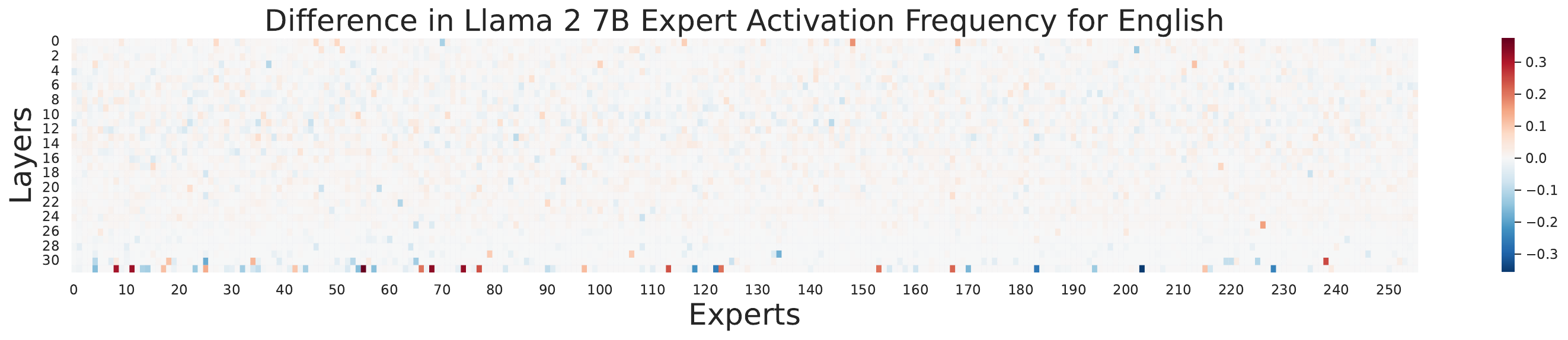}
    \vspace{-6mm}
  \end{minipage}

  \begin{minipage}{\linewidth}
    \centering
    \includegraphics[width=\linewidth]{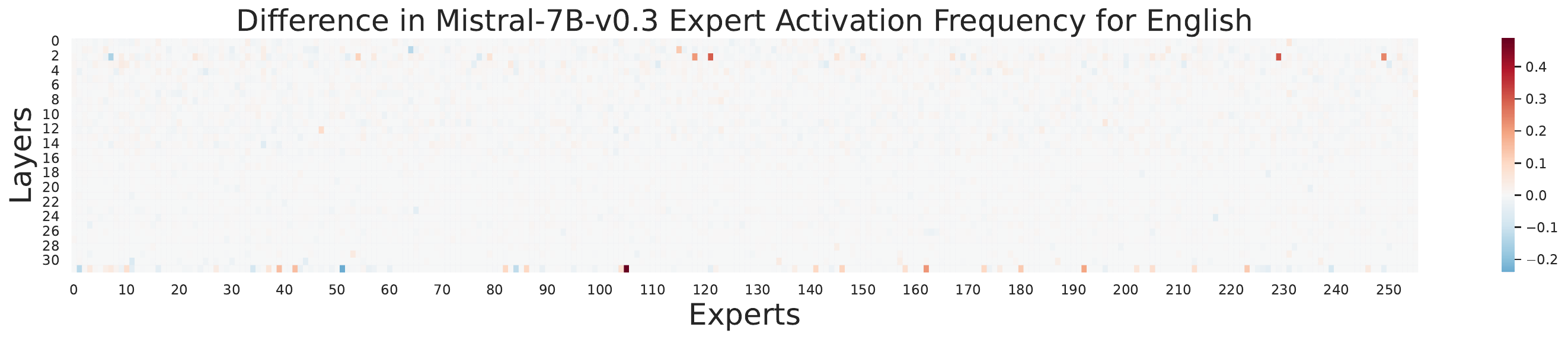}
    \vspace{-8mm}
  \end{minipage}
  
  \caption{Changes in expert activation frequency for the instruction tuning variants of Llama 2 7B and Mistral-7B-v0.3 compared to the original pre-trained models.}
  \label{fig:heatmap_chat}
  \vspace{-6mm}
\end{figure*}

\subsection{Multilingual shared experts}

Based on the previous results, we can observe that some experts are frequently activated across different languages. We are also interested in understanding the distribution of these commonly highly activated experts. We define experts whose activation frequency for a given language is greater or equal to 0.05 as high-frequency experts of that language; otherwise, they are considered low-frequency experts. In Figure~\ref{fig:heatmap_shared}, we display the number of languages in which each expert is a high-frequency expert for Llama 2 7B and Mistral-7B-v0.3. For experts who are high-frequency experts in all 9 languages, we call them multilingual shared experts. For Llama 2 7B, the density of multilingual shared experts is significantly high in the middle layers (2--26 layers), while both shallower (0--2 layers) and deeper (26--32 layers) have sparse dark regions and more light regions. The results for Llama 3 8B and Llama 3 70B are provided in Appendix~\ref{sec:appendix_e}. Their overall trends are consistent with those of Llama 2 7B. As observed in Section~\ref{subsec:41}, the Mistral model has a low-activation band in the middle layers, but multilingual shared experts are still concentrated in the middle layers, while the shallow and deep layers are relatively sparse. This suggests that a large amount of neurons in the middle layers is language-independent and likely serves non-language-specific functions, such as various types of knowledge and abstract concepts that are independent of any particular language. In contrast, the activation patterns in the shallower and deeper layers are more closely related to language, exhibiting different activation patterns for different languages, indicating that these layers may undertake more language-specific processing functions. This may be because, in neural networks, the shallow layers typically learn low-level features of the input data, such as vocabulary, grammatical structures, and linguistic rules of different languages. Thus, the shallow layers contain a significant number of language-specific experts. In layers closer to the output, neurons need to generate outputs in specific languages. This necessitates adaptation to the linguistic structures of specific languages, resulting in a substantial number of language-specific experts in the deeper layers.

\begin{table*}[t]
\centering
\small
\begin{tabular}{lccccccc}
\toprule
& \multicolumn{3}{c}{Expert activation frequency $\geq$ 5\%} & \multicolumn{3}{c}{Expert activation frequency $\geq$ 1\%} & \\
\cmidrule(lr){2-4} \cmidrule(lr){5-7}
Language & Proportion & Random & Experts & Proportion & Random & Experts & Origin \\
\midrule
English & 77.4\% & 3329.60 $\pm$ 4646.53 & 12.91 & 91.4\% & 13.57 $\pm$ 1.32 & 8.01 & 6.89 \\
French & 76.7\% & 37.49 $\pm$ 2.61 & 16.54 & 93.3\% & 11.43 $\pm$ 0.23 & 7.62 & 5.83 \\
Spanish & 76.3\% & 52.42 $\pm$ 4.56 & 21.52 & 93.2\% & 12.29 $\pm$ 2.35 & 8.96 & 6.71 \\
Portuguese & 76.0\% & 3445.22 $\pm$ 4804.65 & 22.68 & 93.4\% & 11.11 $\pm$ 0.40 & 8.34 & 6.19 \\
Bengali & 67.3\% & 274466.25 $\pm$ 387831.01 & 14.23 & 90.2\% & 20.18 $\pm$ 16.63 & 4.89 & 3.43 \\
Urdu & 68.5\% & 416160.50 $\pm$ 568821.95 & 13.24 & 90.3\% & 11.82 $\pm$ 2.85 & 7.18 & 4.54 \\
Hindi & 70.6\% & 22575.46 $\pm$ 23516.65 & 8.24 & 90.6\% & 6.31 $\pm$ 0.36 & 4.13 & 3.51 \\
Arabic & 71.4\% & 512693.39 $\pm$ 725010.36 & 13.52 & 90.8\% & 10.11 $\pm$ 0.46 & 7.14 & 4.58 \\
Chinese & 76.5\% & 19964.56 $\pm$ 28191.94 & 17.19 & 92.9\% & 10.32 $\pm$ 0.65 & 8.54 & 5.57 \\
\midrule
Average & 73.41\% & 139191.65 & 15.56 & 91.79\% & 11.90 & 7.20 & 5.25 \\
\bottomrule
\end{tabular}
\caption{The perplexity results of Llama 2 7B. The smaller the value, the better the model performance.}
\label{tab:ppl-7b}
\end{table*}

\begin{table*}[ht]
\centering
\small
\begin{tabular}{lccccccc}
\toprule
& \multicolumn{3}{c}{Expert activation frequency $\geq$ 5\%} & \multicolumn{3}{c}{Expert activation frequency $\geq$ 1\%} & \\
\cmidrule(lr){2-4} \cmidrule(lr){5-7}
Language & Proportion & Random & Experts & Proportion & Random & Experts & Origin \\
\midrule
English & 77.4\% & 38.6 $\pm$ 3.5 & 45.2 & 91.4\% & 49.3 $\pm$ 2.3 & 51.5 & 53.7 \\
French & 76.7\% & 25.7 $\pm$ 0.7 & 26.5 & 93.3\% & 41.9 $\pm$ 0.6 & \textbf{42.9} & 42.9 \\
Spanish & 76.3\% & 16.3 $\pm$ 12.7 & 30.1 & 93.2\% & 41.5 $\pm$ 1.0 & 42.9 & 43.0 \\
Portuguese & 76.0\% & 18.6 $\pm$ 13.2 & 29.4 & 93.4\% & 35.9 $\pm$ 1.6 & \textbf{36.9} & 36.6 \\
Chinese & 76.5\% & 13.5 $\pm$ 11.5 & 30.3 & 92.9\% & 34.9 $\pm$ 0.9 & \textbf{37.2} & 36.9 \\
\midrule
Average & 76.6\% & 22.5 & 32.3 & 92.8\% & 40.7 & 42.3 & 42.6 \\
\bottomrule
\end{tabular}
\caption{Accuracy (\%) of Llama 2 7B on the X-CSQA dataset.}
\label{tab:X-CSQA-7b}
\end{table*}

\begin{table*}[!ht]
\centering
\small
\begin{tabular}{lccccccc}
\toprule
& \multicolumn{3}{c}{Expert activation frequency $\geq$ 5\%} & \multicolumn{3}{c}{Expert activation frequency $\geq$ 1\%} & \\
\cmidrule(lr){2-4} \cmidrule(lr){5-7}
Language & Proportion & Random & Experts & Proportion & Random & Experts & Origin \\
\midrule
English & 77.4\% & 2.0 $\pm$ 1.4 & 4.4 & 91.4\% & 11.5 $\pm$ 1.0 & 13.6 & 16.0 \\
French & 76.7\% & 1.5 $\pm$ 1.0 & 2.4 & 93.3\% & 8.6 $\pm$ 1.6 & 9.2 & 13.6 \\
Spanish & 76.3\% & 1.1 $\pm$ 1.2 & 3.6 & 93.2\% & 6.8 $\pm$ 0.6 & \textbf{12.4} & 10.0 \\
\midrule
Average & 76.8\% & 1.5 & 3.5 & 92.6\% & 8.9 & 11.7 & 13.2 \\
\bottomrule
\end{tabular}
\caption{Accuracy (\%) of Llama 2 7B on the MGSM dataset.}
\label{tab:MGSM-7b}
\end{table*}

\subsection{Impact of instruction tuning on activation patterns}

In the previous experiments, we primarily focused on pre-trained base models rather than instruction-tuned models to minimize interference from other factors. But we also want to understand how the multilingual activation patterns of LLMs change after instruction tuning. Specifically, does instruction tuning exhibit certain patterns in its impact on multilingual activation? To this end, we tested the changes in expert activation frequency of the instruction tuning variants of the four models mentioned above compared to the original pre-trained model using the same expert split. The results for English using Llama 2 7B and Mistral-7B-v0.3 are shown in Figure~\ref{fig:heatmap_chat}, while the results for other models and languages are provided in Appendix~\ref{sec:appendix_f}. The values in the heatmap represent the frequency of expert activations for the instruction tuning variant minus the frequency of expert activations for the pre-trained model. We found that in all models and languages, the changes in the last layer are significantly larger than those in other layers. In some experts, the activation frequencies increase (red), while in others they decrease (blue). We hope these findings can enhance our understanding of instruction tuning, aiding in more efficient and effective instruction tuning in the future.

\subsection{Can expert activation frequencies guide sparse activation and model pruning?}

After obtaining the multilingual activation patterns of LLM, we can observe differences in the activation frequencies of different experts, which reflects a certain sparsity in model activation. Building upon our prior application of the MoEfication to convert dense LLMs into fine-grained MoE architectures, during the inference process, we can reduce the amount of computation and lower FLOPs by activating only a subset of high-frequency experts, thereby significantly decreasing inference costs. Furthermore, we wonder whether it is possible to perform language-specific model pruning based on the different experts that are frequently activated by each language. Therefore, we further explored two pruning methods: (1) For each language, inference is conducted using only the high-frequency experts whose activation frequency is greater than or equal to a certain threshold. (2) Sort the experts in each layer by activation frequency, and only using the top \( n \% \) of experts based on activation frequency.

\begin{table*}[htb]
\centering
\small
\begin{tabular}{lccccccc}
\toprule
& \multicolumn{3}{c}{Expert activation frequency $\geq$ 0.5\%} & \multicolumn{3}{c}{Expert activation frequency $\geq$ 0.1\%} & \\
\cmidrule(lr){2-4} \cmidrule(lr){5-7}
Language & Proportion & Random & Experts & Proportion & Random & Experts & Origin \\
\midrule
English & 90.1\% & 7.7 $\pm$ 5.8 & 70.3 & 93.3\% & 7.9 $\pm$ 11.1 & \textbf{76.7} & 76.0 \\
French & 87.7\% & 3.3 $\pm$ 2.4 & 66.2 & 93.2\% & 11.9 $\pm$ 16.8 & \textbf{67.0} & 66.9 \\
Spanish & 87.5\% & 8.2 $\pm$ 11.2 & 67.1 & 93.1\% & 15.8 $\pm$ 21.3 & \textbf{69.4} & 68.2 \\
Portuguese & 87.5\% & 25.2 $\pm$ 18.5 & 65.9 & 93.2\% & 46.0 $\pm$ 7.5 & \textbf{68.7} & 66.5 \\
Urdu & 85.7\% & 0.3 $\pm$ 0.4 & 55.0 & 91.7\% & 12.9 $\pm$ 12.4 & 57.6 & 58.4 \\
Hindi & 86.0\% & 3.4 $\pm$ 2.4 & 57.0 & 90.5\% & 14.0 $\pm$ 12.6 & 57.2 & 60.0 \\
Arabic & 84.8\% & 0.0 $\pm$ 0.0 & 63.0 & 91.7\% & 11.8 $\pm$ 3.6 & 63.2 & 64.3 \\
Chinese & 87.0\% & 3.3 $\pm$ 4.2 & 59.9 & 92.8\% & 22.2 $\pm$ 15.7 & \textbf{64.1} & 63.3 \\
\midrule
Average & 87.0\% & 6.4 & 63.1 & 92.4\% & 17.8 & \textbf{65.5} & 65.5 \\
\bottomrule
\end{tabular}
\vspace{-1mm}
\caption{Accuracy (\%) of Llama 3 70B on the X-CSQA dataset.}
\label{tab:X-CSQA-70b}
\vspace{-1mm}
\end{table*}

\begin{table*}[htb]
\centering
\small
\begin{tabular}{lccccccc}
\toprule
& \multicolumn{3}{c}{Expert activation frequency $\geq$ 0.5\%} & \multicolumn{3}{c}{Expert activation frequency $\geq$ 0.1\%} & \\
\cmidrule(lr){2-4} \cmidrule(lr){5-7}
Language & Proportion & Random & Experts & Proportion & Random & Experts & Origin \\
\midrule
English & 90.1\% & 1.6 $\pm$ 1.4 & 62.0 & 93.3\% & 8.3 $\pm$ 5.5 & 79.6 & 81.6 \\
French & 87.7\% & 2.7 $\pm$ 2.7 & 54.8 & 93.2\% & 5.9 $\pm$ 4.2 & 62.0 & 63.2 \\
Spanish & 87.5\% & 1.5 $\pm$ 1.0 & 66.4 & 93.1\% & 8.1 $\pm$ 7.9 & 75.2 & 75.6 \\
Bengali & 87.9\% & 0.0 $\pm$ 0.0 & 30.0 & 91.8\% & 4.1 $\pm$ 3.9 & 35.2 & 36.8 \\
\midrule
Average & 88.3\% & 1.5 & 53.3 & 92.9\% & 6.6 & 63.0 & 64.3 \\
\bottomrule
\end{tabular}
\vspace{-1mm}
\caption{Accuracy (\%) of Llama 3 70B on the MGSM dataset.}
\label{tab:MGSM-70b}
\vspace{-4mm}
\end{table*}

\begin{figure*}[t]
  \centering
  \begin{subfigure}[b]{0.4\textwidth}
    \includegraphics[width=\linewidth]{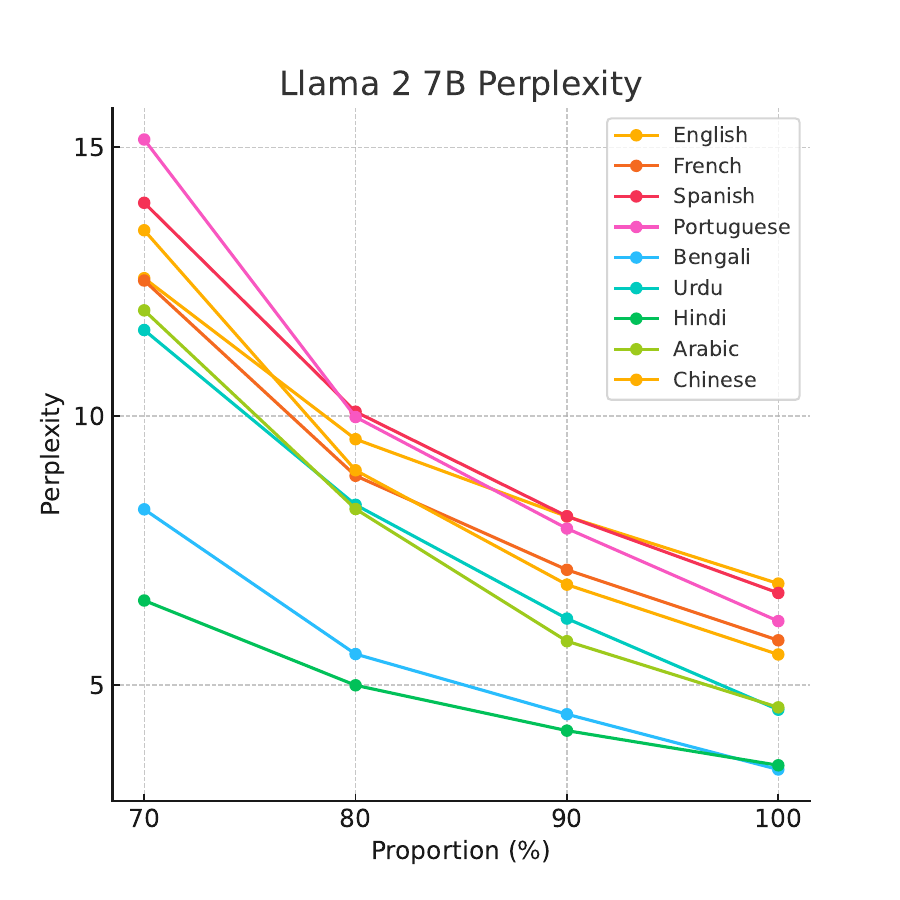}
  \end{subfigure}
  \begin{subfigure}[b]{0.4\textwidth}
    \includegraphics[width=\linewidth]{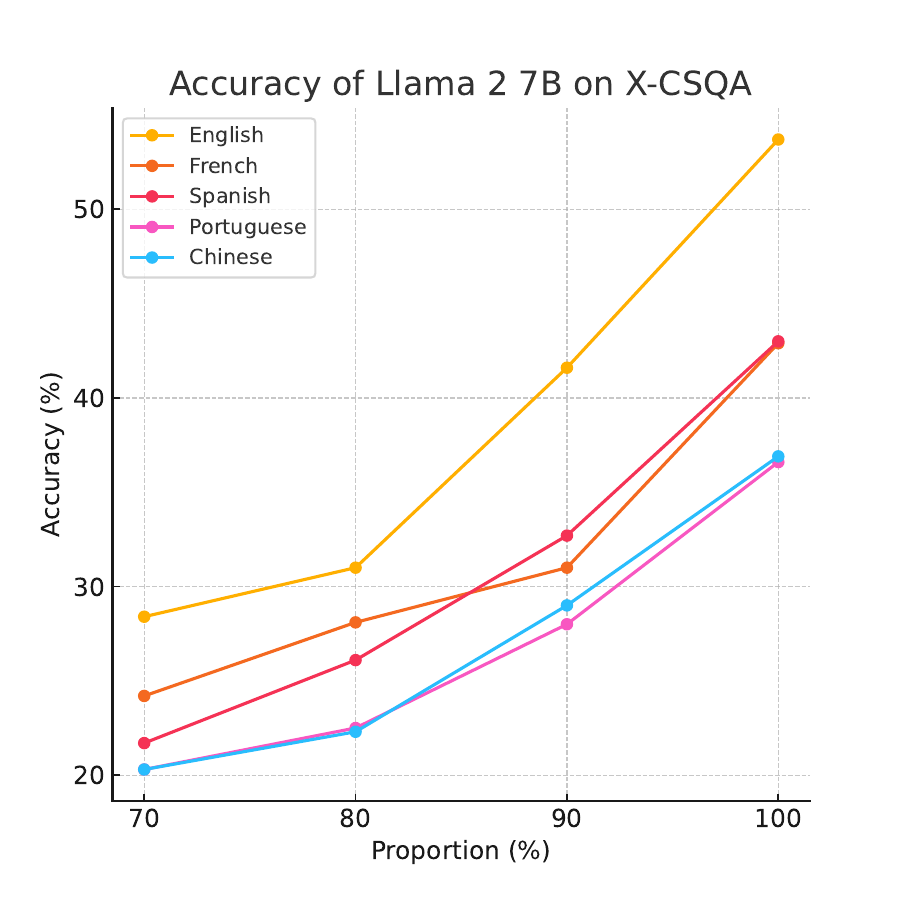}
  \end{subfigure}
  \vspace{-1mm}
  \caption{Results of Llama 2 7B pruning based on frequency sorting.}
  \label{fig:llam2_sort}
  \vspace{-1mm}
\end{figure*}

\paragraph{Evaluation.} We evaluate the performance changes of the models before and after pruning using perplexity (PPL) and accuracy on two datasets: X-CSQA~\citep{lin-etal-2021-xcsr,Talmor2019commonsenseqaaq} (commonsense question answering task) and MGSM~\citep{cobbe2021training,shilanguage} (grade-school math problems). For the perplexity test, we use 1,000 data samples from the ROOTS corpus that are different from the samples used in the expert activation pattern tests, with a maximum input length of 200 tokens per sample. Since the X-CSQA dataset does not publicly provide test set labels, we use 1,000 instances from the dev set for testing, with each question having five options. For each dataset, we selected several languages that overlap with those studied in our research and for which the original model can produce reasonable outputs.\footnote{For instance, the Llama 2 7B and Llama 2-Chat 7B models are almost incapable of correctly answering questions in Urdu, Hindi, and Arabic, so we did not conduct experiments on these languages.}

\subsubsection{Pruning based on frequency thresholds.} To comprehensively study the impact of different model variants and sizes, we conducted experiments on Llama 2 7B, Llama 2-Chat 7B, and Llama 3 70B. Tables~\ref{tab:ppl-7b} to~\ref{tab:MGSM-70b} present partial results, with additional results provided in Appendix~\ref{sec:appendix_g}. During inference, for each language, we only use the experts whose activation frequency for that language is greater than or equal to a specified threshold, excluding the parameters of other experts. For Llama 2 7B and Llama 2-Chat 7B, we experimented with thresholds of 5\% and 1\%, which reduced the FFN layer parameters by approximately 25\% and 10\%, respectively. For Llama 3 70B, due to the generally low activation frequencies of experts, we experimented with thresholds of 0.5\% and 0.1\%, which reduced the FFN layer parameters by approximately 13\% and 8\%, respectively. Based on the model structure calculations, reducing the FFN layer parameters by 25\% can decrease total inference FLOPs by approximately 18\%, while a 13\% reduction can lower FLOPs by approximately 9\%.

To demonstrate the effectiveness of our method, we compare it with random expert selection. Specifically, we conduct experiments using the same proportion of randomly selected experts at each layer. In each table, the Proportion column represents the ratio of the experts used to the total number of experts. The Random column shows the results of using only randomly selected experts. We experimented with three different random seeds and reported the means and standard deviations. The Experts column displays the results of using only experts whose activation frequency is greater than or equal to the specified threshold. The Origin column presents the results of using the original model without pruning. We bold the results where the pruned models perform better than or equal to the original models.

We can observe that the model performance significantly declines when only using randomly selected experts. Additionally, the standard deviation of results from randomly selected experts is quite large, especially when using a higher pruning ratio. This indicates that the selection of different experts greatly affects the model performance. Pruning based on our method, which considers whether the expert activation frequency exceeds a threshold, can maintain the model performance as much as possible, far surpassing the random selection of experts. Surprisingly, when using our method with a lower threshold for pruning, all three models even outperformed the original models in some languages. This further demonstrates the correctness of the expert activation frequency differences identified by our method, providing a new feasible path for model pruning.

\subsubsection{Pruning based on frequency sorting.} We also experimented with sorting the experts in each layer by activation frequency and using only the top 70\%, 80\%, and 90\% of the experts based on activation frequency. The results for Llama 2 7B are shown in Figure~\ref{fig:llam2_sort}, and those for Llama 3 70B in Figure~\ref{fig:llam3_sort}. We found that the model performance deteriorates rapidly with increasing pruning ratios. With similar pruning ratios, the perplexity is comparable to pruning based on frequency thresholds, but the performance on the X-CSQA dataset is significantly worse than pruning based on frequency thresholds. The effect of equal proportional pruning at each layer is inferior to that of unequal pruning, reflecting the inter-layer differences in activation levels of LLMs. This validates our earlier finding that the sparsity of expert activation levels varies across different layers. Therefore, we recommend configuring different pruning rates for different layers based on the differences in activation levels.

\section{Conclusion}

In this study, we investigated the multilingual activation patterns in various LLMs from the perspective of MoE models. We also explored their connections with language families and instruction tuning, as well as the potential for guiding sparse activation and model pruning. These findings can assist us in better developing and utilizing the multilingual capabilities of LLMs. We hope these findings inspire new research in related fields.

\section{Limitations}

Despite achieving some meaningful conclusions in our research, there are still some limitations.

\paragraph{The limitations of pruning ratio.} Section~\ref{subsec:41} and previous studies~\citep{mirzadeh2023relu} have demonstrated that models like Llama 2 exhibit low activation sparsity in most layers, thus the pruning ratio we use is not very high. However, we believe that we offer new perspectives for applications such as model pruning, providing a foundation for further exploration in the future.

\paragraph{The limitations of interpretability by using a simplified model.} Some work~\citep{friedman2023interpretability} on mechanistic interpretability may present the concern that ``even if the simplified representations can accurately approximate the full model on the training set, they may fail to accurately capture the model's behavior out of distribution.'' Thus, in order to verify the effectiveness of the high-frequency activated experts identified by our method for each language, we test the capability of the model pruned based on expert activation frequency and compare it with pruning done at the same ratio randomly. When analyzing the expert activation frequency of LLMs across different languages, we used the ROOTS corpus derived from Wikipedia. However, to evaluate the performance of the pruned models, we employed three distinct tasks for a comprehensive evaluation: the perplexity on another portion of the ROOTS corpus, the accuracy on the X-CSQA dataset and the MGSM dataset. We can see that the tests on the X-CSQA dataset and the MGSM dataset essentially evaluate the model's out-of-distribution capability. The results from these two datasets indicate that our models pruned based on expert activation frequency maintain performance comparable to the original models, and even exceed the original models in some languages. To a certain extent, this provides evidence that our approach enables simplified models to capture the out-of-distribution behaviors of the original models, thereby mitigating concerns.

In the future, we plan to extend our experiments to a broader array of models and languages. In addition, we will explore how to further leverage these insights to utilize and enhance the multilingual capabilities of LLMs.

\section{Ethical Considerations}

\paragraph{The use of AI assistants.} We employed ChatGPT to assist us in polishing our paper and writing code.

\section*{Acknowledgements}

This research was partially supported by the Key R\&D Program of Zhejiang under grant No.~2024SSYS0026.

\bibliography{anthology,custom}

\clearpage
\appendix

\section{The FFN Structure}
\label{sec:appendix_a}

\begin{figure}[ht]
  \centering
  \includegraphics[width=\linewidth]{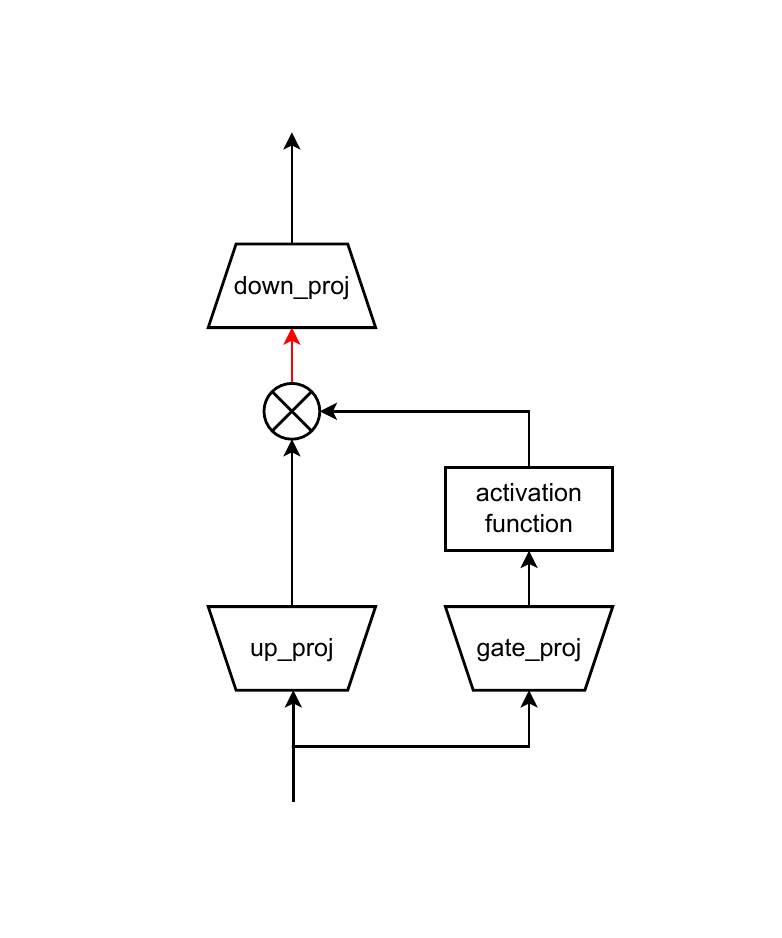}
  \caption{Schematic diagram of the FFN structure in Llama/Mistral model series.}
  \label{fig:FFN}
\end{figure}

When calculating the scores for each expert, the activation values that we use are the hidden representations before the down-projection layer, as indicated by the red section in Figure~\ref{fig:FFN}.

\section{Data Sources}
\label{sec:appendix_b}

We obtained the ROOTS corpus data from Hugging Face, with their website URL presented in Table~\ref{tab:url}.

\begin{table}[ht]
\centering
\begin{tabular}{l p{4.8cm}}
\toprule
Language & URL \\
\midrule
English & \url{https://huggingface.co/datasets/bigscience-data/roots_en_wikipedia} \\
French & \url{https://huggingface.co/datasets/bigscience-data/roots_fr_wikipedia} \\
Spanish & \url{https://huggingface.co/datasets/bigscience-data/roots_es_wikipedia} \\
Portuguese & \url{https://huggingface.co/datasets/bigscience-data/roots_pt_wikipedia} \\
Bengali & \url{https://huggingface.co/datasets/bigscience-data/roots_indic-bn_wikipedia} \\
Urdu & \url{https://huggingface.co/datasets/bigscience-data/roots_indic-ur_wikipedia} \\
Hindi & \url{https://huggingface.co/datasets/bigscience-data/roots_indic-hi_wikipedia} \\
Arabic & \url{https://huggingface.co/datasets/bigscience-data/roots_ar_wikipedia} \\
Chinese & \url{https://huggingface.co/datasets/bigscience-data/roots_zh-cn_wikipedia} \\
\bottomrule
\end{tabular}
\caption{Wikipedia dataset URLs for various languages.}
\label{tab:url}
\end{table}

\section{Expert Activation Pattern Heatmaps}
\label{sec:appendix_c}

\begin{figure*}[ht]
  \centering
  \begin{minipage}{\linewidth}
    \centering
    \includegraphics[width=\linewidth]{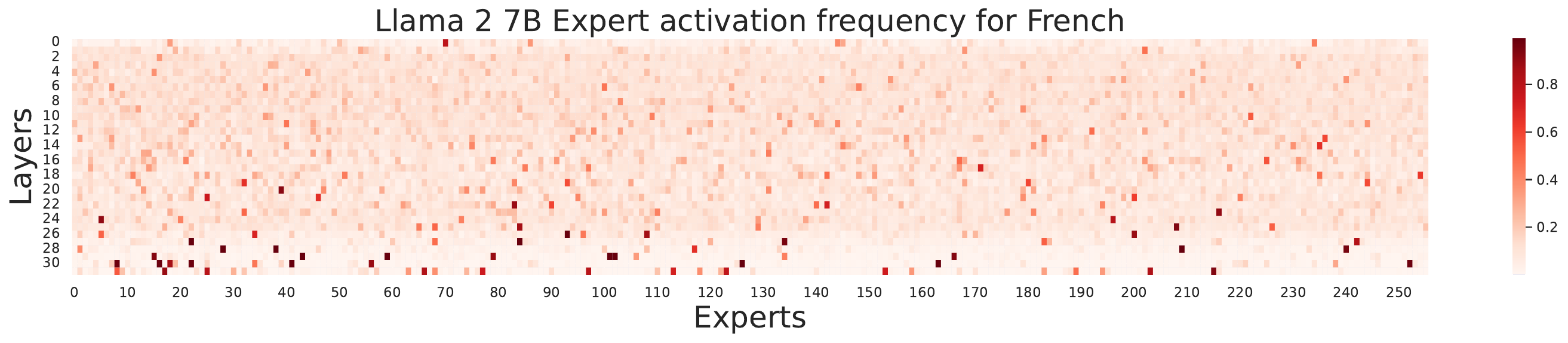}
    \vspace{-6mm}
  \end{minipage}

  \begin{minipage}{\linewidth}
    \centering
    \includegraphics[width=\linewidth]{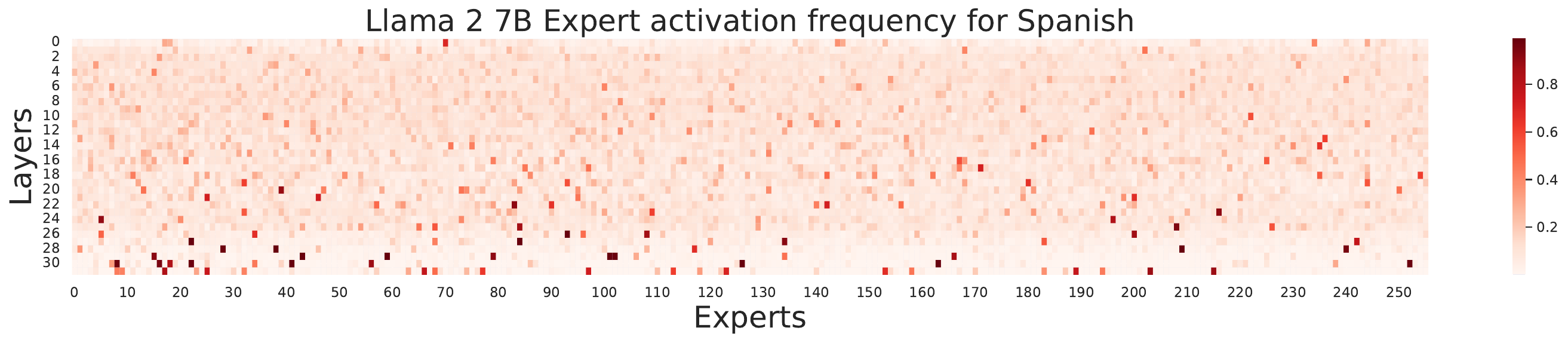}
    \vspace{-6mm}
  \end{minipage}

  \begin{minipage}{\linewidth}
    \centering
    \includegraphics[width=\linewidth]{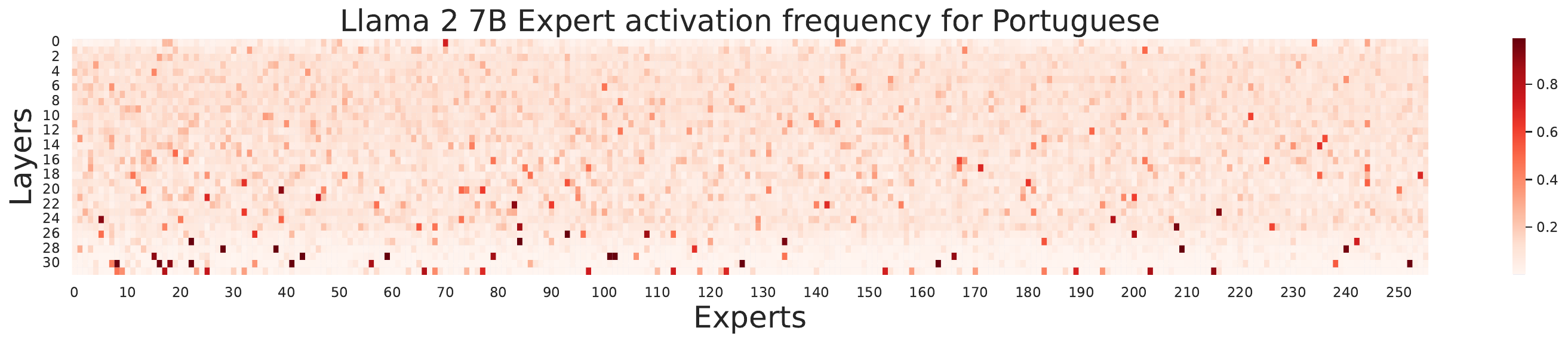}
    \vspace{-6mm}
  \end{minipage}

  \begin{minipage}{\linewidth}
    \centering
    \includegraphics[width=\linewidth]{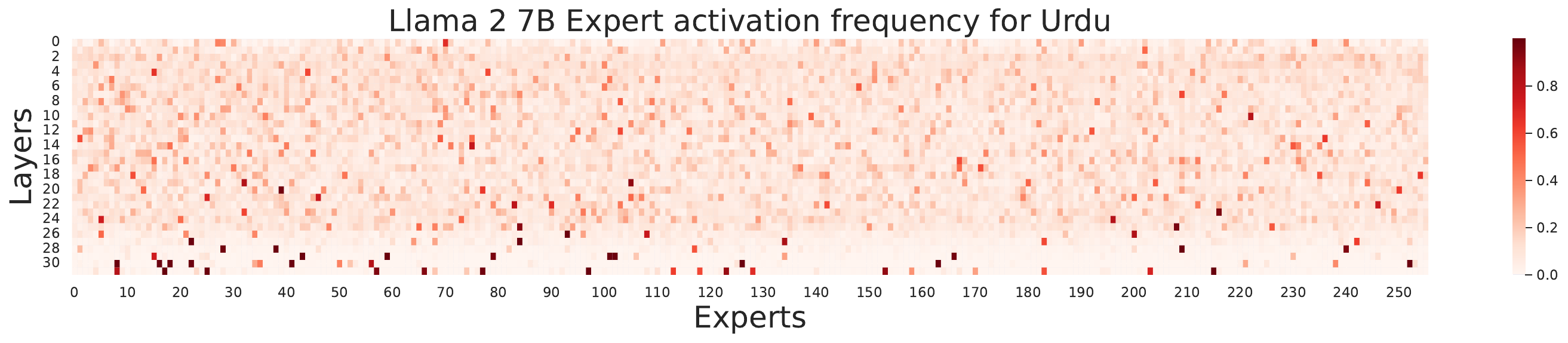}
    \vspace{-6mm}
  \end{minipage}

  \begin{minipage}{\linewidth}
    \centering
    \includegraphics[width=\linewidth]{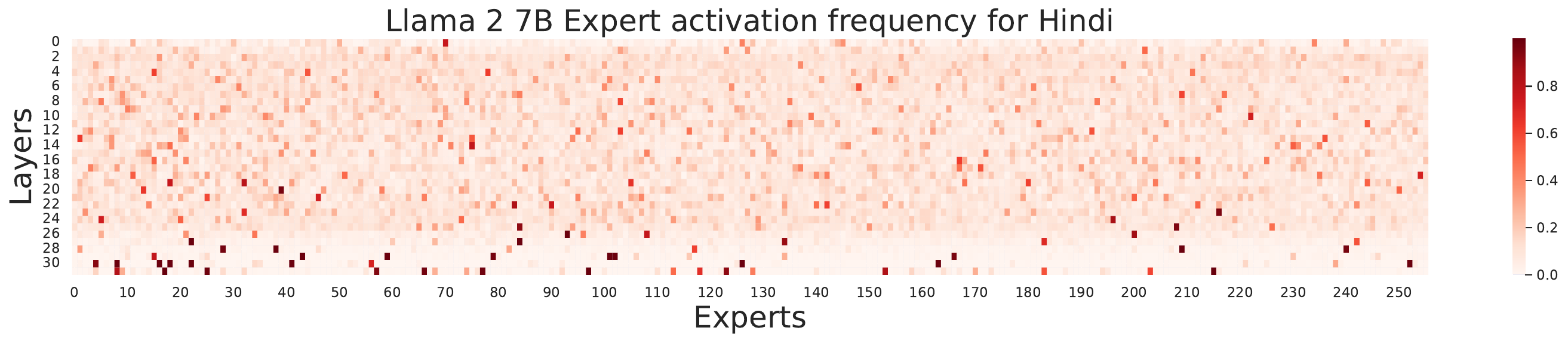}
    \vspace{-6mm}
  \end{minipage}

  \begin{minipage}{\linewidth}
    \centering
    \includegraphics[width=\linewidth]{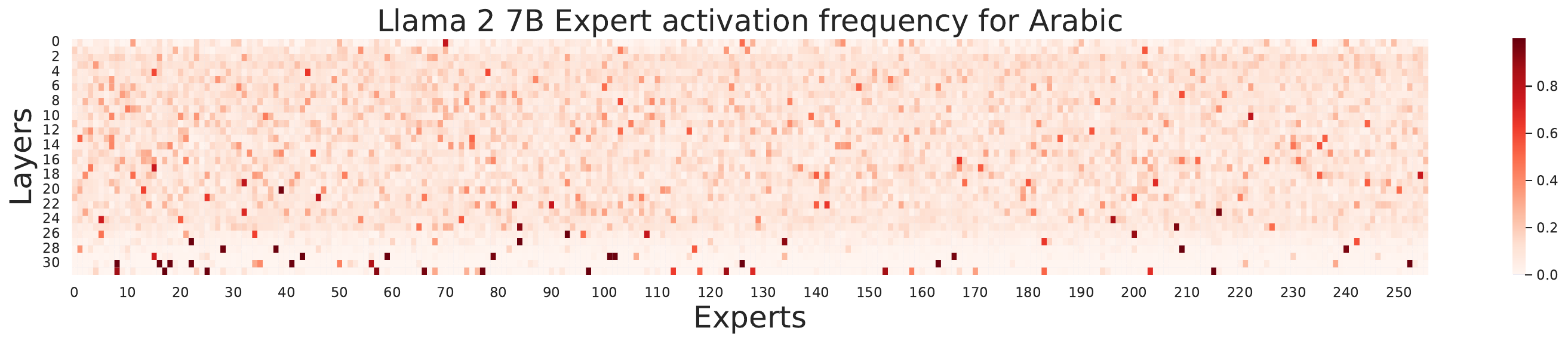}
    \vspace{-6mm}
  \end{minipage}

  \begin{minipage}{\linewidth}
    \centering
    \includegraphics[width=\linewidth]{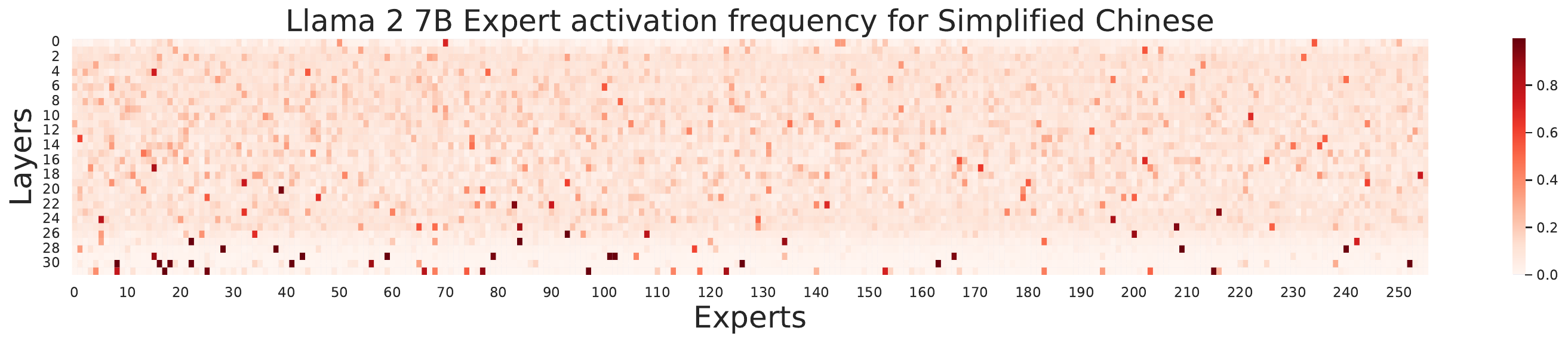}
    \vspace{-6mm}
  \end{minipage}
  
  \caption{Heatmaps of activation patterns across languages for Llama 2 7B.}
  \label{fig:heatmap_llama2-7b}
  \vspace{-3mm}
\end{figure*}

\begin{figure*}[t]
  \centering  
  \begin{minipage}{\linewidth}
    \centering
    \includegraphics[width=\linewidth]{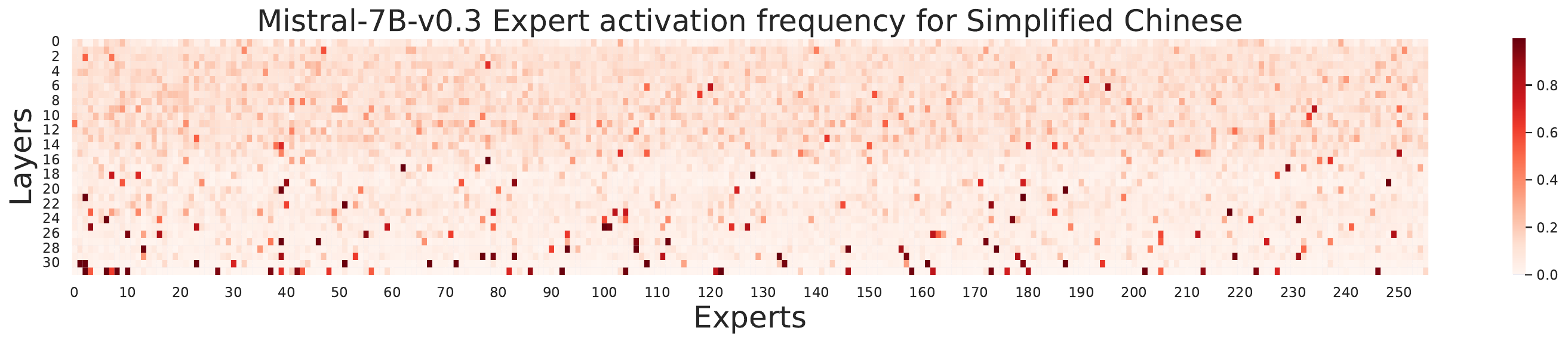}
    \vspace{-6mm}
  \end{minipage}

  \caption{Heatmaps of activation patterns across languages for Mistral-7B-v0.3.}
  \label{fig:heatmap_mistral-7b}
\end{figure*}

\begin{figure*}[t]
  \centering
  \begin{minipage}{\linewidth}
    \centering
    \includegraphics[width=\linewidth]{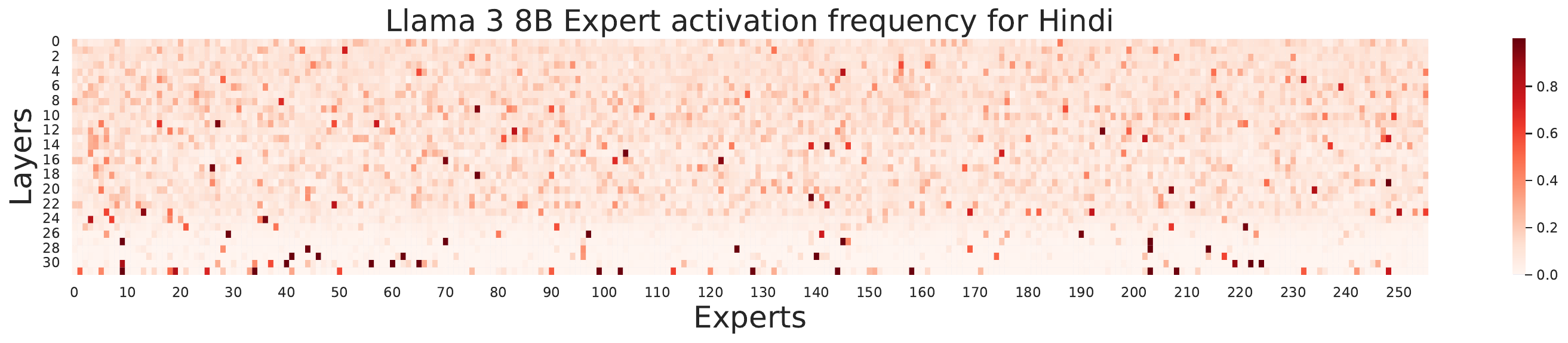}
    \vspace{-6mm}
  \end{minipage}

  \begin{minipage}{\linewidth}
    \centering
    \includegraphics[width=\linewidth]{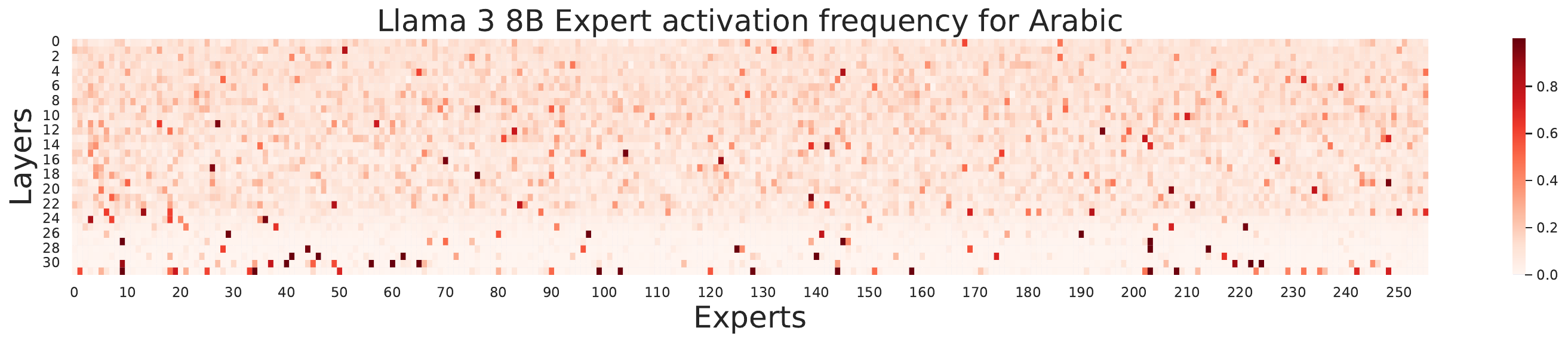}
    \vspace{-6mm}
  \end{minipage}

  \begin{minipage}{\linewidth}
    \centering
    \includegraphics[width=\linewidth]{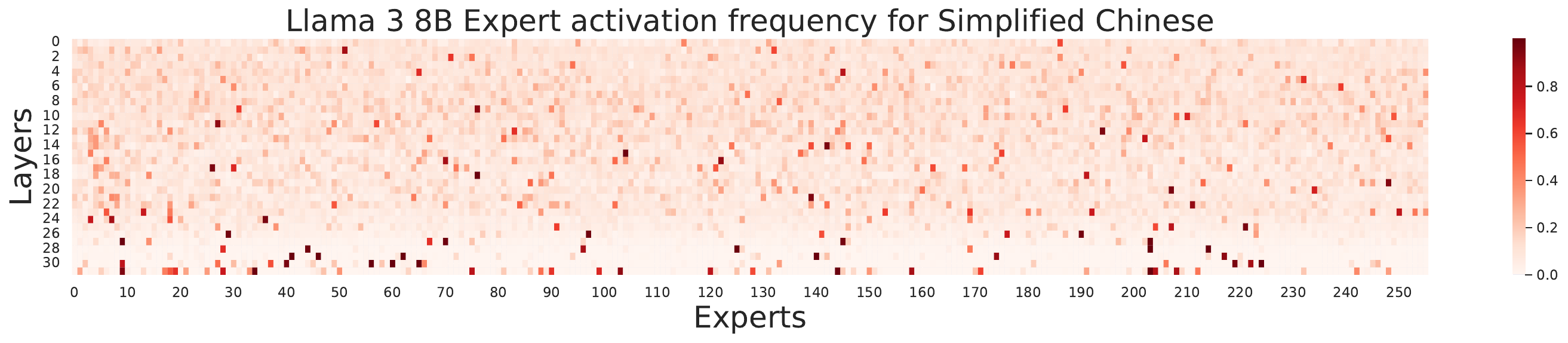}
    \vspace{-6mm}
  \end{minipage}

  \caption{Heatmaps of activation patterns across languages for Llama 3 8B.}
  \label{fig:heatmap_llama3-8b}
\end{figure*}

\begin{figure*}[t]
  \begin{minipage}{\linewidth}
    \centering
    \includegraphics[width=\linewidth]{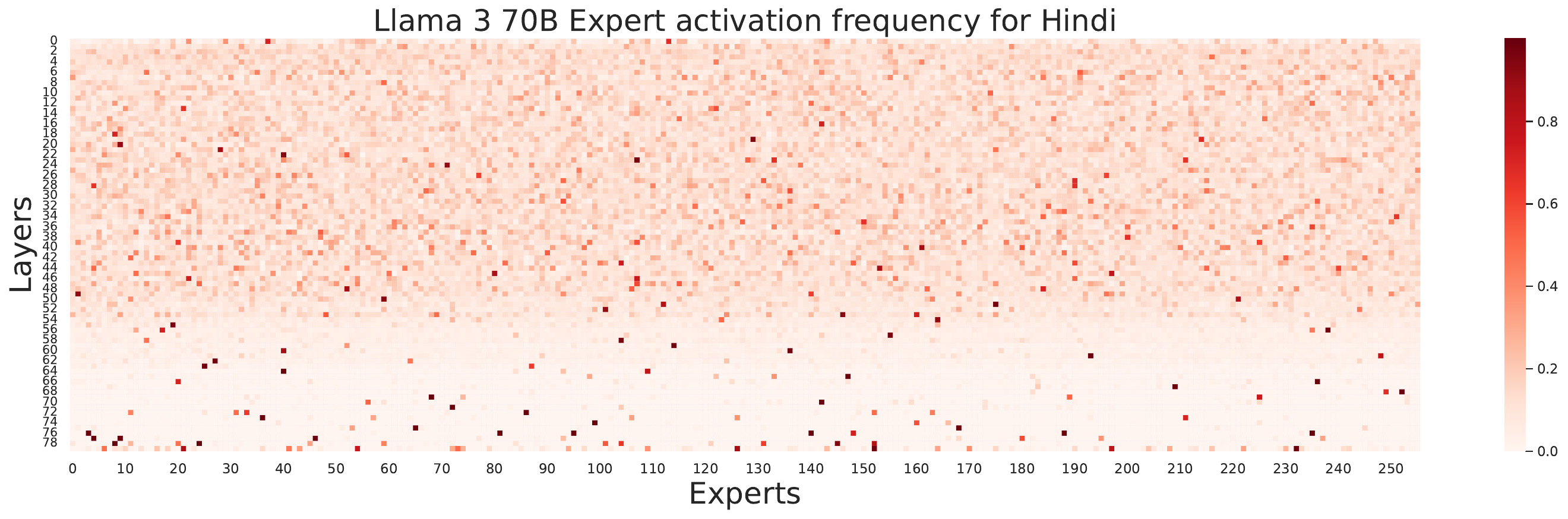}
    \vspace{-6mm}
  \end{minipage}

  \centering
  \begin{minipage}{\linewidth}
    \centering
    \includegraphics[width=\linewidth]{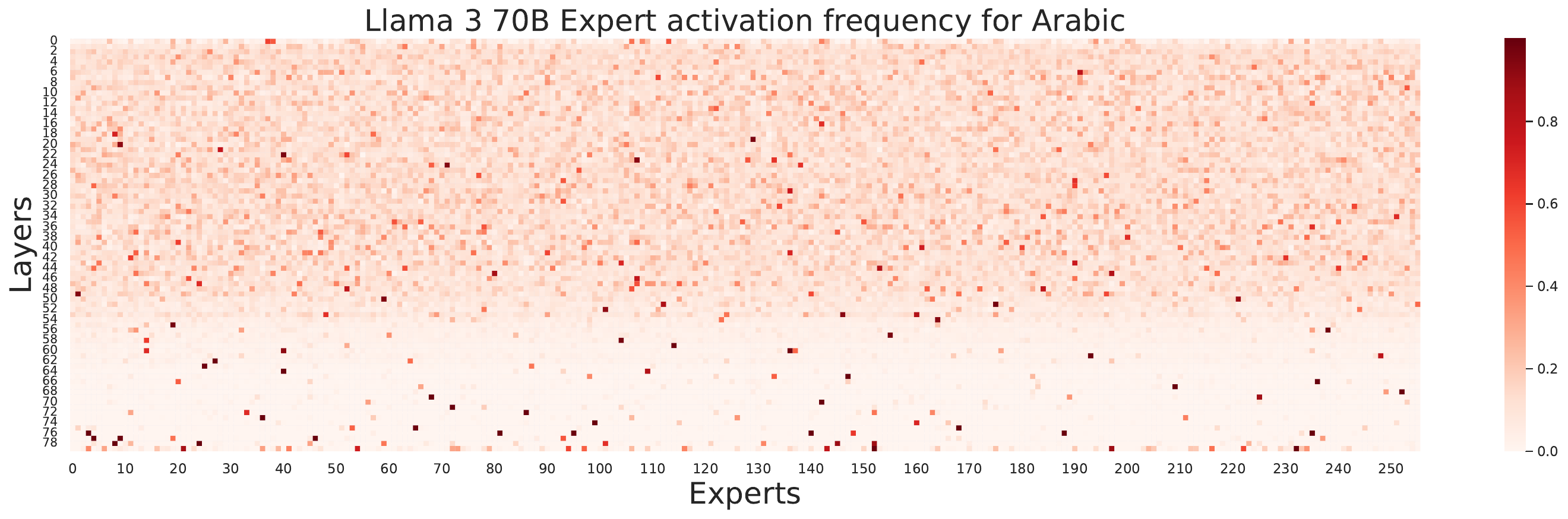}
    \vspace{-6mm}
  \end{minipage}

  \caption{Heatmaps of activation patterns across languages for Llama 3 70B.}
  \label{fig:heatmap_llama3-70b}
\end{figure*}

In Figures~\ref{fig:heatmap_llama2-7b} to~\ref{fig:heatmap_llama3-70b}, we present the activation pattern heatmaps for the remaining models and languages.

\section{Similarity between Different Activation Matrices}
\label{sec:appendix_d}

\begin{figure*}[ht]
  \centering
  \begin{subfigure}[b]{0.325\textwidth}
    \includegraphics[width=\linewidth]{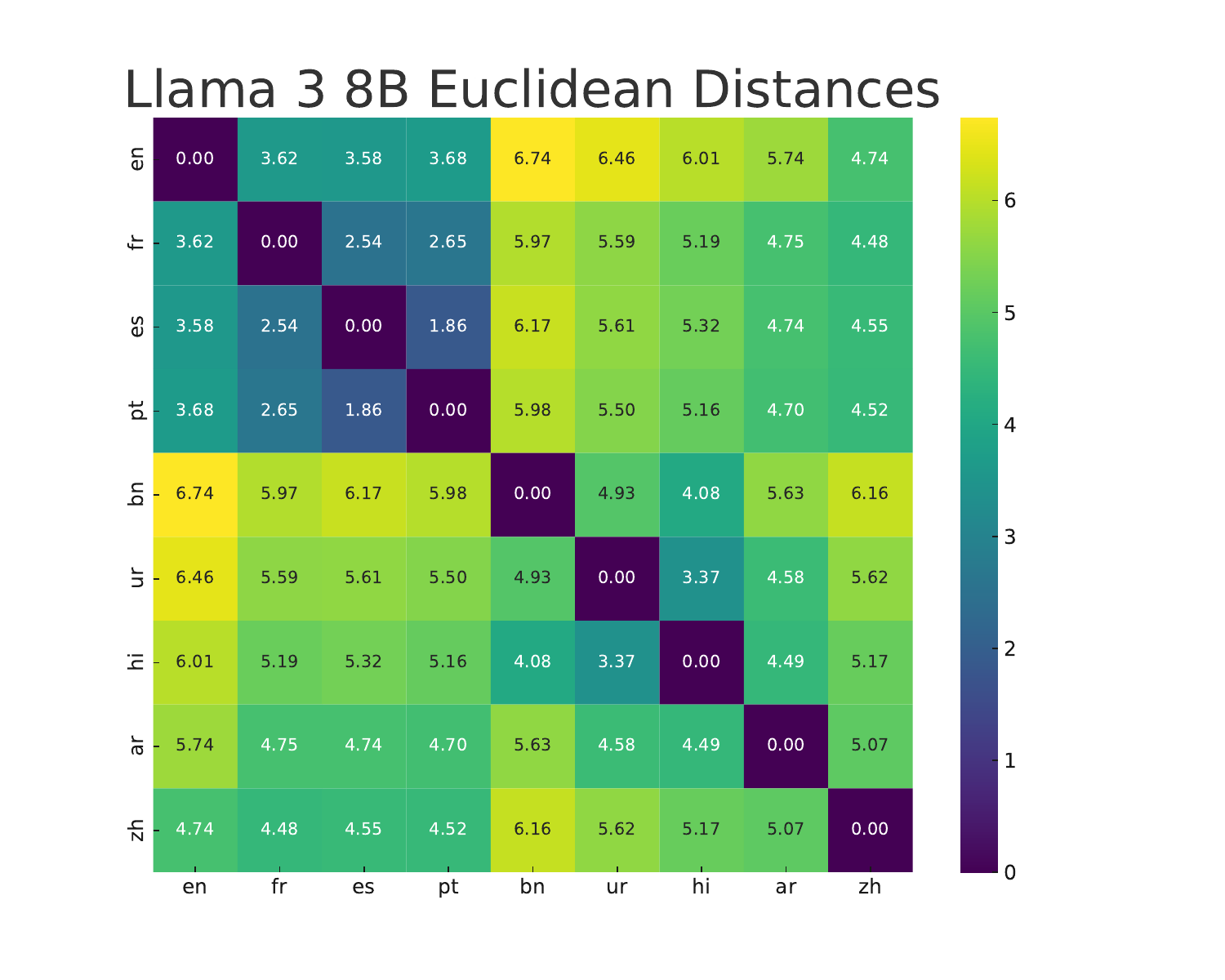}
  \end{subfigure}
  \vspace{-2mm}
  \begin{subfigure}[b]{0.325\textwidth}
    \includegraphics[width=\linewidth]{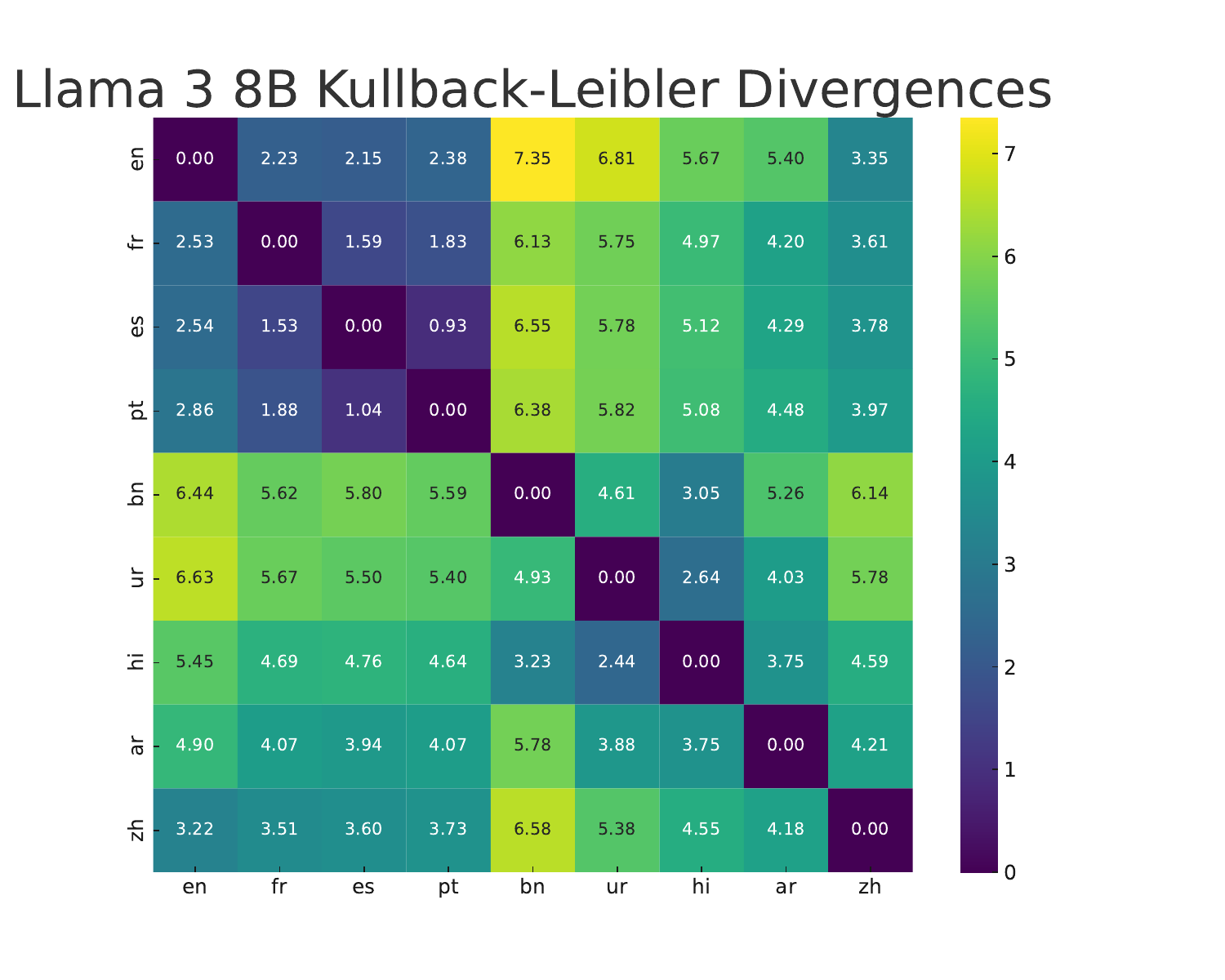}
  \end{subfigure}
  \begin{subfigure}[b]{0.325\textwidth}
    \includegraphics[width=\linewidth]{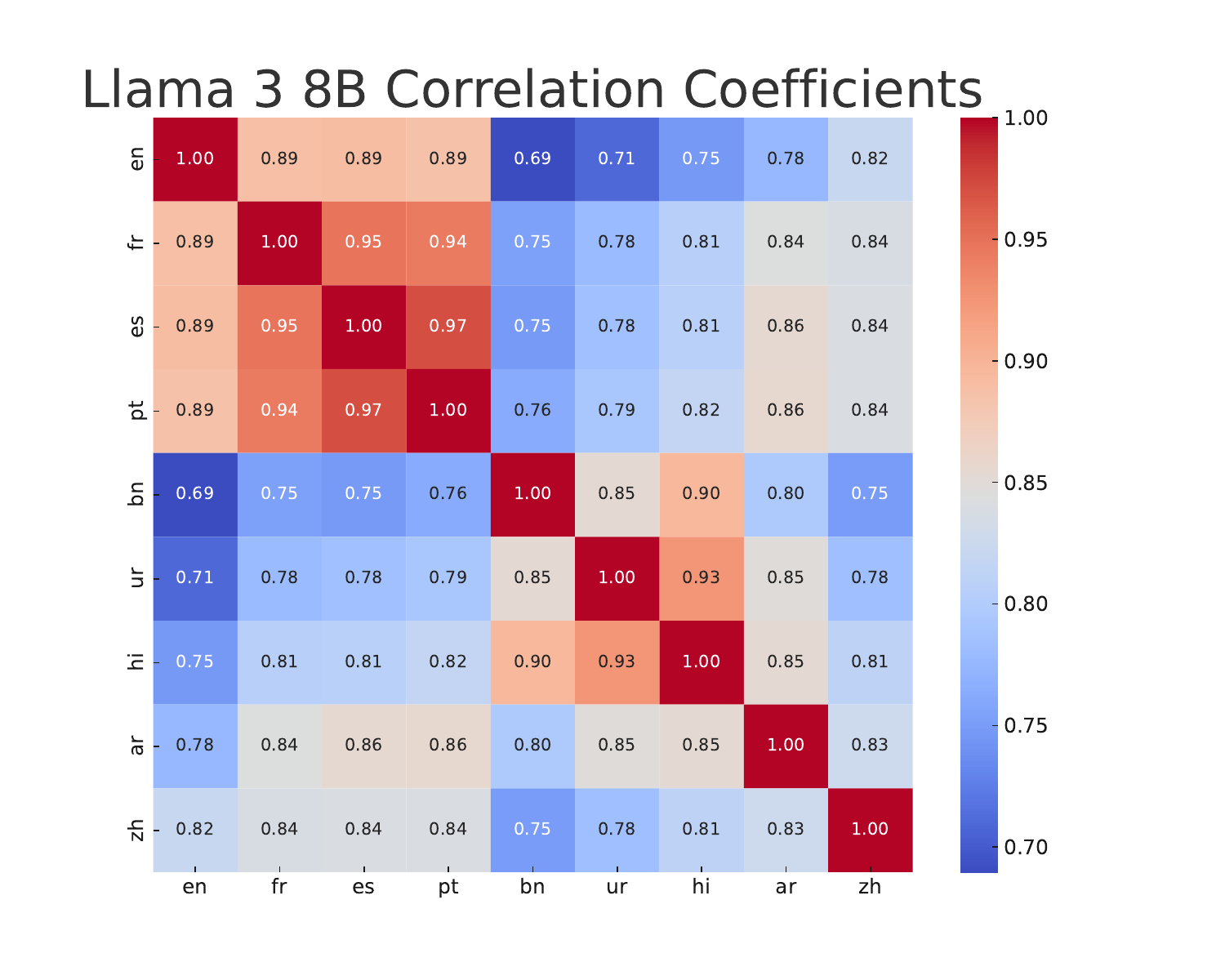}
  \end{subfigure}
  \begin{subfigure}[b]{0.325\textwidth}
    \includegraphics[width=\linewidth]{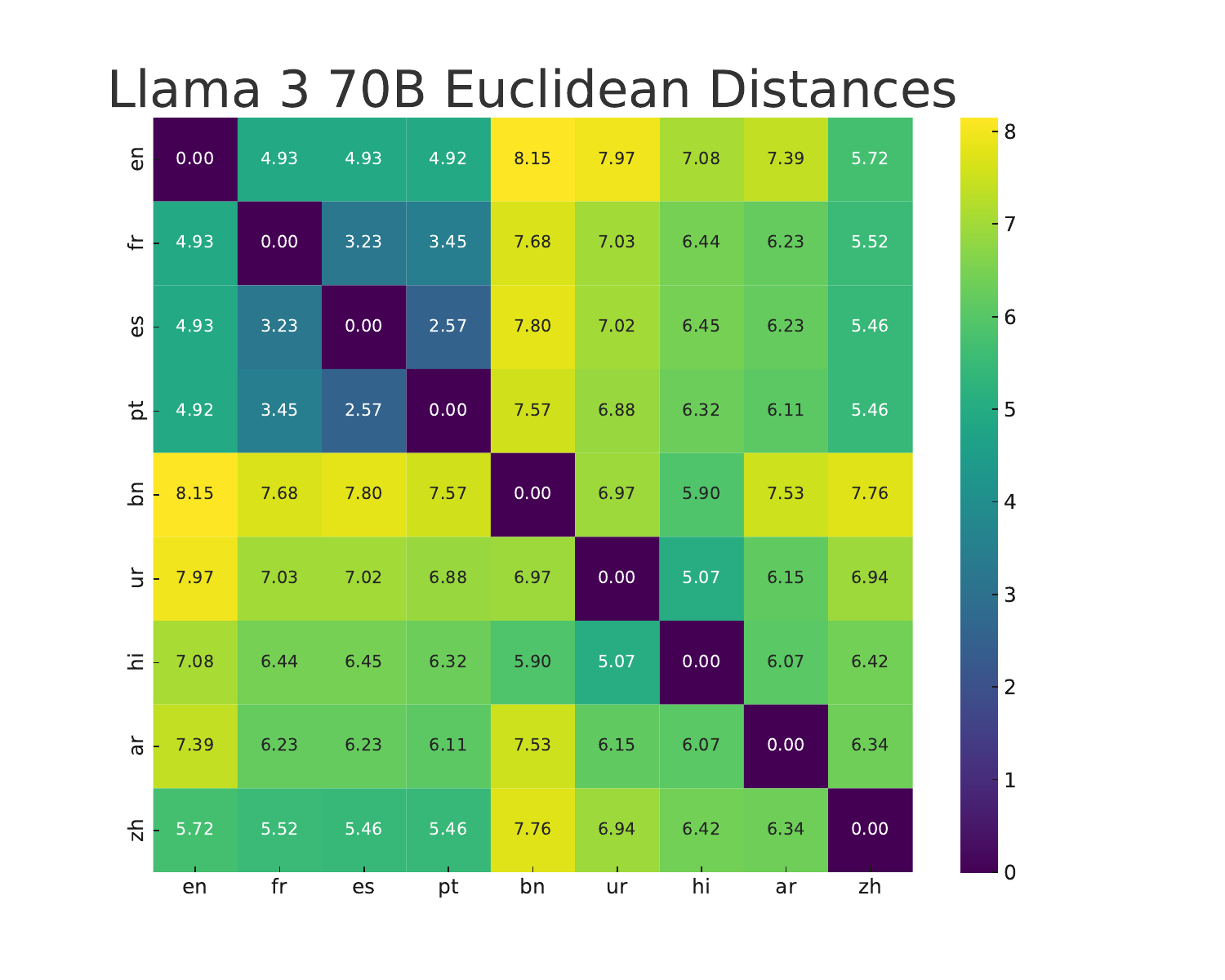}
  \end{subfigure}
  \begin{subfigure}[b]{0.325\textwidth}
    \includegraphics[width=\linewidth]{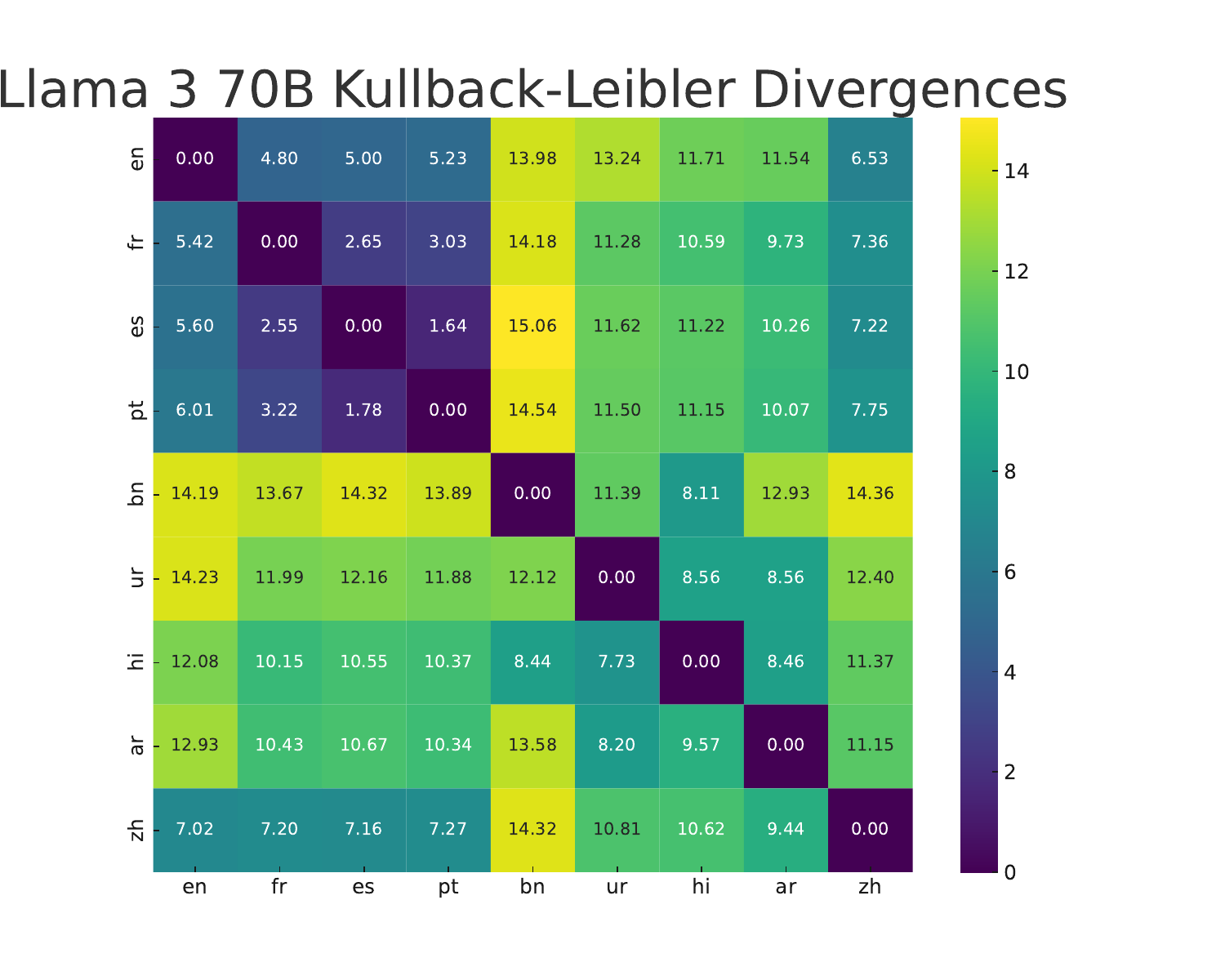}
  \end{subfigure}
  \begin{subfigure}[b]{0.325\textwidth}
    \includegraphics[width=\linewidth]{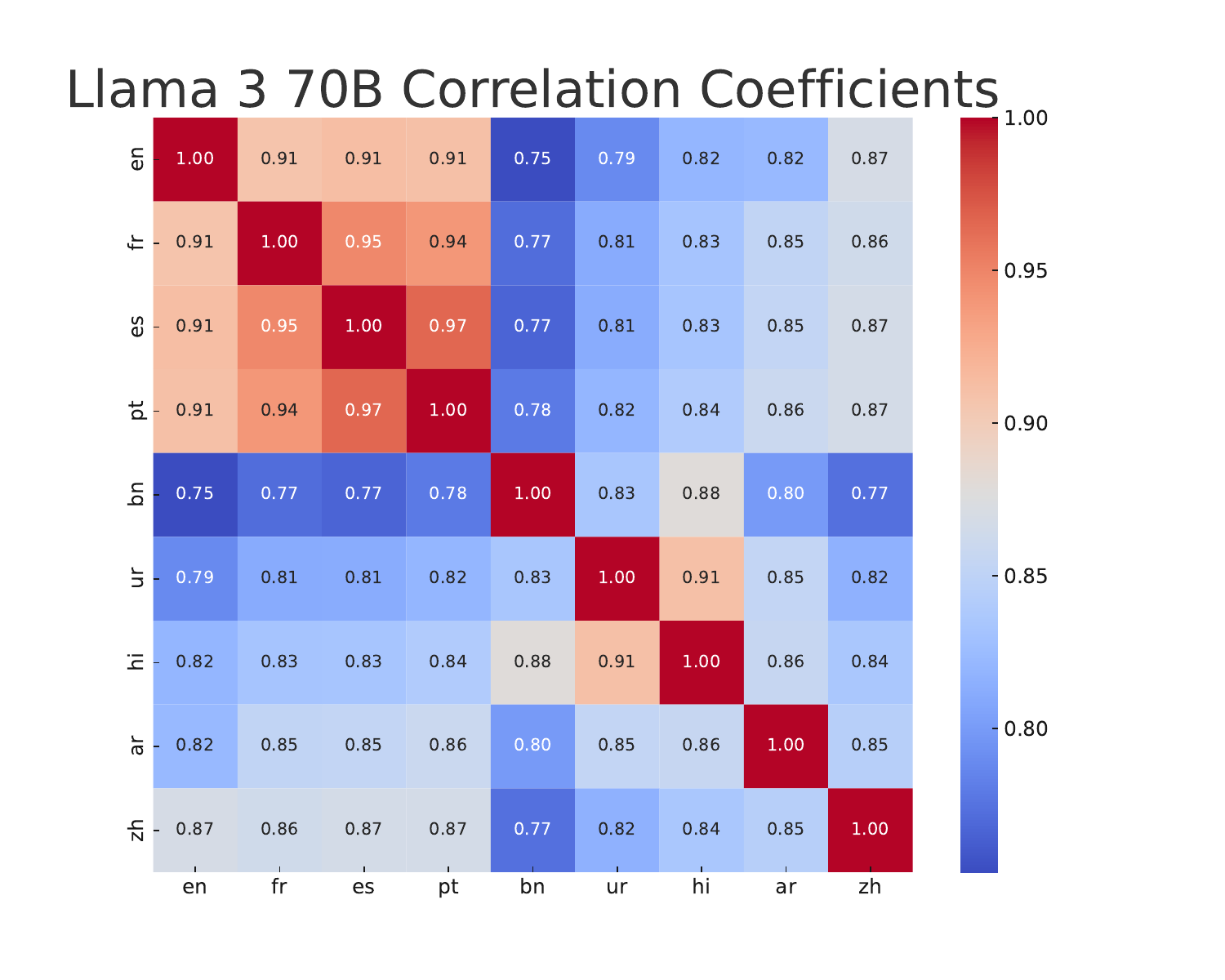}
  \end{subfigure}
  \vspace{-4mm}
  \caption{Heatmaps of similarity between activation pattern matrices for different languages in Llama 3 8B and Llama 3 70B. Each value in the Euclidean distance heatmaps represents the square root of the sum of the squares of the differences between corresponding elements of two matrices. Each value in the Kullback-Leibler (KL) divergence heatmaps represents the cumulative sum of the KL divergences computed row-wise between two matrices. Each value in the correlation coefficients heatmaps represents the mean of the Pearson correlation coefficients calculated row-wise between two matrices. The smaller the Euclidean distance/KL divergence, the more similar the two matrices are. The larger the correlation coefficient, the more similar the two matrices are.}
  \label{fig:combined_heatmaps-2}
  \vspace{-4mm}
\end{figure*}

Figure~\ref{fig:combined_heatmaps-2} presents the heatmaps of similarity between activation pattern matrices for different languages in Llama 3 8B and Llama 3 70B. It can be seen that these results are generally consistent with the patterns observed in Figure~\ref{fig:combined_heatmaps}.

\section{Multilingual shared expert Heatmaps}
\label{sec:appendix_e}

\begin{figure*}[ht]
  \centering
  \begin{minipage}{\linewidth}
    \centering
    \includegraphics[width=\linewidth]{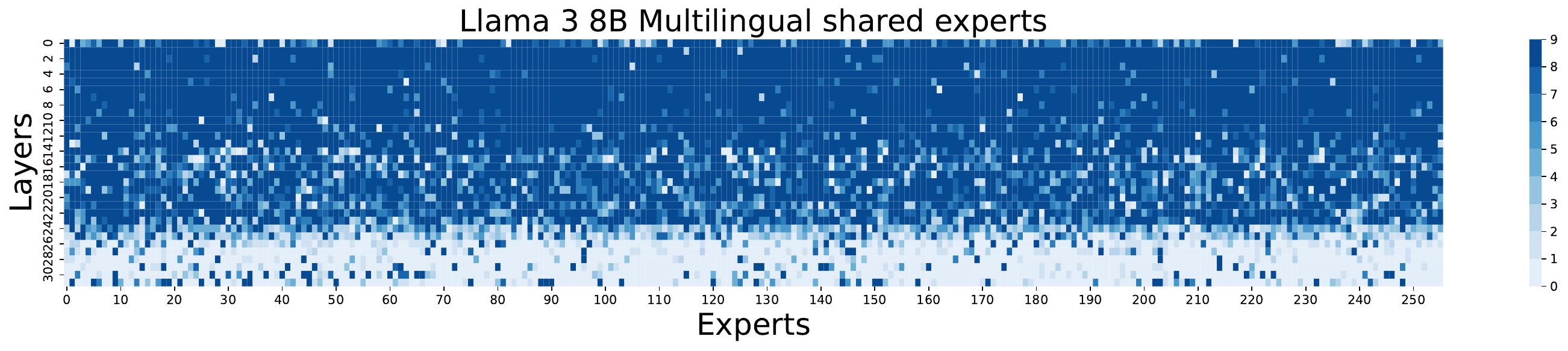}
  \end{minipage}

  \begin{minipage}{\linewidth}
    \centering
    \includegraphics[width=\linewidth]{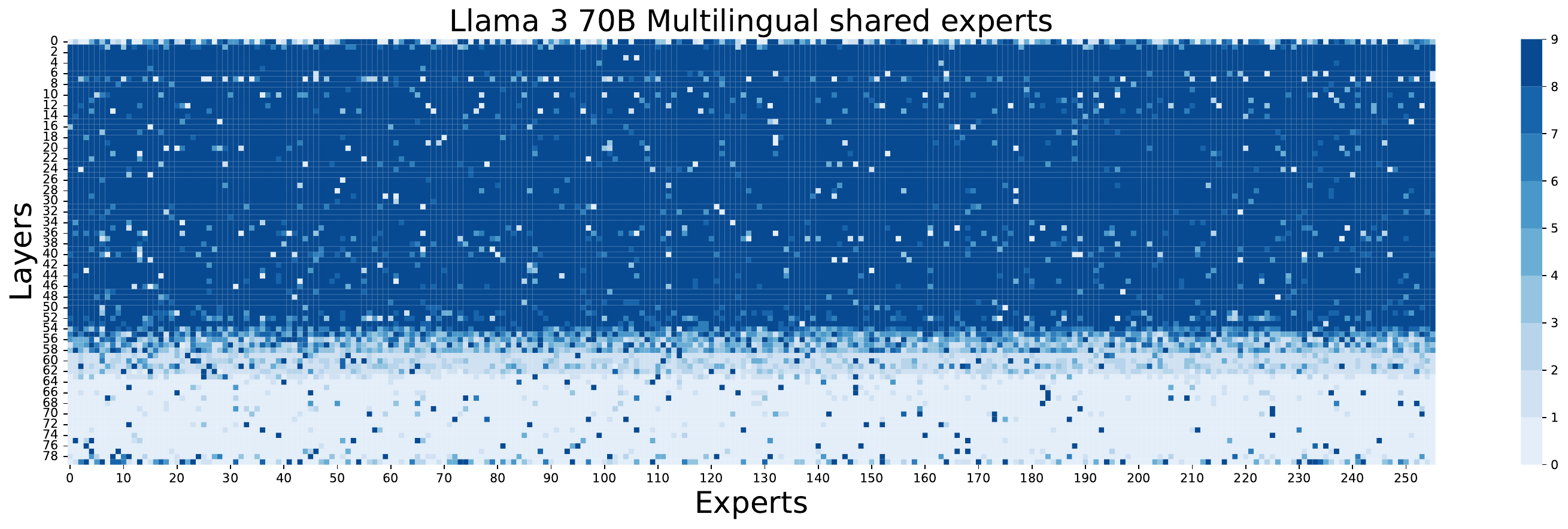}
  \end{minipage}
  
  \caption{The heatmaps of multilingual shared experts for Llama 3 8B and Llama 3 70B. The color shade of each cell indicates how many languages the expert is a high-frequency expert in.}
  \label{fig:heatmap_shared-2}
  \vspace{-3mm}
\end{figure*}

In Figure~\ref{fig:heatmap_shared-2}, we display the number of languages in which each expert is a high-frequency expert for Llama 3 8B and Llama 3 70B.

\section{Heatmaps of the impact of instruction tuning}
\label{sec:appendix_f}

\begin{figure*}[t]
  \centering  
  \begin{minipage}{\linewidth}
    \centering
    \includegraphics[width=\linewidth]{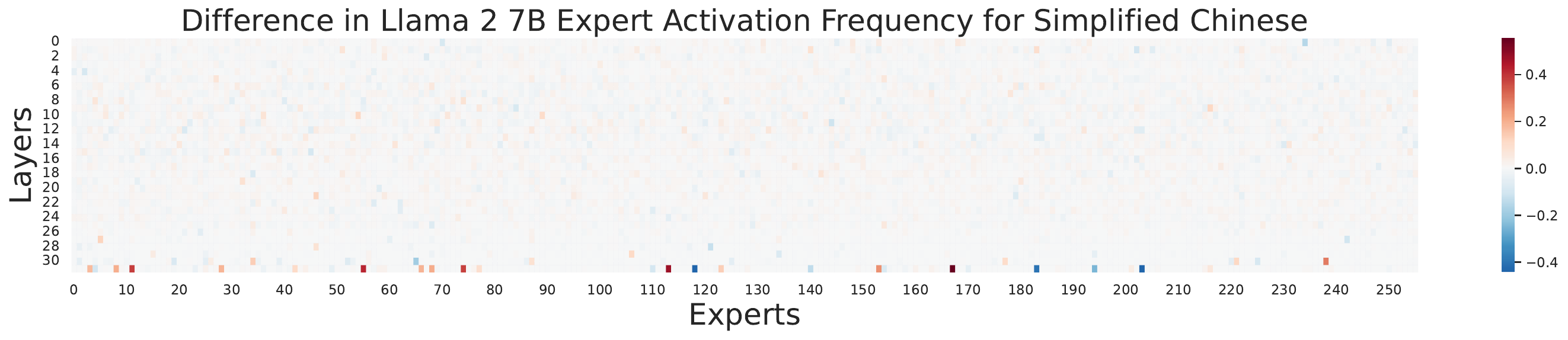}
    \vspace{-7mm}
  \end{minipage}
  
  \caption{Changes in expert activation frequency of Llama 2 7B instruction tuning variants across different languages compared to the original pre-trained model.}
  \label{fig:chat_llama2-7b}
  \vspace{-2mm}
\end{figure*}

\begin{figure*}[t]
  \centering
  \begin{minipage}{\linewidth}
    \centering
    \includegraphics[width=\linewidth]{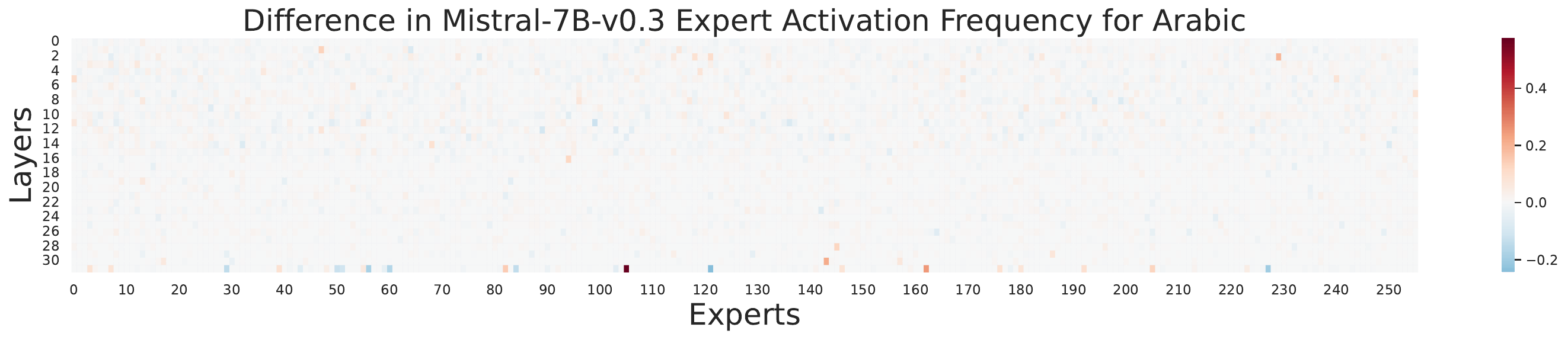}
    \vspace{-6mm}
  \end{minipage}

  \begin{minipage}{\linewidth}
    \centering
    \includegraphics[width=\linewidth]{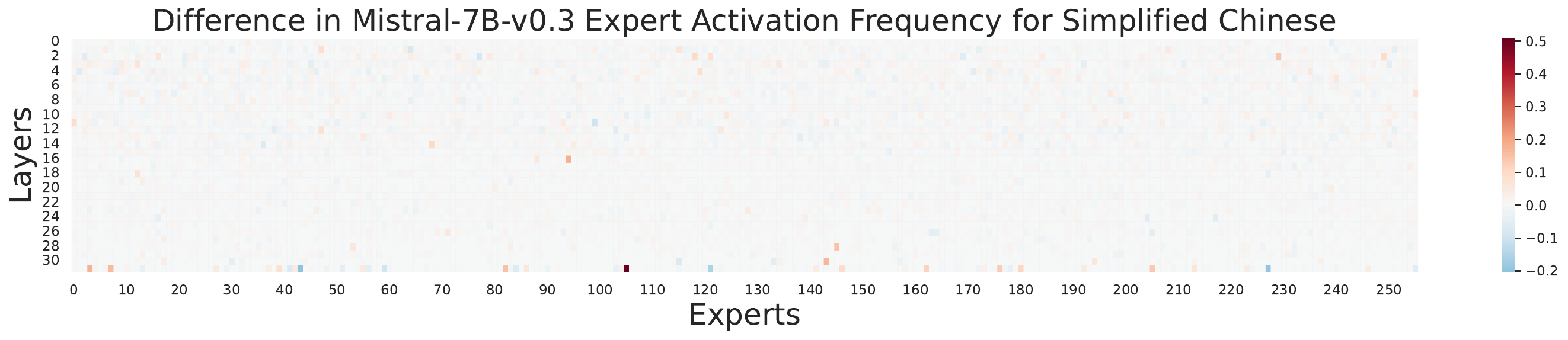}
    \vspace{-7mm}
  \end{minipage}

  \caption{Changes in expert activation frequency of Mistral-7B-v0.3 instruction tuning variants across different languages compared to the original pre-trained model.}
  \label{fig:chat_mistral}
  \vspace{-2mm}
\end{figure*}

\begin{figure*}[t]
  \centering
  \begin{minipage}{\linewidth}
    \centering
    \includegraphics[width=\linewidth]{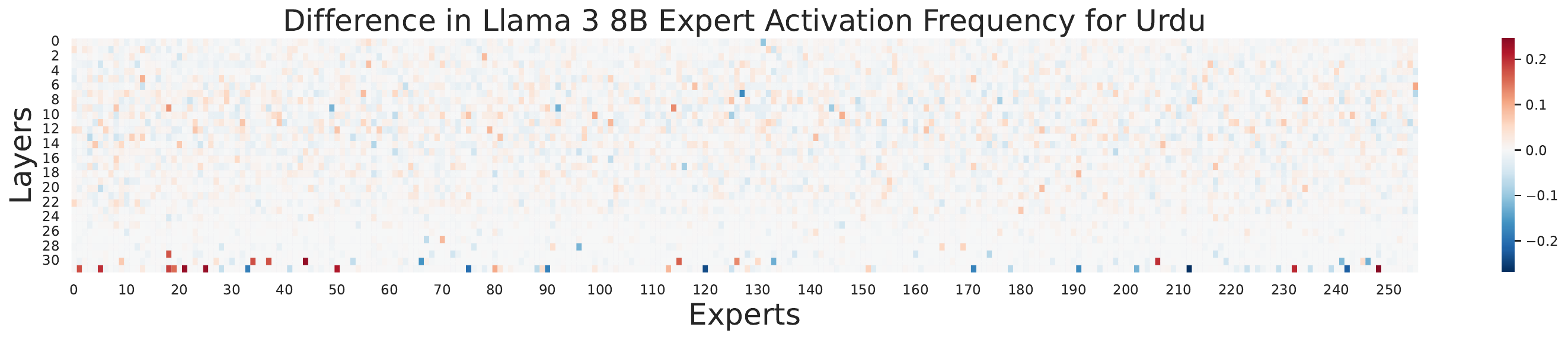}
    \vspace{-6mm}
  \end{minipage}

  \begin{minipage}{\linewidth}
    \centering
    \includegraphics[width=\linewidth]{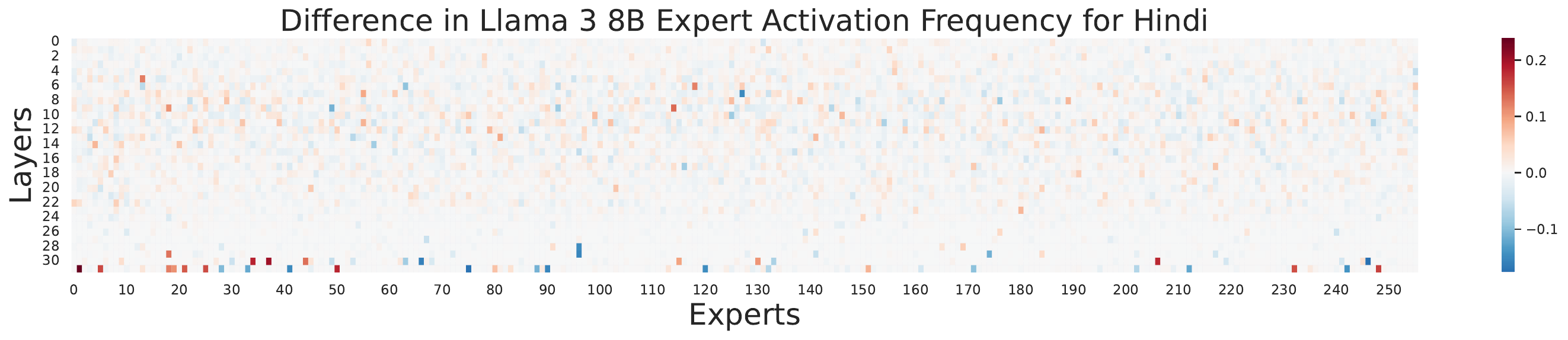}
    \vspace{-6mm}
  \end{minipage}

  \begin{minipage}{\linewidth}
    \centering
    \includegraphics[width=\linewidth]{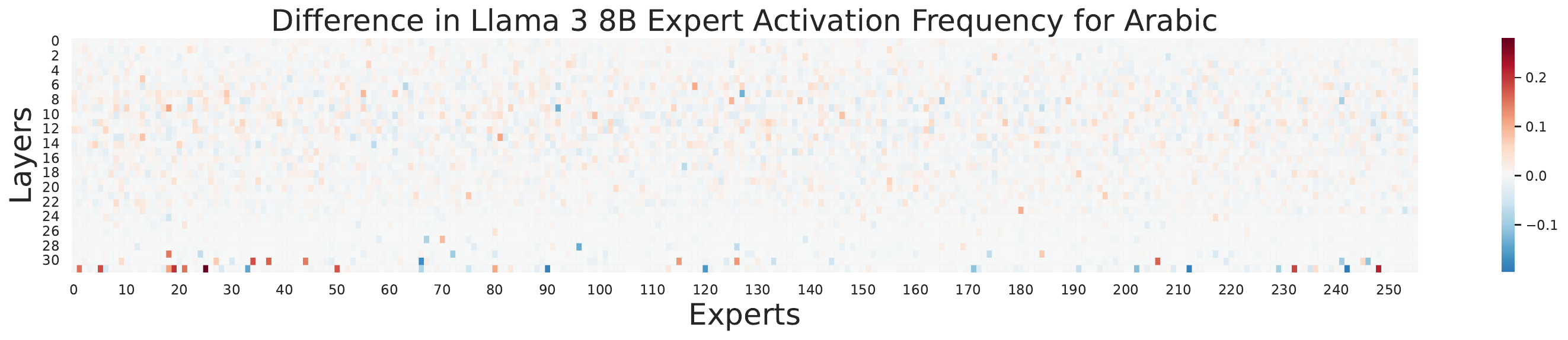}
    \vspace{-6mm}
  \end{minipage}

  \begin{minipage}{\linewidth}
    \centering
    \includegraphics[width=\linewidth]{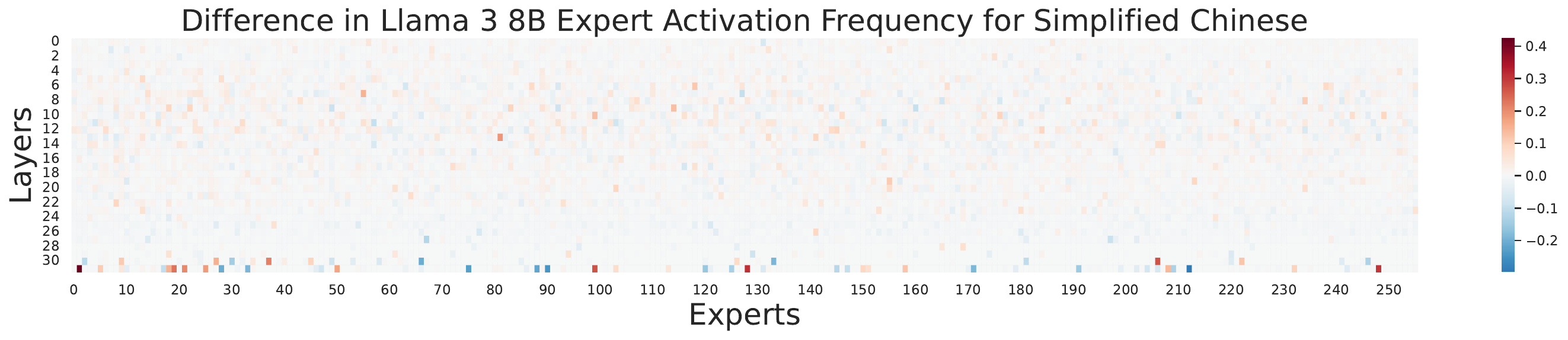}
    \vspace{-7mm}
  \end{minipage}

  \caption{Changes in expert activation frequency of Llama 3 8B instruction tuning variants across different languages compared to the original pre-trained model.}
  \label{fig:chat_llama3-8b}
  \vspace{-2mm}
\end{figure*}

\begin{figure*}[t]
  \begin{minipage}{\linewidth}
    \centering
    \includegraphics[width=\linewidth]{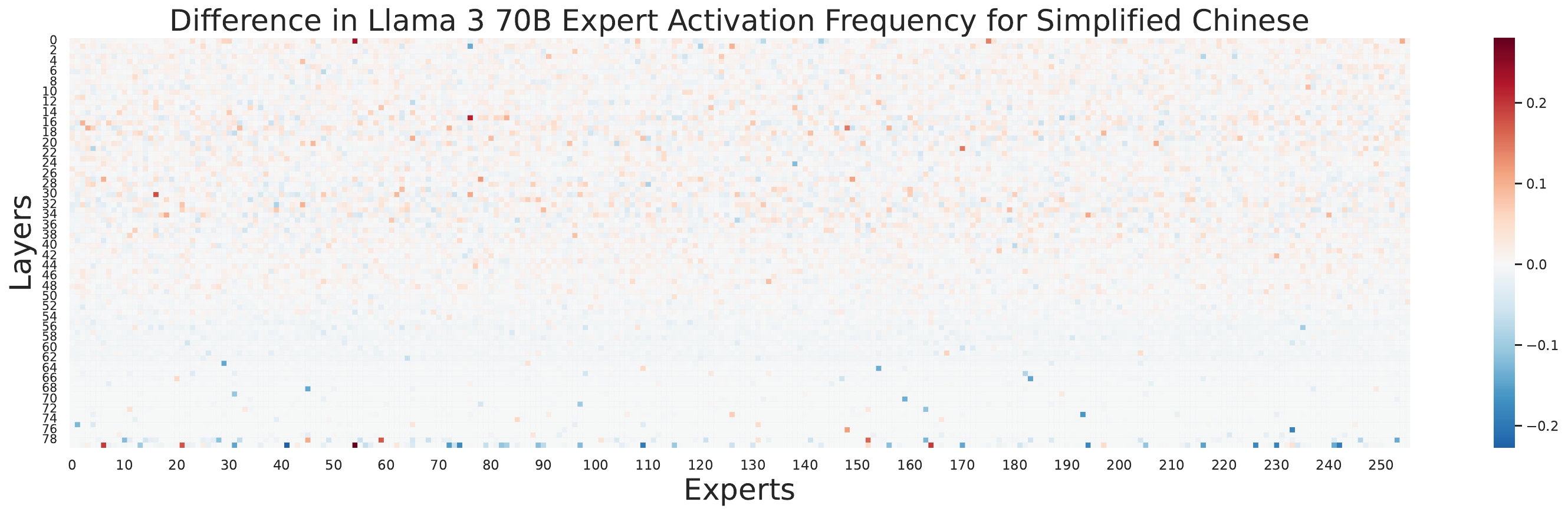}
    \vspace{-6mm}
  \end{minipage}

  \caption{Changes in expert activation frequency of Llama 3 70B instruction tuning variants across different languages compared to the original pre-trained model.}
  \label{fig:chat_llama3-70b}
  \vspace{-3mm}
\end{figure*}

In Figures~\ref{fig:chat_llama2-7b} to~\ref{fig:chat_llama3-70b}, we present the heatmaps of changes after instruction tuning for the remaining models and languages.

\section{Pruning results}
\label{sec:appendix_g}

\begin{figure*}[t]
  \centering
  \begin{subfigure}[b]{0.4\textwidth}
    \includegraphics[width=\linewidth]{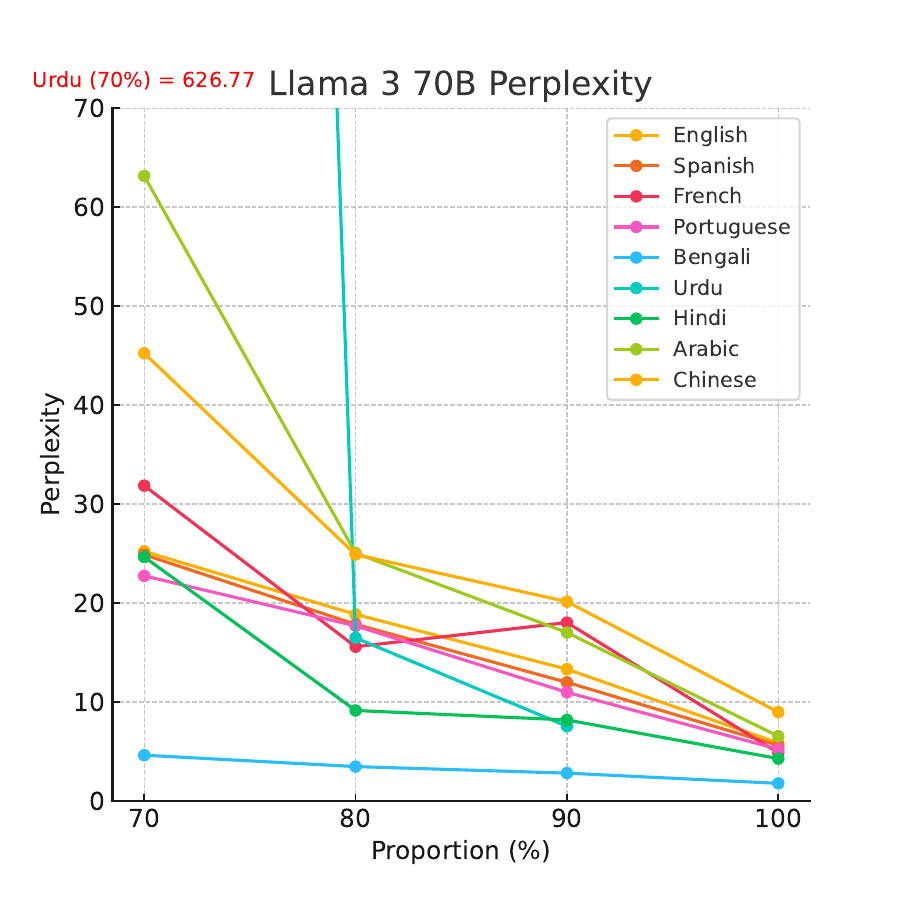}
  \end{subfigure}
  \begin{subfigure}[b]{0.4\textwidth}
    \includegraphics[width=\linewidth]{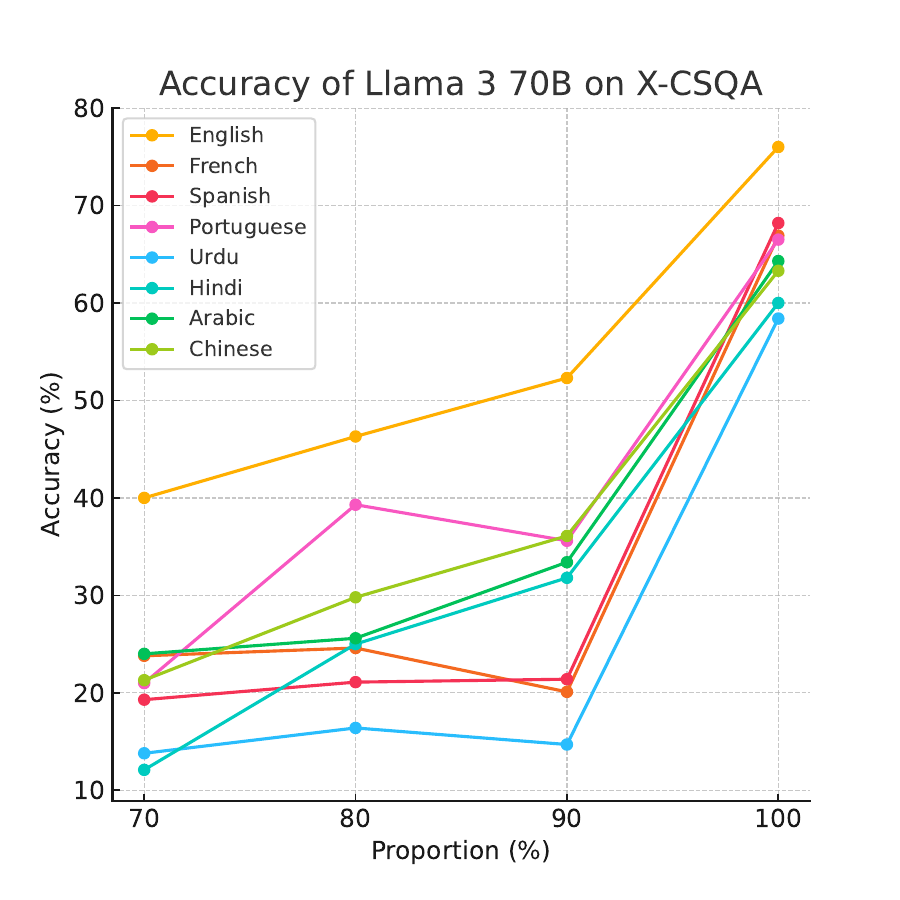}
  \end{subfigure}
  \vspace{-4mm}
  \caption{Results of Llama 3 70B pruning based on frequency sorting.}
  \label{fig:llam3_sort}
  \vspace{-4mm}
\end{figure*}

Tables~\ref{tab:ppl-70b}, ~\ref{tab:ppl-7b-chat}, and~\ref{tab:X-CSQA-7b-chat} present the remaining results of pruning based on frequency thresholds. Figure~\ref{fig:llam3_sort} illustrates the pruning based on frequency sorting results for Llama 3 70B.

\begin{figure*}[ht]
  \centering
  \begin{minipage}{\linewidth}
    \centering
    \includegraphics[width=\linewidth]{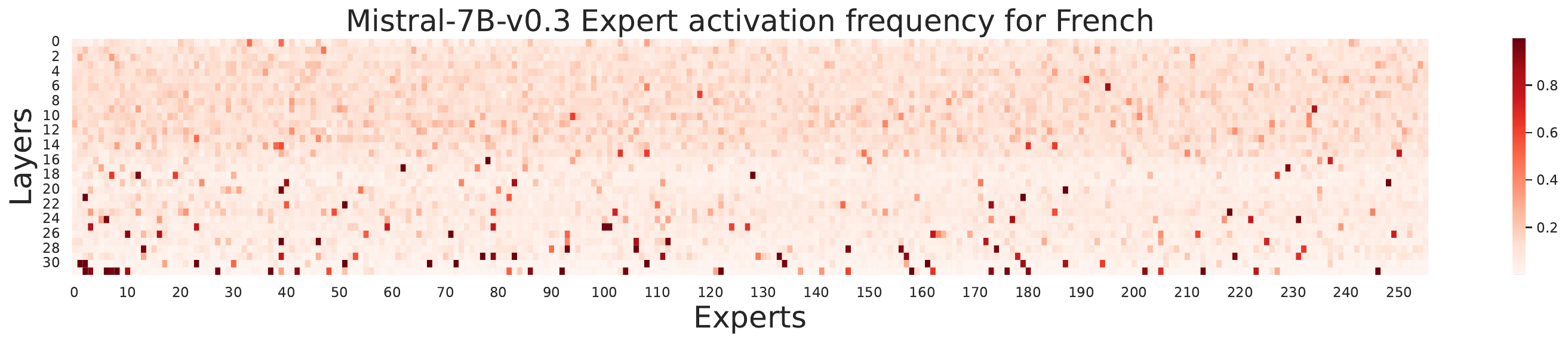}
    \vspace{-6mm}
  \end{minipage}

  \begin{minipage}{\linewidth}
    \centering
    \includegraphics[width=\linewidth]{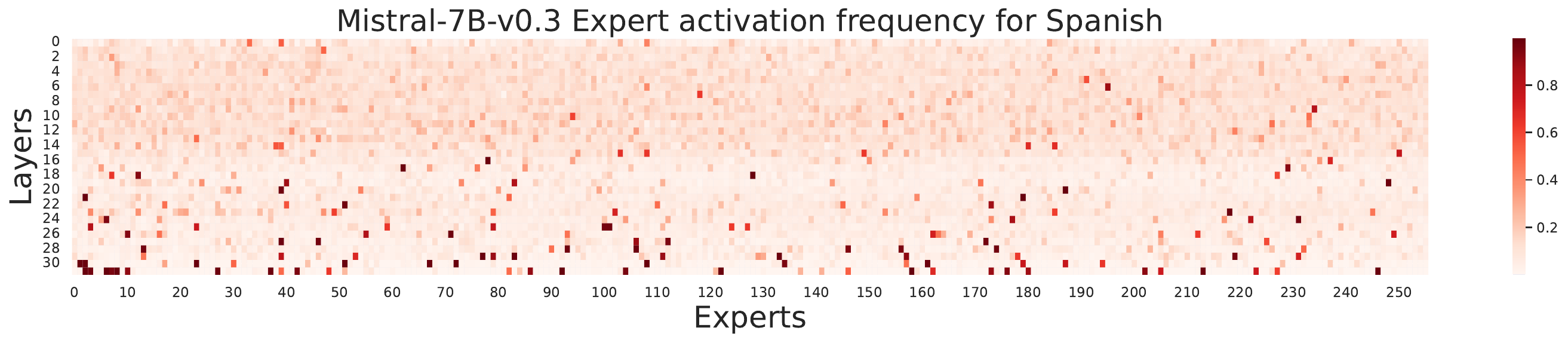}
    \vspace{-6mm}
  \end{minipage}

  \begin{minipage}{\linewidth}
    \centering
    \includegraphics[width=\linewidth]{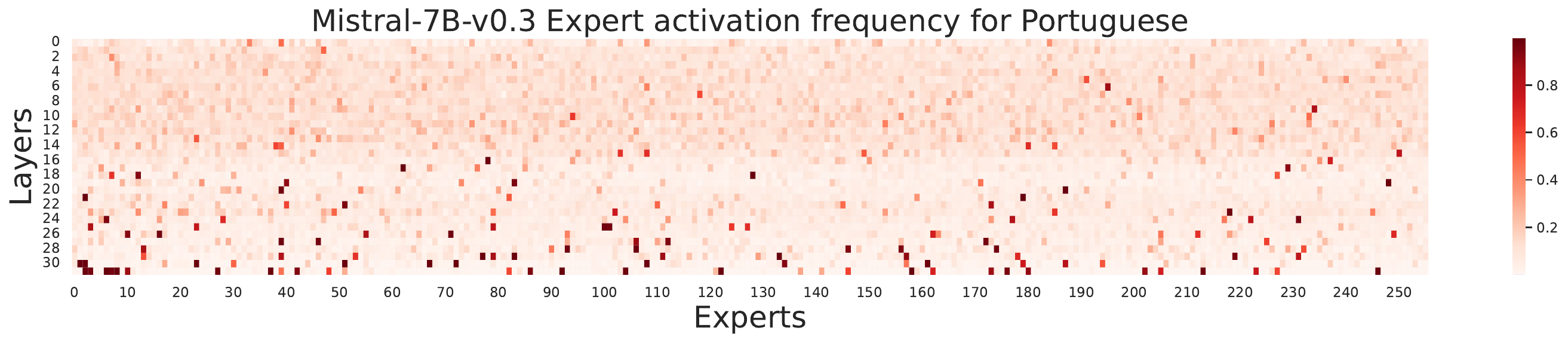}
    \vspace{-6mm}
  \end{minipage}
  
  \begin{minipage}{\linewidth}
    \centering
    \includegraphics[width=\linewidth]{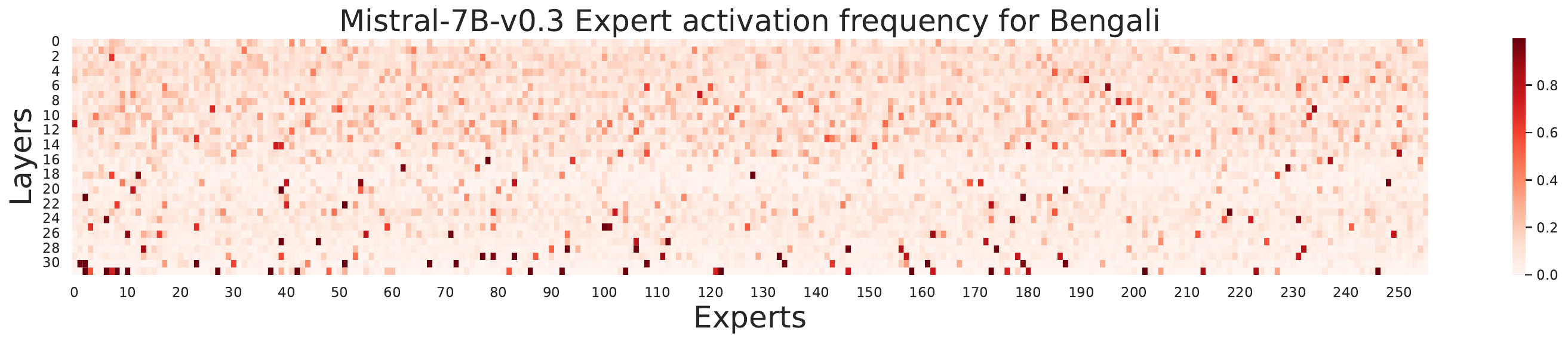}
    \vspace{-6mm}
  \end{minipage}

  \begin{minipage}{\linewidth}
    \centering
    \includegraphics[width=\linewidth]{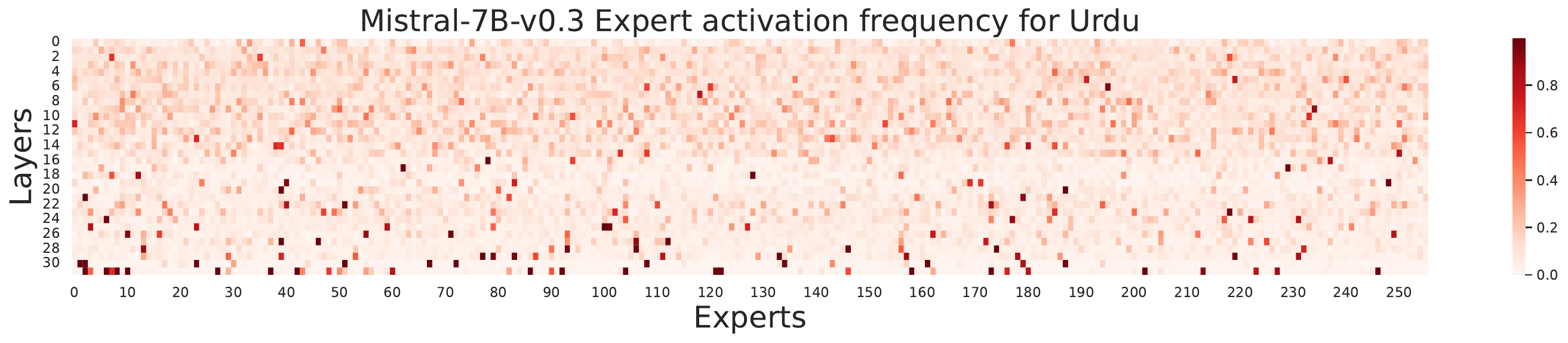}
    \vspace{-6mm}
  \end{minipage}
  
  \begin{minipage}{\linewidth}
    \centering
    \includegraphics[width=\linewidth]{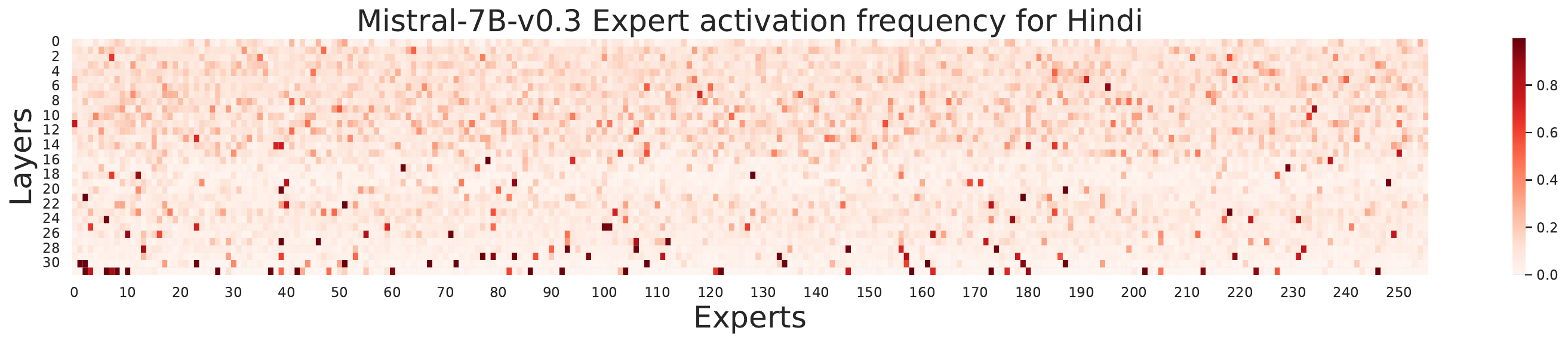}
    \vspace{-6mm}
  \end{minipage}
  
  \begin{minipage}{\linewidth}
    \centering
    \includegraphics[width=\linewidth]{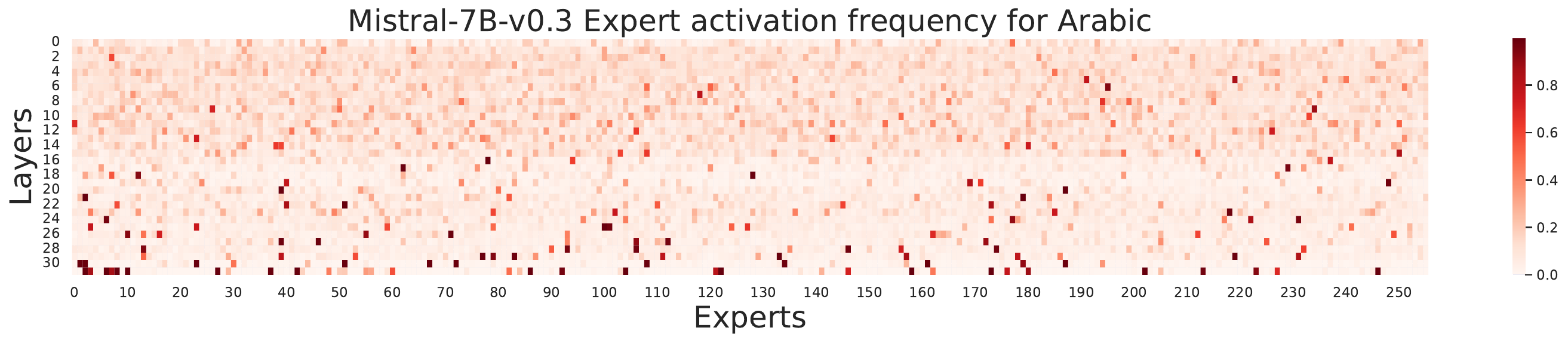}
    \vspace{-6mm}
  \end{minipage}
\end{figure*}

\begin{figure*}[ht]
  \centering
  \begin{minipage}{\linewidth}
    \centering
    \includegraphics[width=\linewidth]{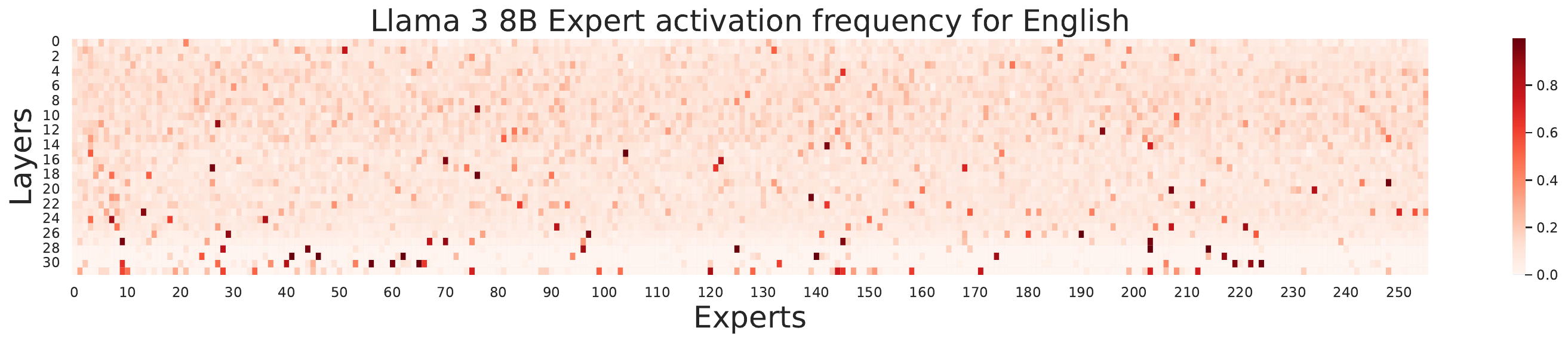}
    \vspace{-6mm}
  \end{minipage}

  \begin{minipage}{\linewidth}
    \centering
    \includegraphics[width=\linewidth]{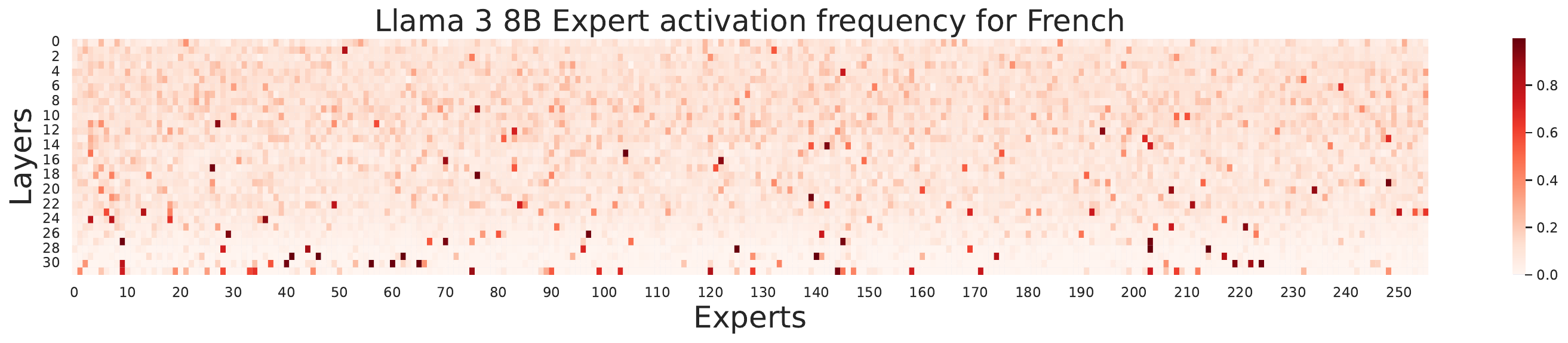}
    \vspace{-6mm}
  \end{minipage}

  \begin{minipage}{\linewidth}
    \centering
    \includegraphics[width=\linewidth]{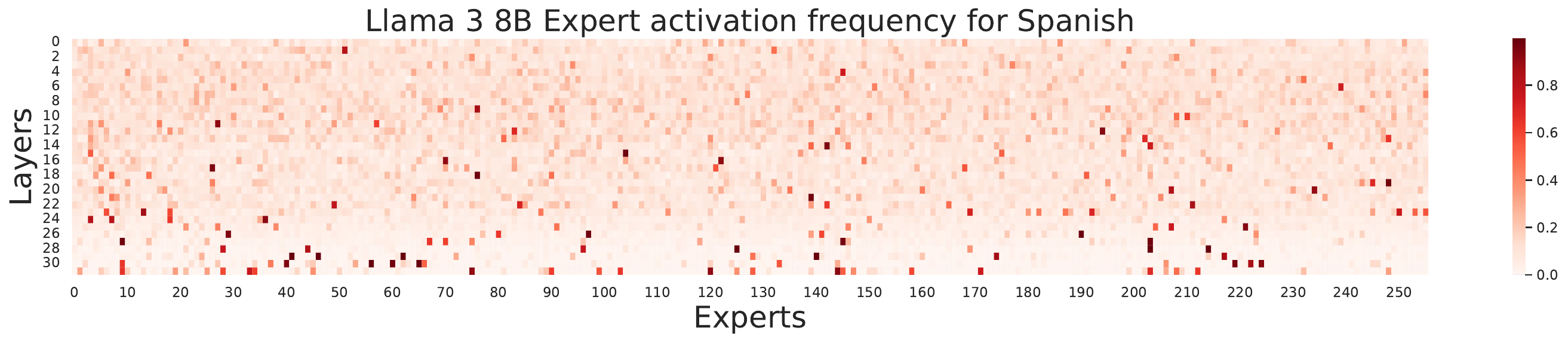}
    \vspace{-6mm}
  \end{minipage}
  
  \begin{minipage}{\linewidth}
    \centering
    \includegraphics[width=\linewidth]{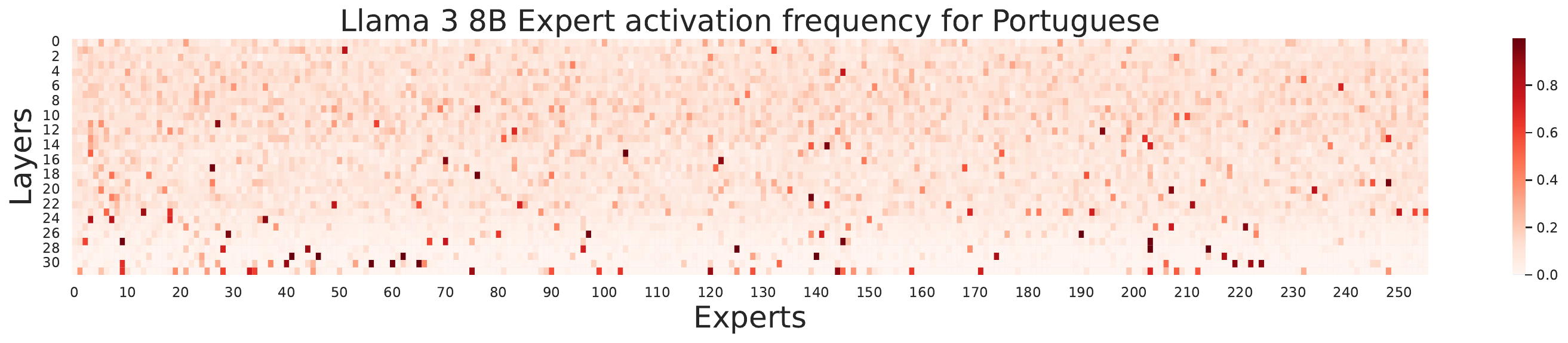}
    \vspace{-6mm}
  \end{minipage}

  \begin{minipage}{\linewidth}
    \centering
    \includegraphics[width=\linewidth]{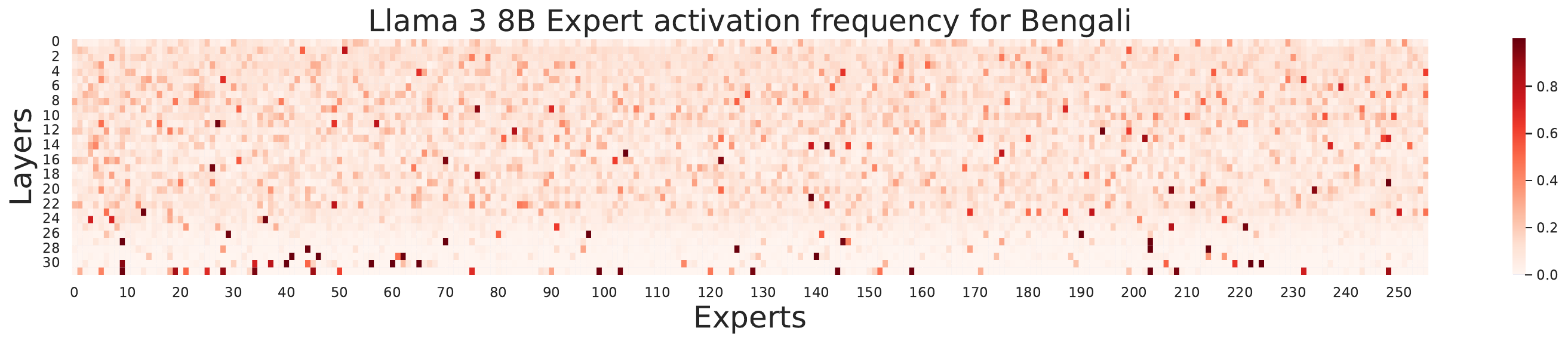}
    \vspace{-6mm}
  \end{minipage}
  
  \begin{minipage}{\linewidth}
    \centering
    \includegraphics[width=\linewidth]{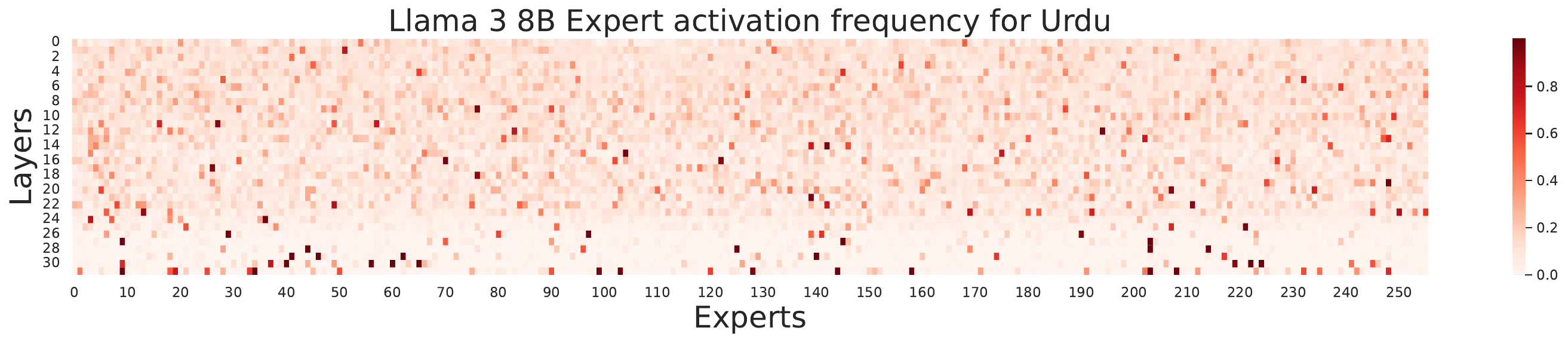}
    \vspace{-6mm}
  \end{minipage}
\end{figure*}

\begin{figure*}[ht]
  \centering
  \begin{minipage}{\linewidth}
    \centering
    \includegraphics[width=\linewidth]{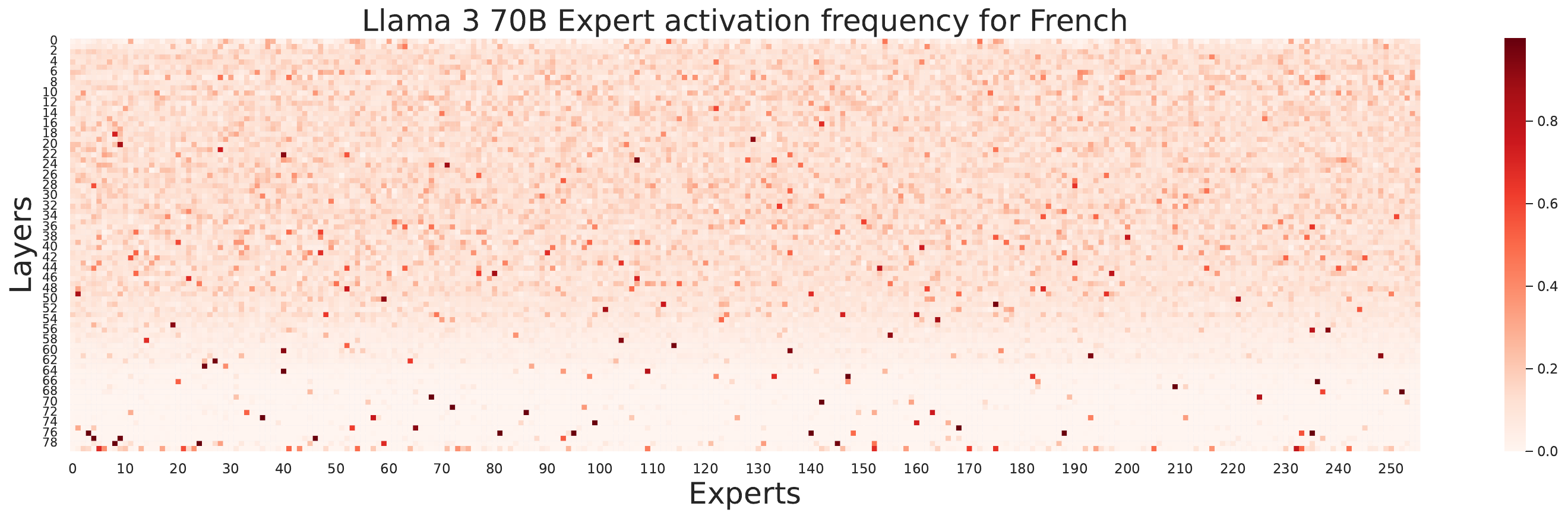}
    \vspace{-6mm}
  \end{minipage}

  \begin{minipage}{\linewidth}
    \centering
    \includegraphics[width=\linewidth]{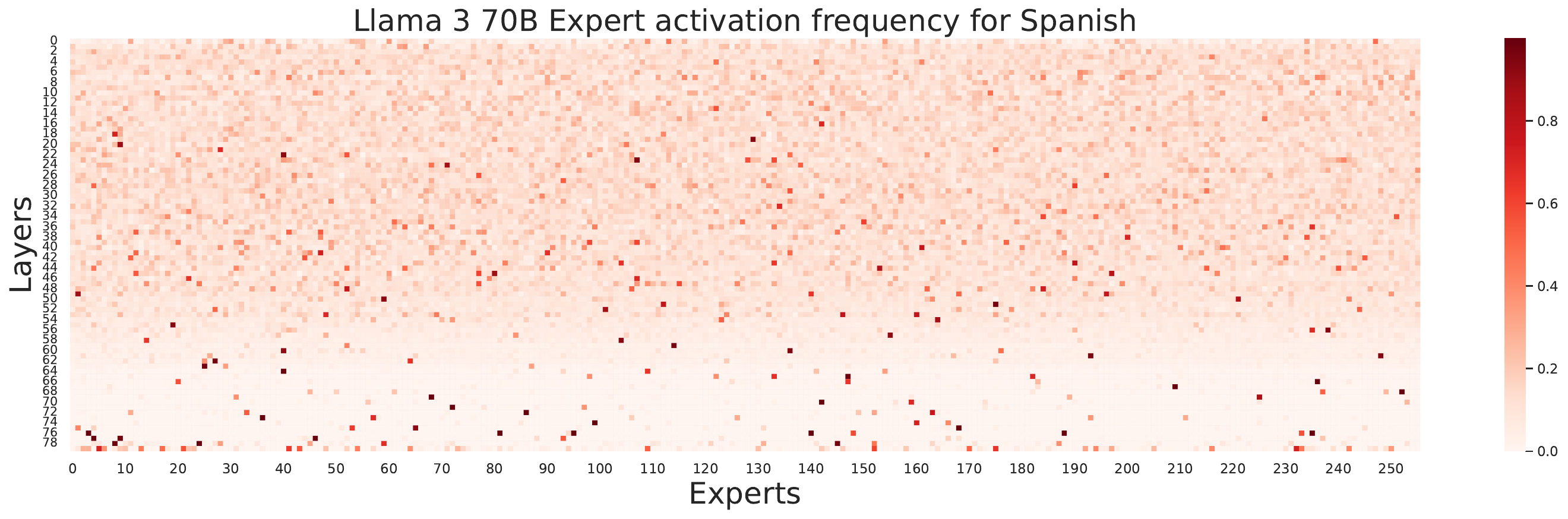}
    \vspace{-6mm}
  \end{minipage}
\end{figure*}

\begin{figure*}[t]
  \begin{minipage}{\linewidth}
    \centering
    \includegraphics[width=\linewidth]{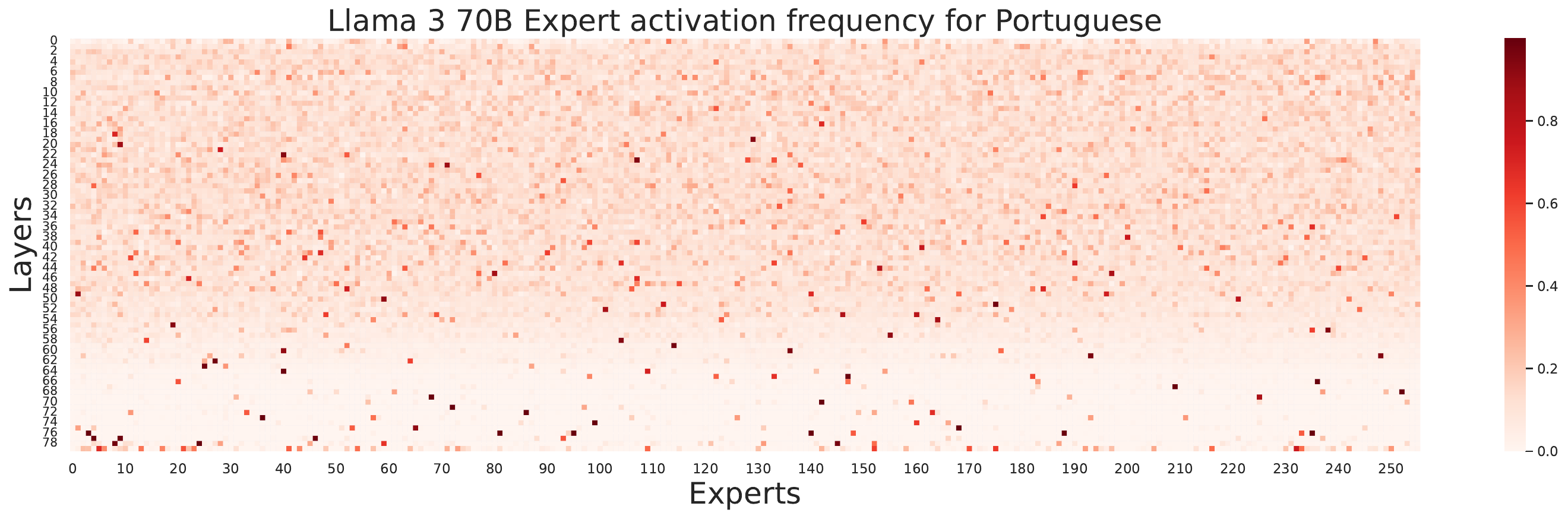}
    \vspace{-6mm}
  \end{minipage}
  
  \begin{minipage}{\linewidth}
    \centering
    \includegraphics[width=\linewidth]{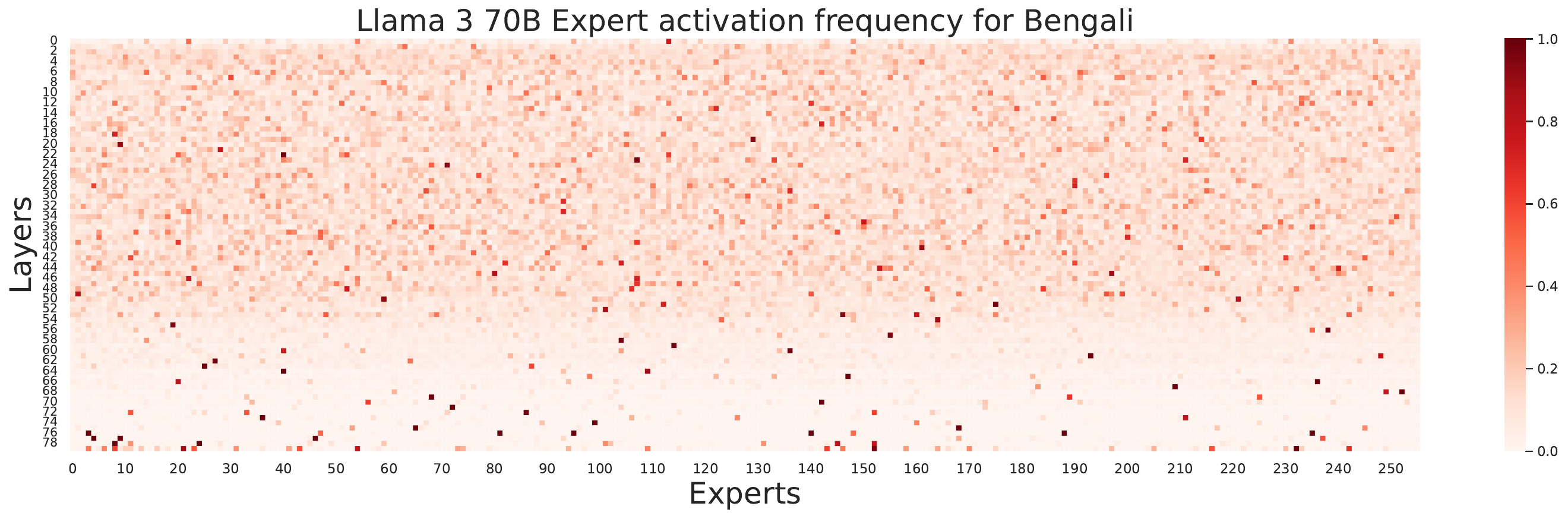}
    \vspace{-6mm}
  \end{minipage}

  \begin{minipage}{\linewidth}
    \centering
    \includegraphics[width=\linewidth]{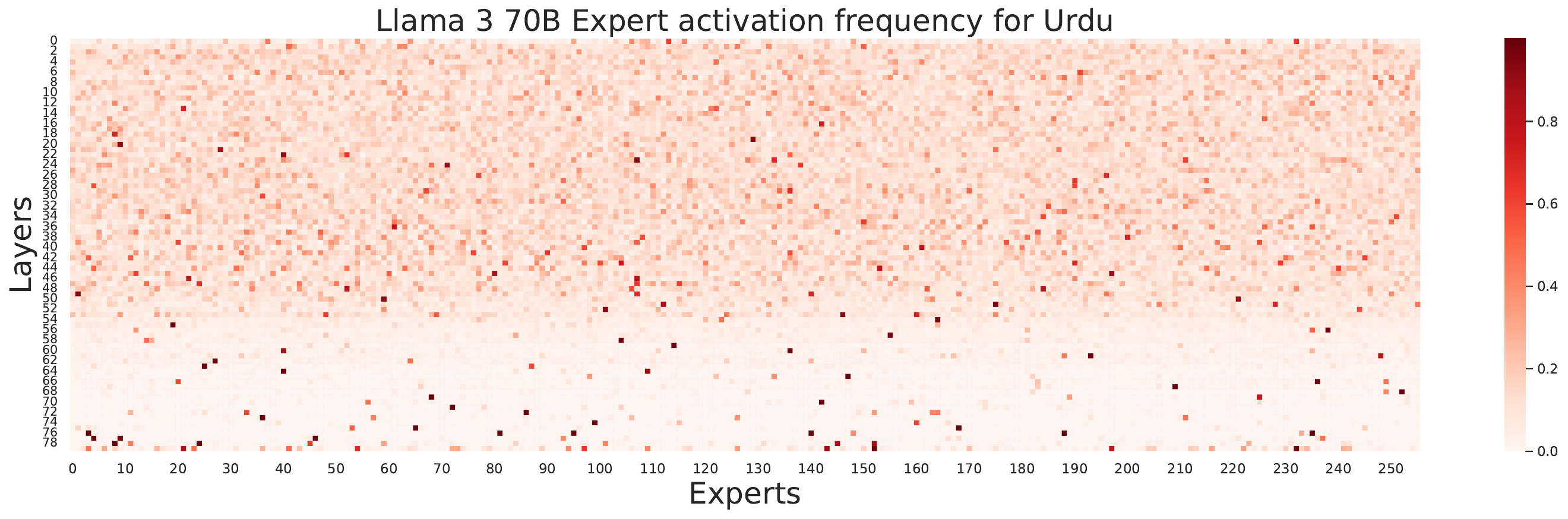}
    \vspace{-6mm}
  \end{minipage}
\end{figure*}

\begin{figure*}[ht]
  \centering
  \begin{minipage}{\linewidth}
    \centering
    \includegraphics[width=\linewidth]{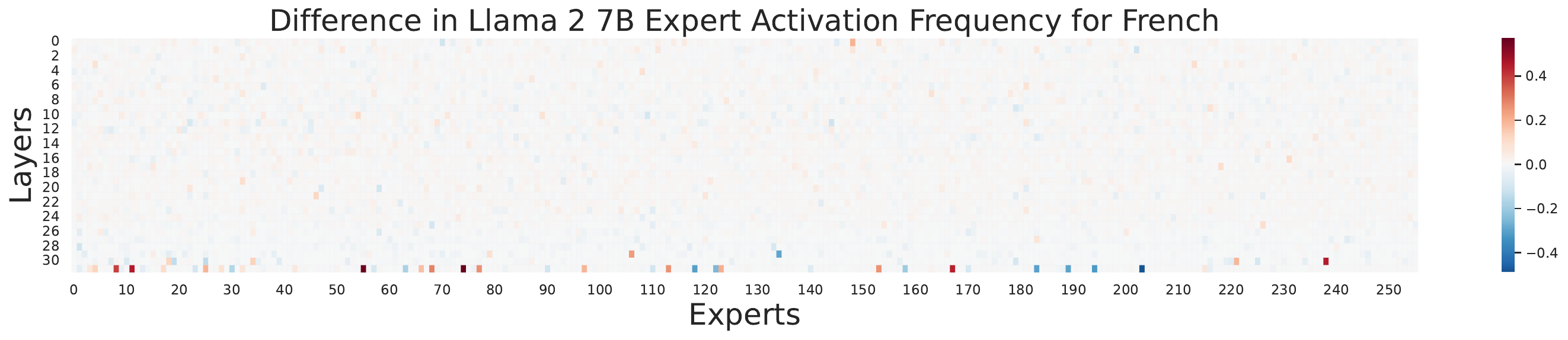}
    \vspace{-6mm}
  \end{minipage}

  \begin{minipage}{\linewidth}
    \centering
    \includegraphics[width=\linewidth]{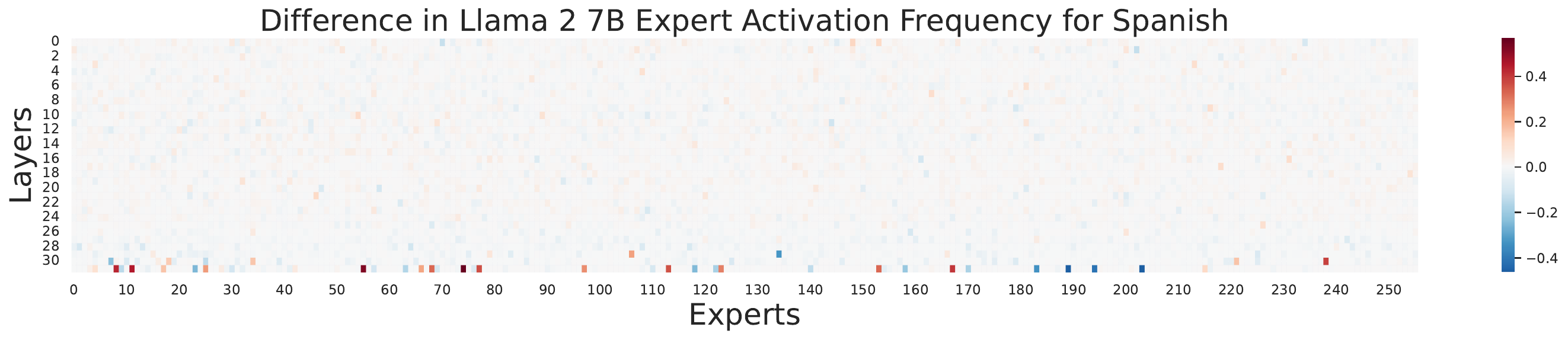}
    \vspace{-6mm}
  \end{minipage}

  \begin{minipage}{\linewidth}
    \centering
    \includegraphics[width=\linewidth]{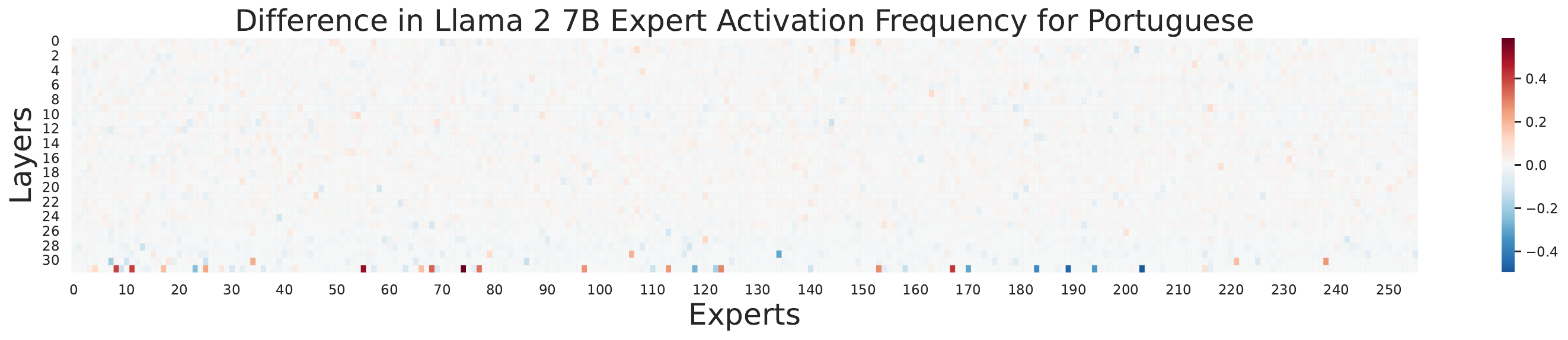}
    \vspace{-6mm}
  \end{minipage}

  \begin{minipage}{\linewidth}
    \centering
    \includegraphics[width=\linewidth]{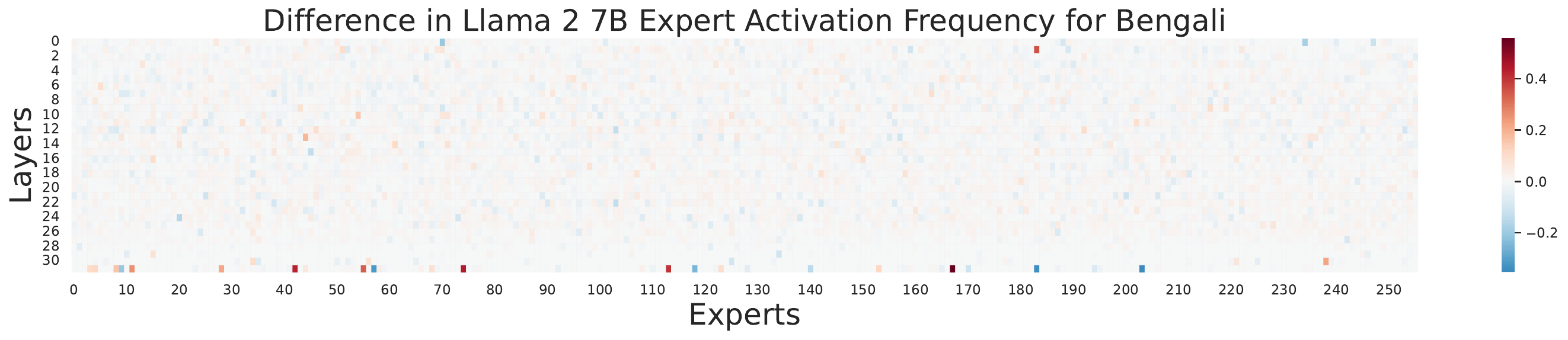}
    \vspace{-6mm}
  \end{minipage}

  \begin{minipage}{\linewidth}
    \centering
    \includegraphics[width=\linewidth]{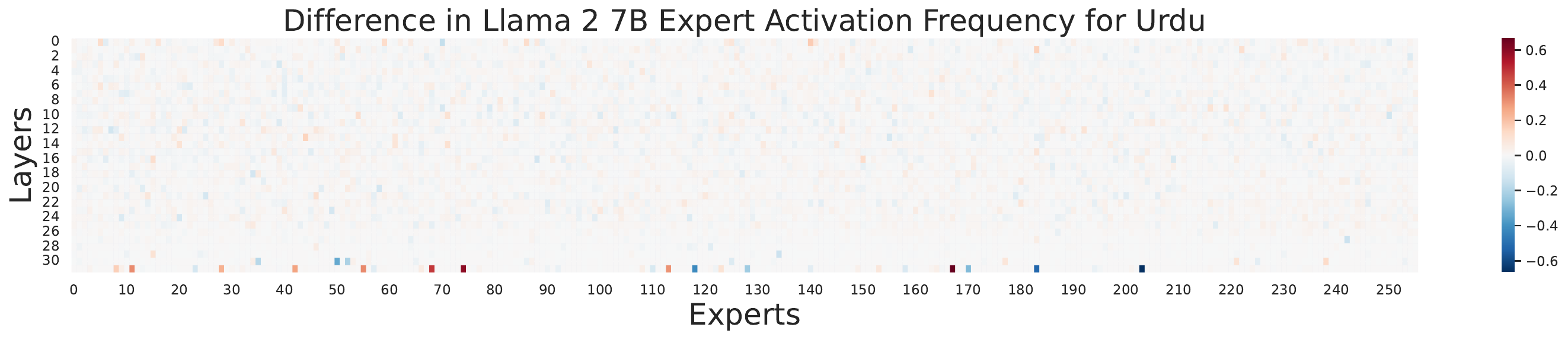}
    \vspace{-6mm}
  \end{minipage}

  \begin{minipage}{\linewidth}
    \centering
    \includegraphics[width=\linewidth]{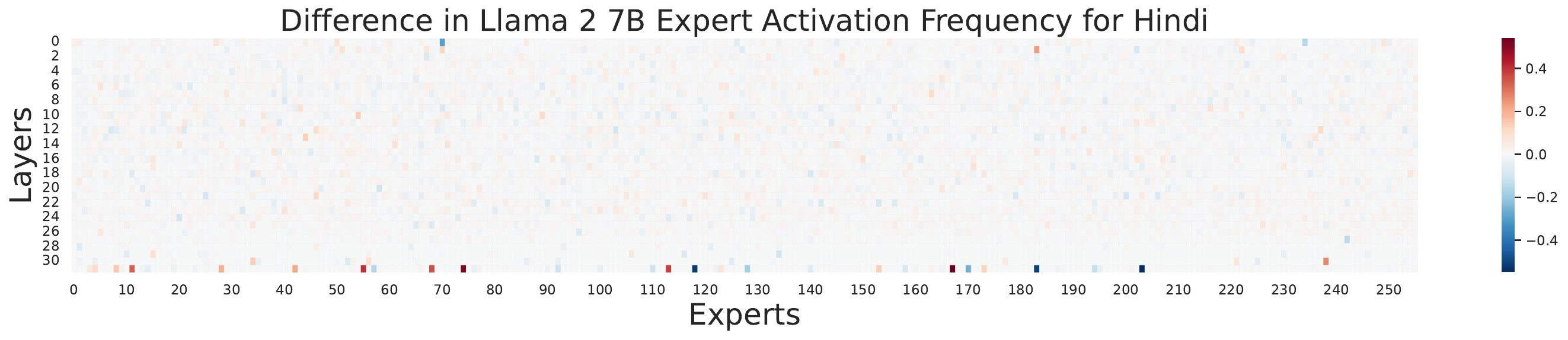}
    \vspace{-6mm}
  \end{minipage}

  \begin{minipage}{\linewidth}
    \centering
    \includegraphics[width=\linewidth]{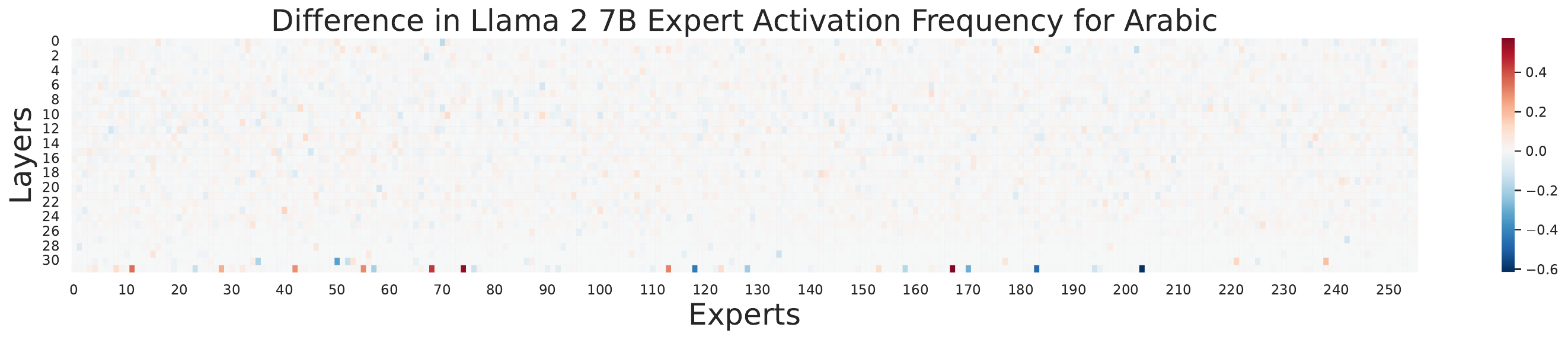}
    \vspace{-6mm}
  \end{minipage}
\end{figure*}

\begin{figure*}[ht]
  \centering
  \begin{minipage}{\linewidth}
    \centering
    \includegraphics[width=\linewidth]{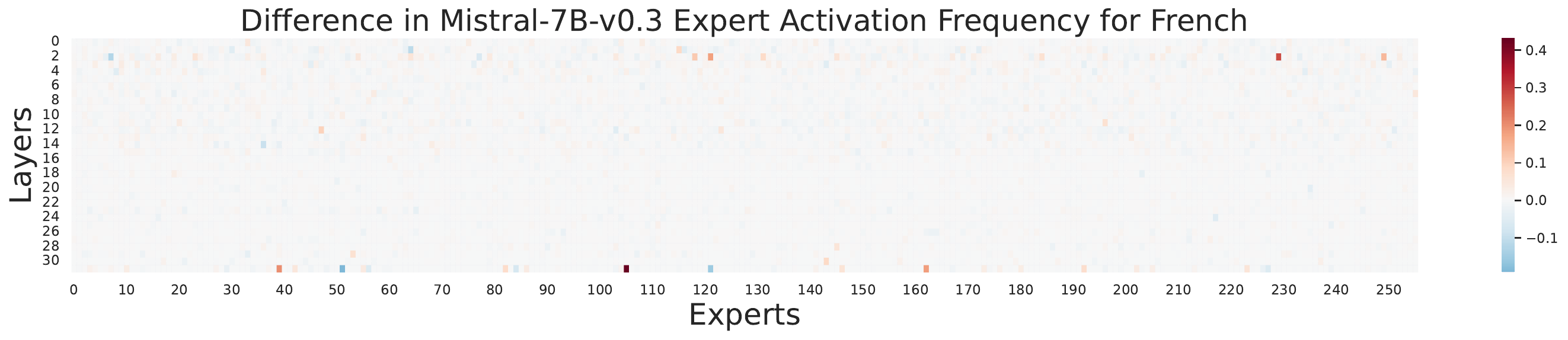}
    \vspace{-6mm}
  \end{minipage}

  \begin{minipage}{\linewidth}
    \centering
    \includegraphics[width=\linewidth]{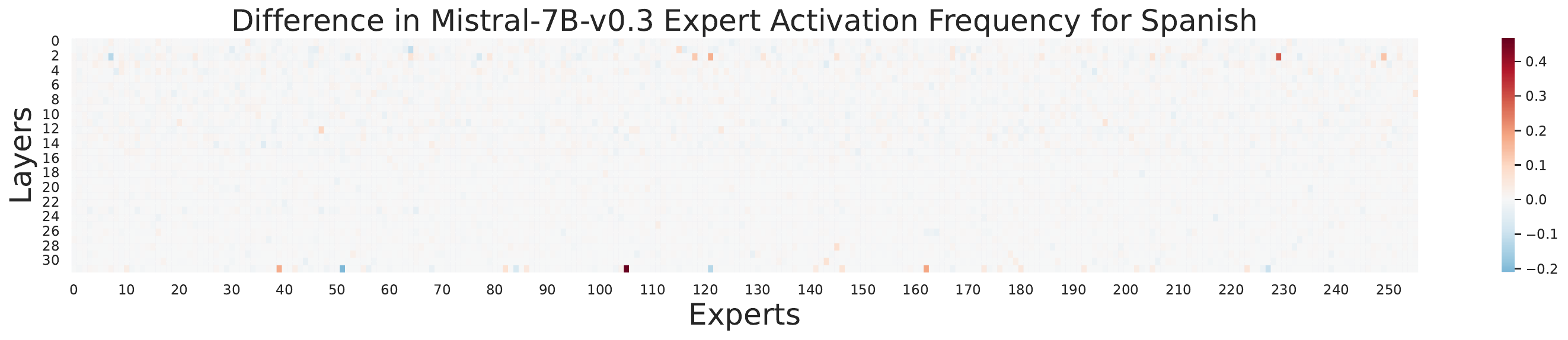}
    \vspace{-6mm}
  \end{minipage}

  \begin{minipage}{\linewidth}
    \centering
    \includegraphics[width=\linewidth]{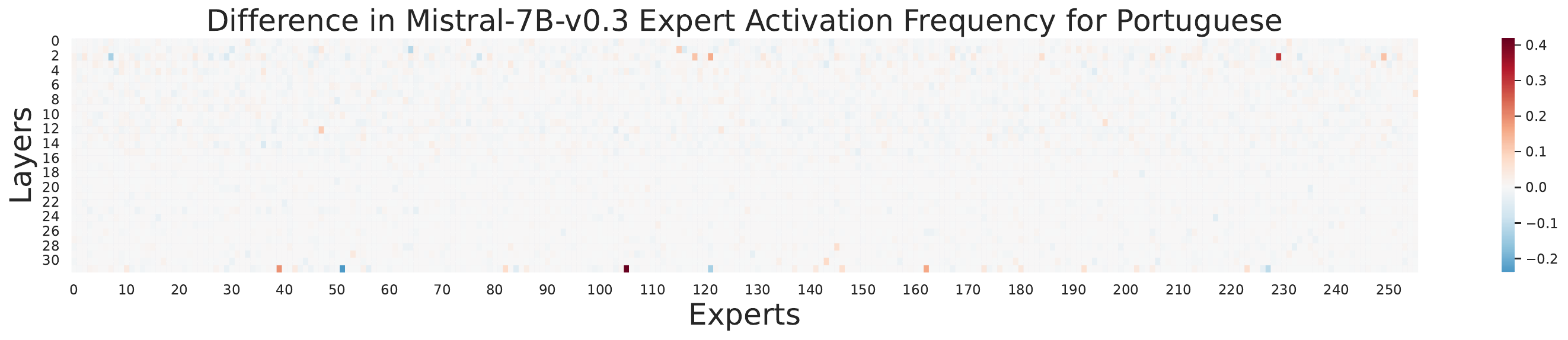}
    \vspace{-6mm}
  \end{minipage}
  
  \begin{minipage}{\linewidth}
    \centering
    \includegraphics[width=\linewidth]{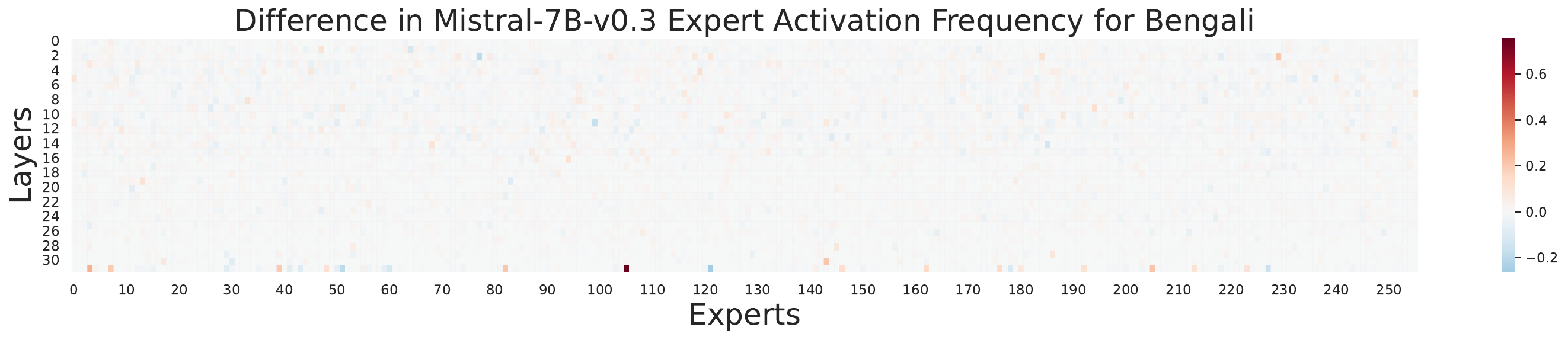}
    \vspace{-6mm}
  \end{minipage}

  \begin{minipage}{\linewidth}
    \centering
    \includegraphics[width=\linewidth]{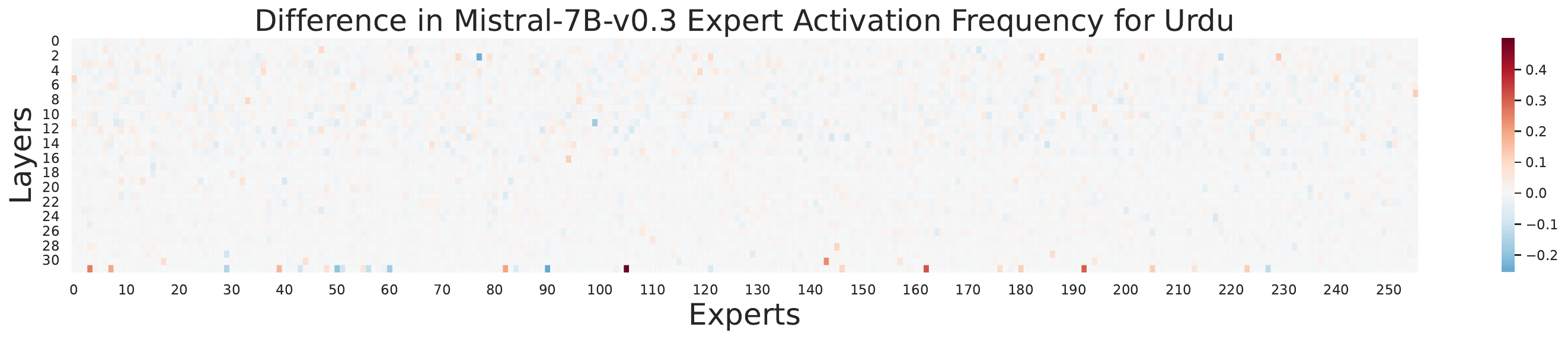}
    \vspace{-6mm}
  \end{minipage}
  
  \begin{minipage}{\linewidth}
    \centering
    \includegraphics[width=\linewidth]{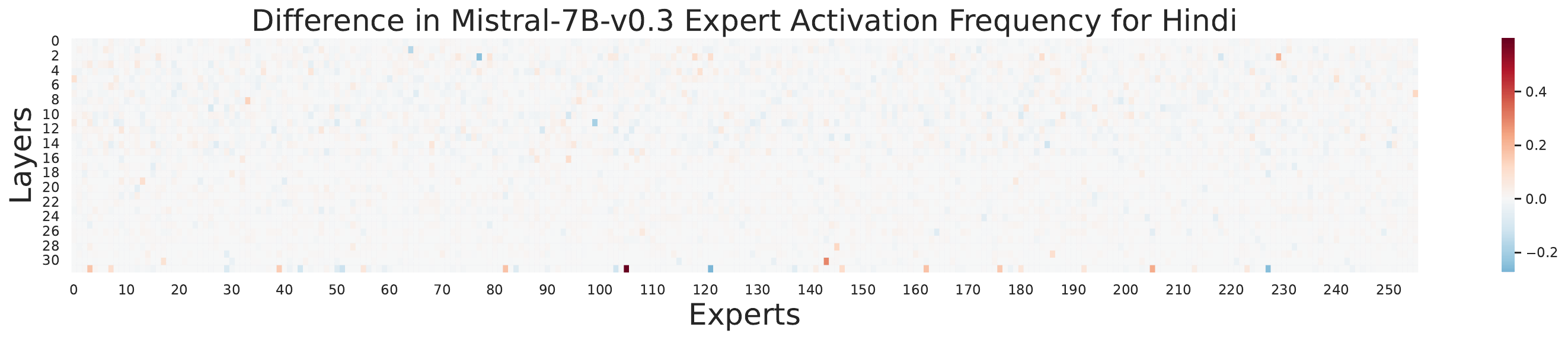}
    \vspace{-6mm}
  \end{minipage}
\end{figure*}

\begin{figure*}[ht]
  \centering
  \begin{minipage}{\linewidth}
    \centering
    \includegraphics[width=\linewidth]{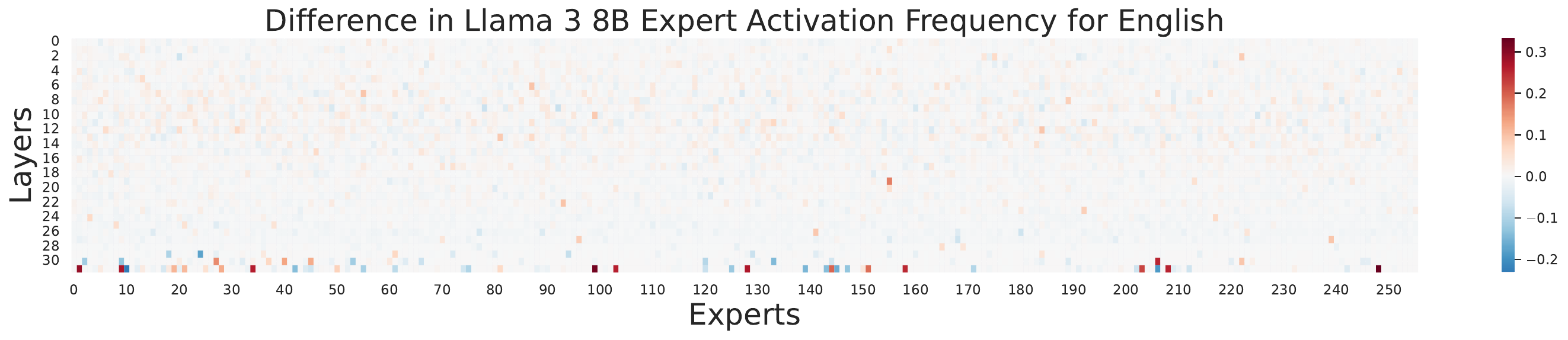}
    \vspace{-6mm}
  \end{minipage}

  \begin{minipage}{\linewidth}
    \centering
    \includegraphics[width=\linewidth]{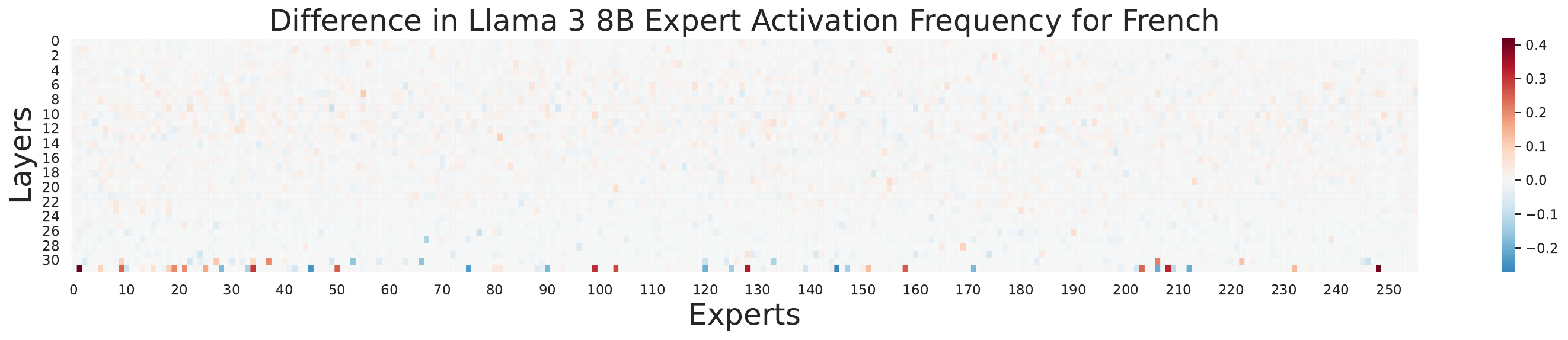}
    \vspace{-6mm}
  \end{minipage}

  \begin{minipage}{\linewidth}
    \centering
    \includegraphics[width=\linewidth]{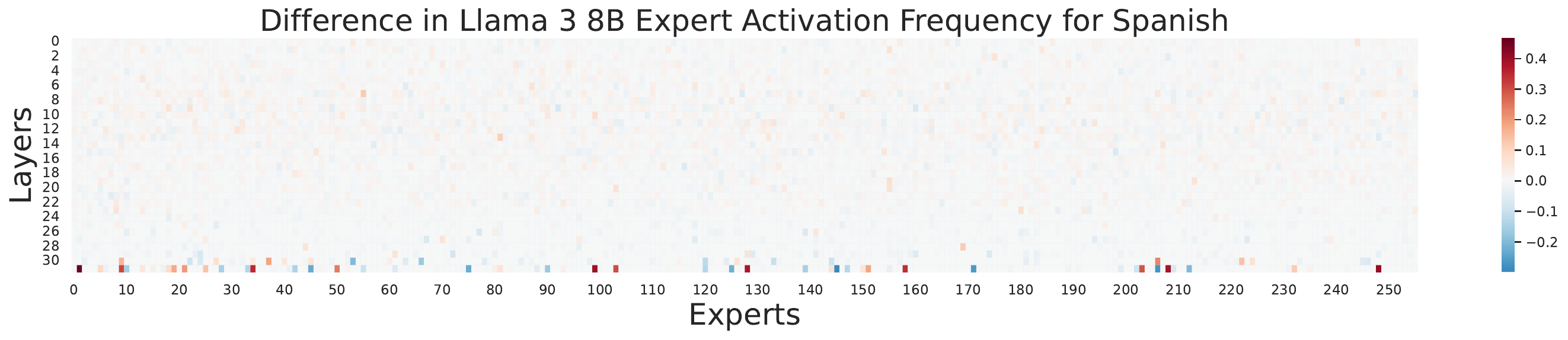}
    \vspace{-6mm}
  \end{minipage}
  
  \begin{minipage}{\linewidth}
    \centering
    \includegraphics[width=\linewidth]{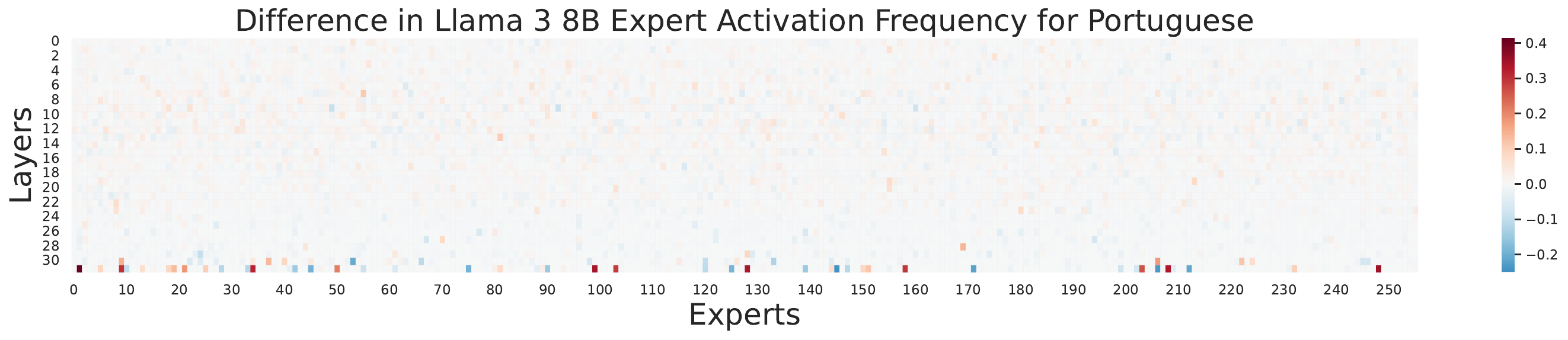}
    \vspace{-6mm}
  \end{minipage}

  \begin{minipage}{\linewidth}
    \centering
    \includegraphics[width=\linewidth]{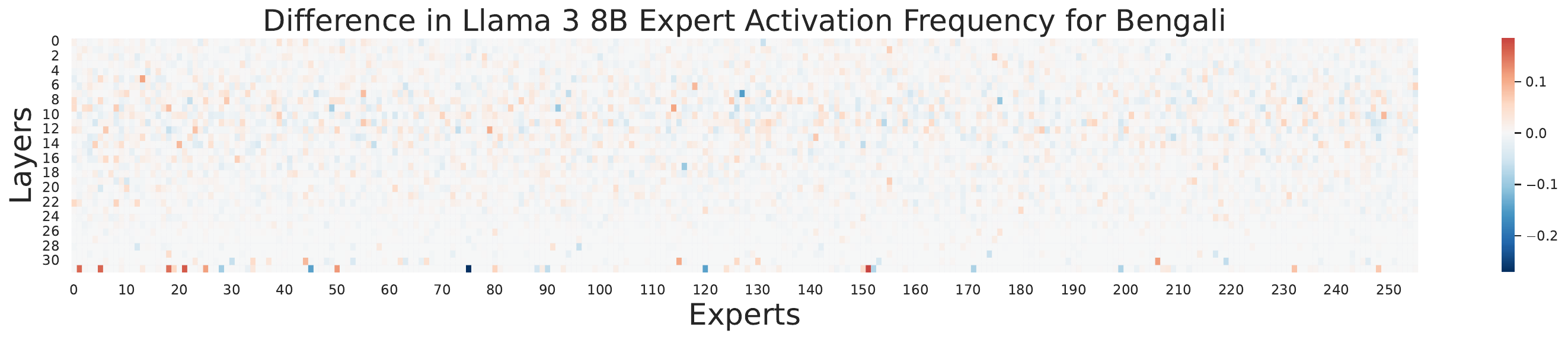}
    \vspace{-6mm}
  \end{minipage}
\end{figure*}

\begin{figure*}[ht]
  \centering
  \begin{minipage}{\linewidth}
    \centering
    \includegraphics[width=\linewidth]{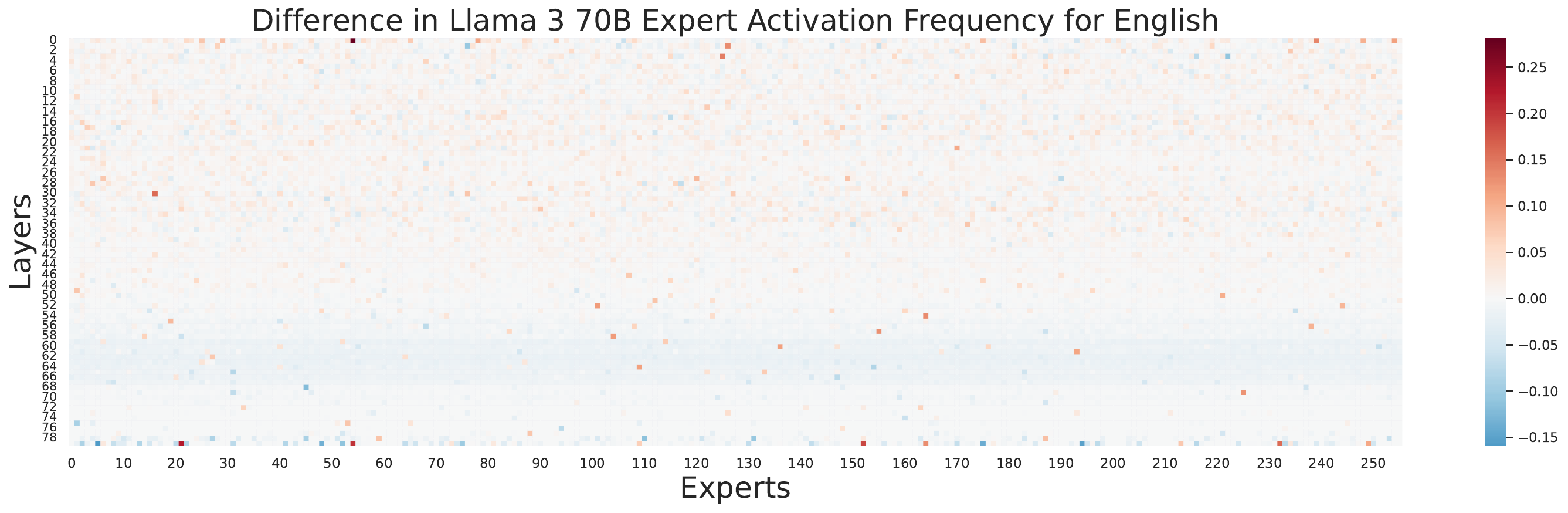}
    \vspace{-6mm}
  \end{minipage}

  \begin{minipage}{\linewidth}
    \centering
    \includegraphics[width=\linewidth]{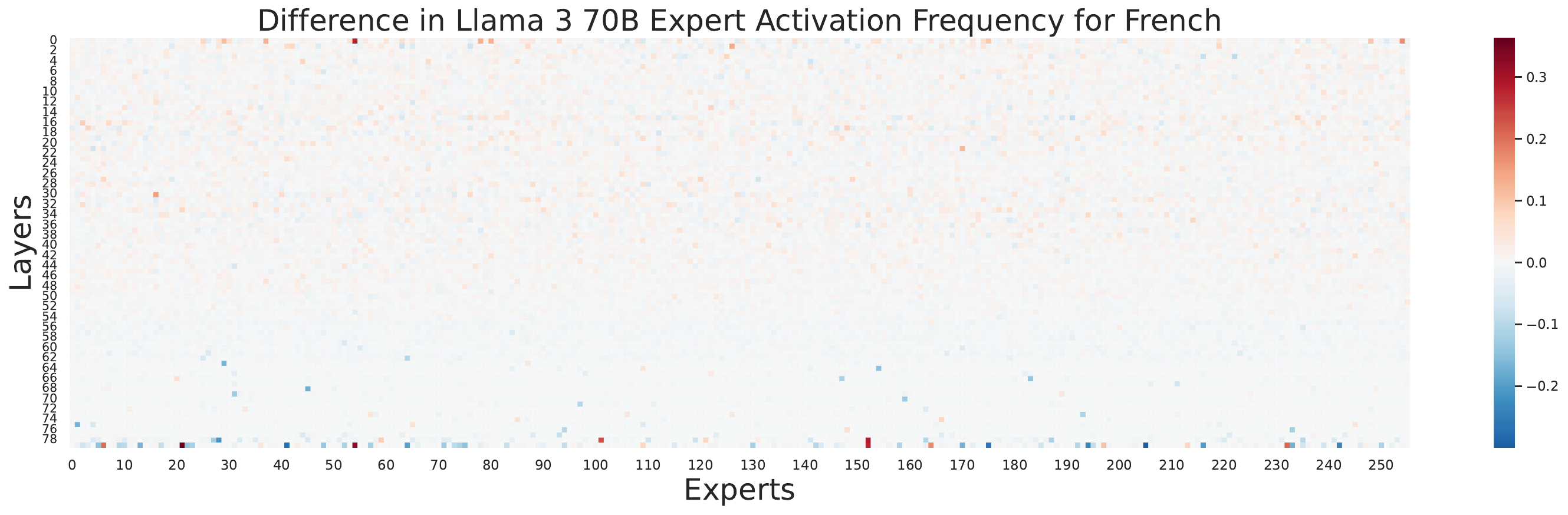}
    \vspace{-6mm}
  \end{minipage}

  \begin{minipage}{\linewidth}
    \centering
    \includegraphics[width=\linewidth]{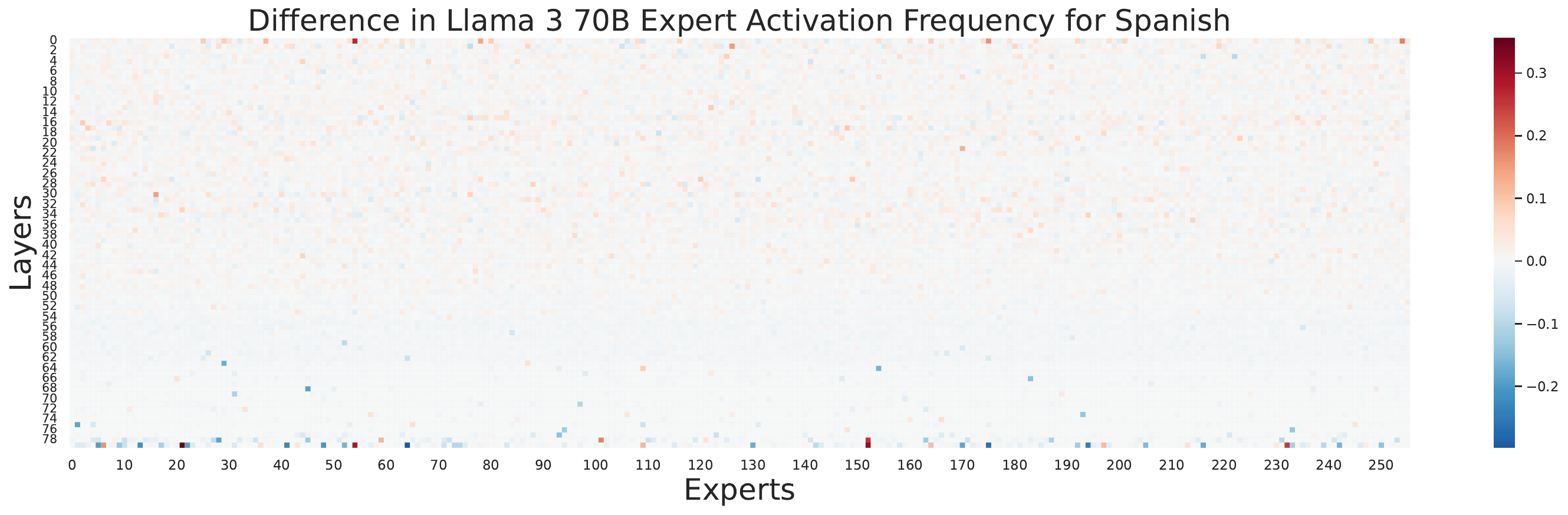}
    \vspace{-6mm}
  \end{minipage}
  
  \begin{minipage}{\linewidth}
    \centering
    \includegraphics[width=\linewidth]{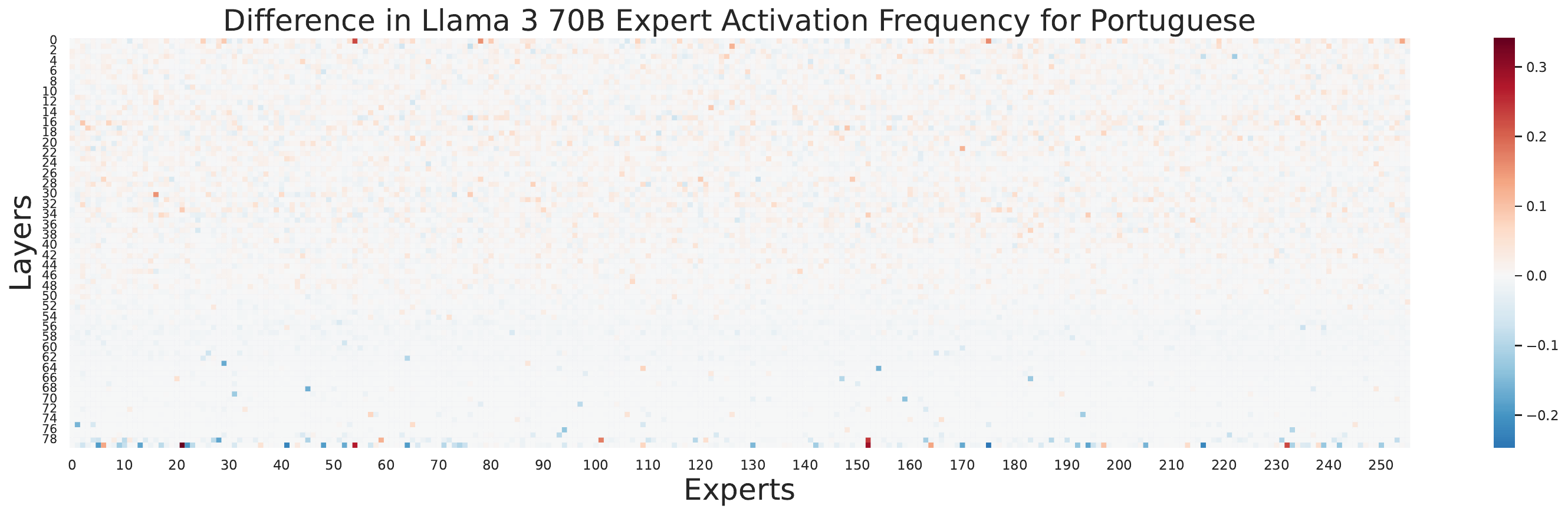}
    \vspace{-6mm}
  \end{minipage}
\end{figure*}

\begin{figure*}[t]
  \begin{minipage}{\linewidth}
    \centering
    \includegraphics[width=\linewidth]{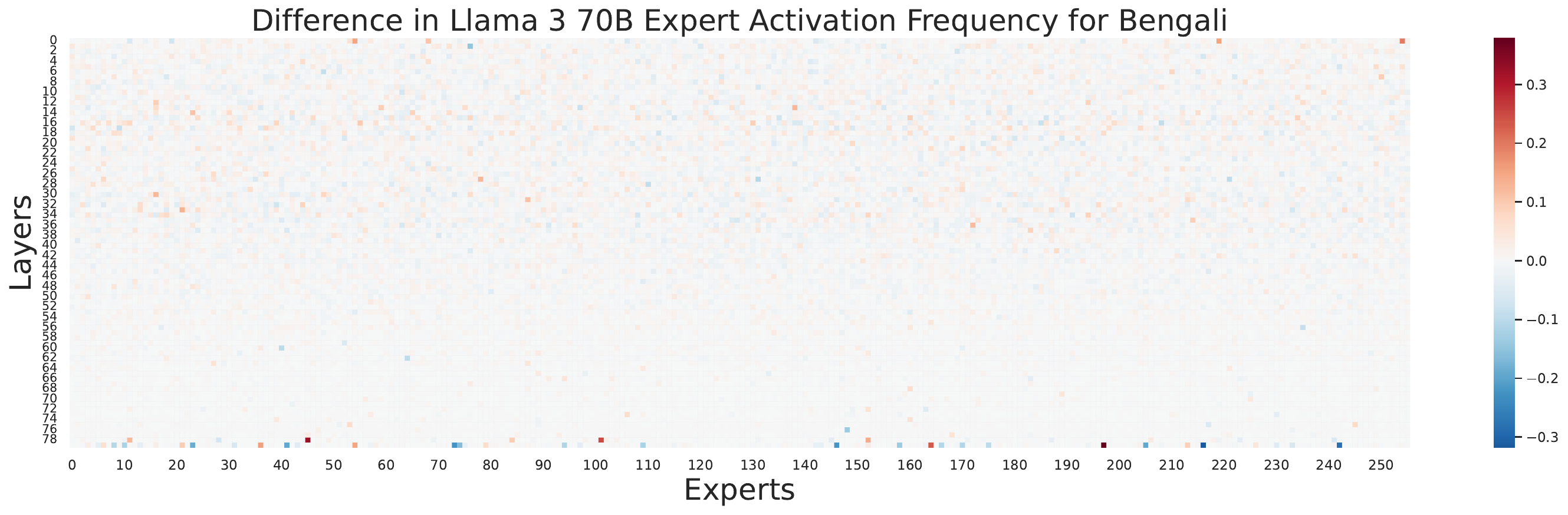}
    \vspace{-6mm}
  \end{minipage}

  \centering
  \begin{minipage}{\linewidth}
    \centering
    \includegraphics[width=\linewidth]{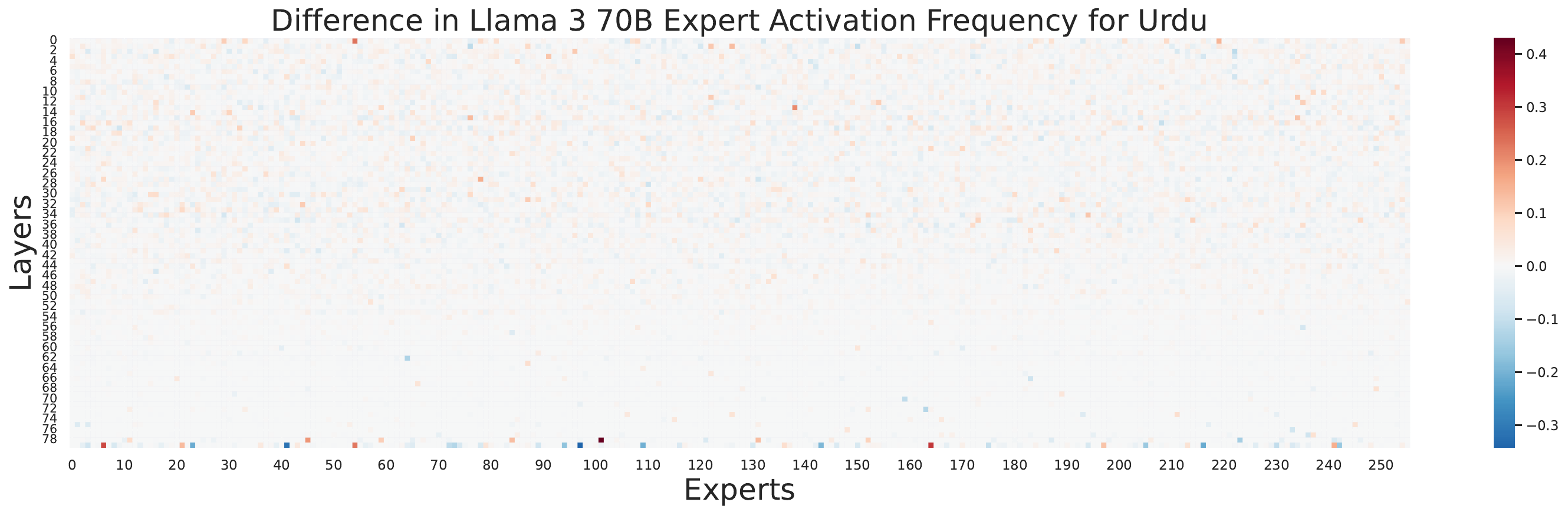}
    \vspace{-6mm}
  \end{minipage}

  \begin{minipage}{\linewidth}
    \centering
    \includegraphics[width=\linewidth]{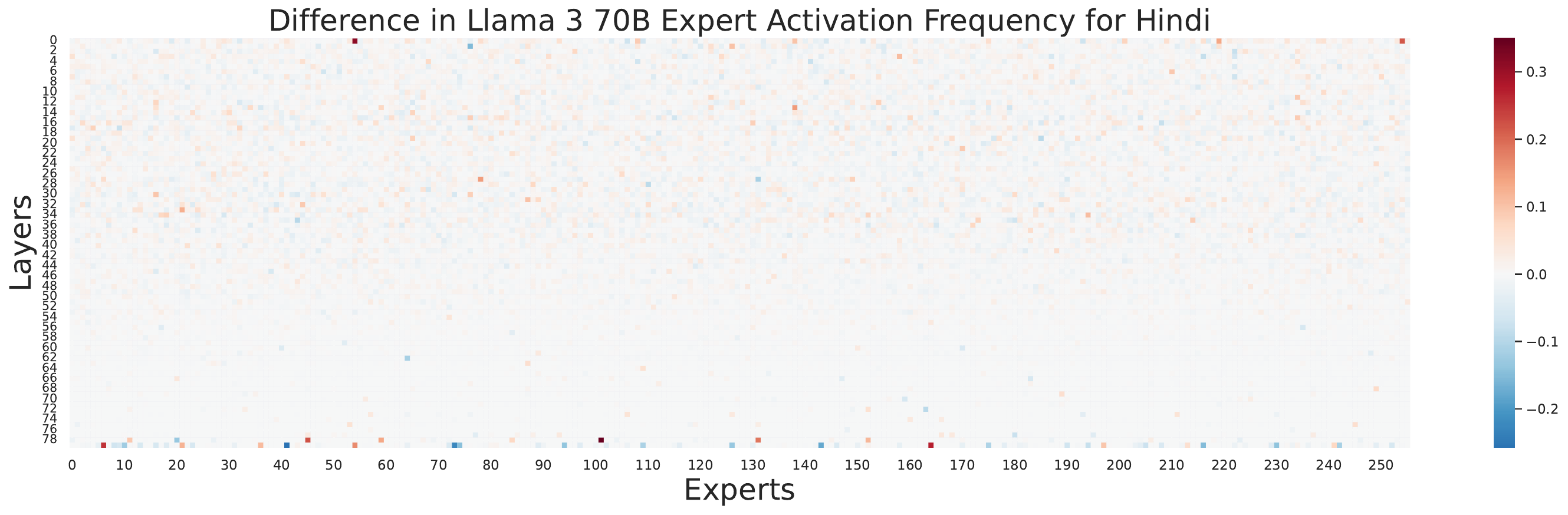}
    \vspace{-6mm}
  \end{minipage}

  \begin{minipage}{\linewidth}
    \centering
    \includegraphics[width=\linewidth]{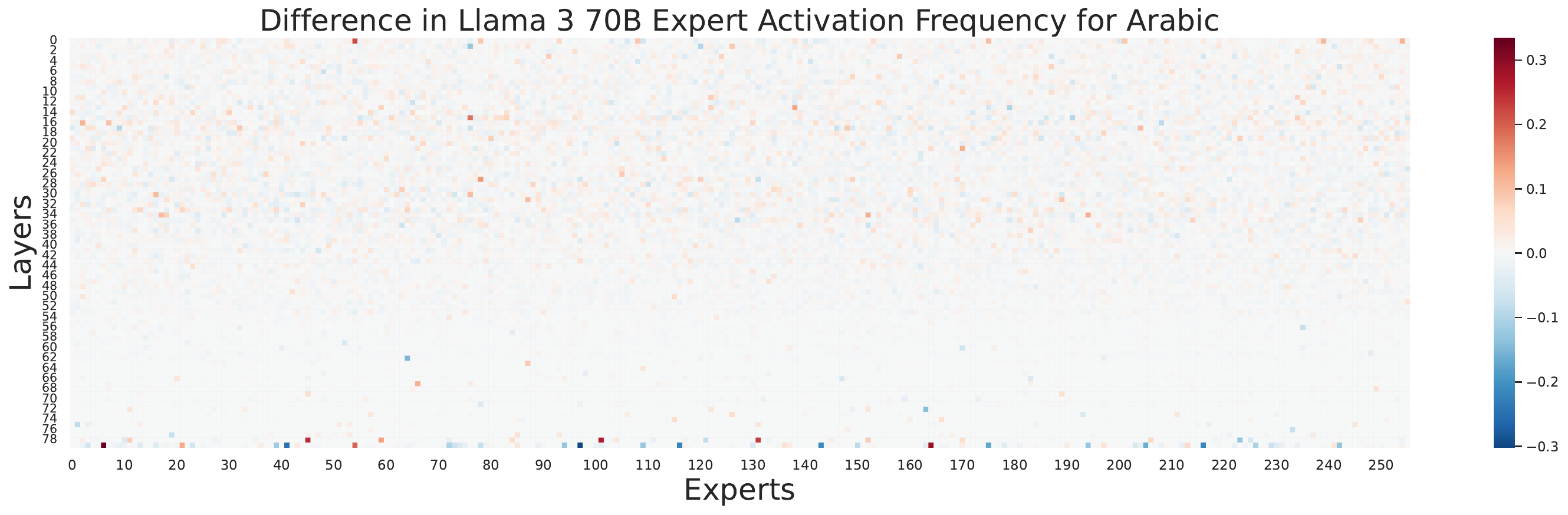}
    \vspace{-6mm}
  \end{minipage}
\end{figure*}

\clearpage

\begin{table*}[htb]
\centering
\small
\begin{tabular}{lccccccc}
\toprule
& \multicolumn{3}{c}{Expert activation frequency $\geq$ 0.5\%} & \multicolumn{3}{c}{Expert activation frequency $\geq$ 0.1\%} & \\
\cmidrule(lr){2-4} \cmidrule(lr){5-7}
Language & Proportion & Random & Experts & Proportion & Random & Experts & Origin \\
\midrule
English & 90.1\% & 4698.40 $\pm$ 5822.59 & 8.55 & 93.3\% & 65061.69 $\pm$ 91423.24 & 7.05 & 5.86 \\
French & 87.7\% & 2995.53 $\pm$ 3686.66 & 7.34 & 93.2\% & 3318.31 $\pm$ 3260.60 & 5.56 & 4.86 \\
Spanish & 87.5\% & 1680.02 $\pm$ 2196.96 & 7.87 & 93.1\% & 60.45 $\pm$ 21.81 & 6.41 & 5.62 \\
Portuguese & 87.5\% & 187.94 $\pm$ 137.34 & 7.61 & 93.2\% & 19.27 $\pm$ 5.33 & 6.02 & 5.23 \\
Bengali & 87.9\% & 21.47 $\pm$ 15.56 & 2.14 & 91.8\% & 16.43 $\pm$ 11.99 & 1.93 & 1.78 \\
Urdu & 85.7\% & 295.45 $\pm$ 166.41 & 5.08 & 91.7\% & 18.14 $\pm$ 3.96 & 3.91 & 3.23 \\
Hindi & 86.0\% & 1433.74 $\pm$ 1322.22 & 6.13 & 90.5\% & 94.81 $\pm$ 96.82 & 5.12 & 4.28 \\
Arabic & 84.8\% & 364.06 $\pm$ 106.11 & 13.28 & 91.7\% & 57.79 $\pm$ 32.34 & 8.97 & 6.54 \\
Chinese & 87.0\% & 1437.81 $\pm$ 1704.64 & 15.64 & 92.8\% & 36.01 $\pm$ 7.77 & 11.28 & 8.98 \\
\midrule
Average & 87.13\% & 1457.16 & 8.18 & 92.37\% & 7631.43 & 6.25 & 5.15 \\
\bottomrule
\end{tabular}
\caption{The perplexity results of Llama 3 70B. The smaller the value, the better the model performance.}
\label{tab:ppl-70b}
\end{table*}

\begin{table*}[htb]
\centering
\small
\begin{tabular}{lccccccc}
\toprule
& \multicolumn{3}{c}{Expert activation frequency $\geq$ 5\%} & \multicolumn{3}{c}{Expert activation frequency $\geq$ 1\%} & \\
\cmidrule(lr){2-4} \cmidrule(lr){5-7}
Language & Proportion & Random & Experts & Proportion & Random & Experts & Origin \\
\midrule
English & 77.1\% & 5641.07 $\pm$ 6999.41 & 20.54 & 89.6\% & 19.14 $\pm$ 1.41 & 11.78 & 9.43 \\
French & 76.3\% & 866.76 $\pm$ 100.47 & 34.77 & 90.8\% & 21.78 $\pm$ 0.80 & 14.50 & 10.09 \\
Spanish & 76.3\% & 89.55 $\pm$ 5.88 & 43.13 & 90.7\% & 23.45 $\pm$ 0.46 & 15.78 & 11.70 \\
Portuguese & 75.9\% & 177709.36 $\pm$ 250709.26 & 39.95 & 91.0\% & 23.26 $\pm$ 1.72 & 15.97 & 10.78 \\
Bengali & 67.5\% & 181486.00 $\pm$ 254425.86 & 37.24 & 89.5\% & 21.63 $\pm$ 3.99 & 10.76 & 5.22 \\
Urdu & 68.7\% & 145160.34 $\pm$ 165880.84 & 27.36 & 89.7\% & 16.27 $\pm$ 0.67 & 8.85 & 7.24 \\
Hindi & 70.9\% & 68728.70 $\pm$ 69301.79 & 11.02 & 89.8\% & 10.23 $\pm$ 0.21 & 6.05 & 4.86 \\
Arabic & 72.1\% & 426638.39 $\pm$ 602646.86 & 20.44 & 90.0\% & 14.08 $\pm$ 2.30 & 10.63 & 7.11 \\
Chinese & 76.4\% & 67081.13 $\pm$ 94749.98 & 34.38 & 91.7\% & 21.78 $\pm$ 1.68 & 11.21 & 9.28 \\
\midrule
Average & 73.5\% & 119266.81 & 29.87 & 90.3\% & 19.07 & 11.73 & 8.41 \\
\bottomrule
\end{tabular}
\caption{The perplexity results of Llama 2-Chat 7B. The smaller the value, the better the model performance.}
\label{tab:ppl-7b-chat}
\end{table*}

\begin{table*}[htb]
\centering
\small
\begin{tabular}{lccccccc}
\toprule
& \multicolumn{3}{c}{Expert activation frequency $\geq$ 5\%} & \multicolumn{3}{c}{Expert activation frequency $\geq$ 1\%} & \\
\cmidrule(lr){2-4} \cmidrule(lr){5-7}
Language & Proportion & Random & Experts & Proportion & Random & Experts & Origin \\
\midrule
English & 77.1\% & 41.7 $\pm$ 4.5 & 50.0 & 89.6\% & 49.3 $\pm$ 2.3 & \textbf{57.2} & 56.8 \\
French & 76.3\% & 15.3 $\pm$ 8.9 & 33.4 & 90.8\% & 41.9 $\pm$ 0.6 & 45.0 & 45.3 \\
Spanish & 76.3\% & 19.9 $\pm$ 14.5 & 38.7 & 90.7\% & 41.5 $\pm$ 1.0 & 42.4 & 44.5 \\
Portuguese & 75.9\% & 19.0 $\pm$ 13.5 & 31.9 & 91.0\% & 35.9 $\pm$ 1.6 & 38.2 & 39.8 \\
Chinese & 76.4\% & 28.5 $\pm$ 5.5 & 34.2 & 91.7\% & 37.6 $\pm$ 0.8 & 38.4 & 38.4 \\
\midrule
Average & 76.4\% & 24.9 & 37.6 & 90.8\% & 41.2 & 44.2 & 45.0 \\
\bottomrule
\end{tabular}
\caption{Accuracy (\%) of Llama 2-Chat 7B on the X-CSQA dataset.}
\label{tab:X-CSQA-7b-chat}
\end{table*}

\end{document}